\documentclass[lettersize,journal]{IEEEtran}

\usepackage{amsmath,amsfonts}
\usepackage{algorithmic}
\usepackage{algorithm}
\usepackage{array}
\usepackage[font=normalsize]{subfig}
\usepackage{textcomp}
\usepackage{stfloats}
\usepackage{url}
\usepackage{verbatim}
\usepackage{cite}
\usepackage{graphbox}

\usepackage{makecell}
\usepackage{graphicx} 
\usepackage{multirow}
\usepackage{multicol}
\usepackage{color}

\usepackage{bbm}

\newcommand{\figref}[1]{Fig. \ref{#1}}
\newcommand{\tabref}[1]{Table \ref{#1}}
\newcommand{\equref}[1]{Eq. (\ref{#1})}
\newcommand{\secref}[1]{Sec. \ref{#1}}

\definecolor{srcolor}{rgb}{1,0,0}

\usepackage[switch]{lineno}
\usepackage{pgfplots}
\pgfplotsset{width=4cm,compat=1.9}
\pgfplotsset{every tick label/.append style={font=\tiny}}
\usepackage{pifont}
\newcommand{\cmark}{\ding{51}}%
\newcommand{\xmark}{\ding{55}}%
\newcommand*\samethanks[1][\value{footnote}]{}
\renewcommand\footnotemark{}
\newcolumntype{M}[1]{>{\centering\arraybackslash}m{#1}}
\newcolumntype{N}{@{}m{0pt}@{}}

\usepackage[pagebackref=true,breaklinks=true,letterpaper=true,colorlinks,bookmarks=true]{hyperref}
\makeatletter
\def\hlinewd#1{%
\noalign{\ifnum0=`}\fi\hrule \@height #1 \futurelet
\reserved@a\@xhline}

\begin{document}

\title{AggMatch: Aggregating Pseudo Labels for Semi-Supervised Learning}

\author{
Jiwon Kim$^{*}$, Kwangrok Ryoo$^{*}$, Gyuseong Lee, Seokju Cho, Junyoung Seo,\\ Daehwan Kim, Hansang Cho, and Seungryong Kim$^\dagger$,
\thanks{$^*$Equal contribution}
\thanks{$^\dagger$Corresponding author}
\thanks{J. Kim, K. Ryoo, G. Lee, J. Seo, and S. Kim are with the Department of Computer Science and Engineering, Korea University, Seoul 02841, Korea. (E-mail: \{naancoco, kwangrok21, jpl358, se780, seungryong\_kim\}@korea.ac.kr).}
\thanks{S. Cho is with the Department of Computer Science and Engineering, Yonsei University, Seoul 03722, Korea. (E-mail: hamacojr@yonsei.ac.kr).}
\thanks{D. Kim and H. Cho are with Samsung Electro-Mechanics, Suwon 16674, Korea. (E-mail: 1378kdh@gmail.com, hansang.cho@samsung.com).}
}



\maketitle

\begin{abstract}
Semi-supervised learning (SSL) has recently proven to be an effective paradigm for leveraging a huge amount of unlabeled data while mitigating the reliance on large labeled data. Conventional methods focused on extracting a pseudo label from individual unlabeled data sample and thus they mostly struggled to handle inaccurate or noisy pseudo labels, which degenerate performance.   

In this paper, we address this limitation with a novel SSL framework for aggregating pseudo labels, called AggMatch, which refines initial pseudo labels by using different confident instances.
Specifically, we introduce an aggregation module for consistency regularization framework that aggregates the initial pseudo labels based on the similarity between the instances. To enlarge the aggregation candidates beyond the mini-batch, we present a class-balanced confidence-aware queue built with the momentum model, encouraging to provide more stable and consistent aggregation. We also propose a novel uncertainty-based confidence measure for the pseudo label by considering the consensus among multiple hypotheses with different subsets of the queue. We conduct experiments to demonstrate the effectiveness of AggMatch over the latest methods on standard benchmarks and provide extensive analyses.
\end{abstract}


\section{Introduction}
Recently, semi-supervised learning (SSL), an approach for learning a model on a large amount of \textit{unlabeled} data with few \textit{labeled} data, has become a promising solution to mitigate the reliance on large labeled dataset in numerous image processing and computer vision tasks~\cite{radosavovic2018data, yalniz2019billion, xie2020self, sohn2020fixmatch, chen2021semi}. Unlike the supervised learning paradigm~\cite{dosovitskiy2020image,chen2020dynamic, tolstikhin2021mlp, liu2021swin} that relies on \textit{ground-truth} labels, SSL poses additional challenges in that good \textit{pseudo} labels should be extracted from unlabeled data and used to learn the model.  

\begin{figure}[!t]
\begin{center}
\includegraphics[width=0.8\linewidth]{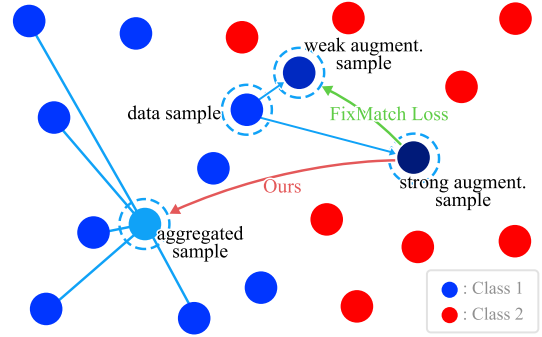}
\end{center}
\caption{\textbf{Intuition of our AggMatch in comparison to FixMatch~\cite{sohn2020fixmatch}.} By aggregating initial pseudo labels, AggMatch generates more confident pseudo labels, which boosts the classification performance.}
\label{Fig:intution}
\end{figure}

Most recent SSL approaches mainly followed two trends; \textit{pseudo-labeling} which encourages the model to follow the pseudo label from model's prediction itself~\cite{lee2013pseudo,shi2018transductive, xie2019unsupervised, arazo2020pseudo, zoph2020rethinking, pham2021meta}, which is closely related to entropy minimization~\cite{grandvalet2005semi, sajjadi2016mutual} and \textit{consistency regularization} which enables the model to produce the same prediction when perturbations are applied to the input~\cite{sajjadi2016regularization,laine2016temporal, tarvainen2017mean,miyato2018virtual,park2018adversarial,ke2019dual,xie2019unsupervised, verma2019interpolation,xie2020self} or the model~\cite{belkin2004regularization, laine2016temporal, tarvainen2017mean, ke2019dual, park2018adversarial, verma2019interpolation, xie2020self,zhang2020wcp}.
Very recently, advanced consistency regularization methods, called \textit{holistic approach}, have been presented by integrating the two above paradigms in a unified framework. For instance, FixMatch~\cite{sohn2020fixmatch} first generates a pseudo label using the model's prediction on weakly-augmented unlabeled data and then encourages the prediction from strongly-augmented unlabeled data to follow the pseudo label with a confidence threshold. Although FixMatch~\cite{sohn2020fixmatch} serves as one of the state-of-the-art frameworks~\cite{berthelot2019mixmatch, berthelot2019remixmatch, sohn2020fixmatch, kuo2020featmatch, hu2021simple}, it inherits the limitations of pseudo-labeling, i.e., confirmation bias and noise sensitivity. 
In addition, confidence, measured by only considering the prediction itself as in FixMatch~\cite{sohn2020fixmatch}, has limited discriminative power and often cannot catch false positives at the early stage of training, which may be confident pseudo label candidates, thus hindering performance boosting. 

\begin{table}
\caption{\textbf{Summary of our key results on standard CIFAR-10~\cite{krizhevsky2009learning} with 40 labels and on asymmetric label noise setting.}}
\begin{center}
\scalebox{1}{
\begin{tabular}{c|c|c}
\hlinewd{0.8pt}
Datasets &
  \begin{tabular}[c]{@{}c@{}}CIFAR-10~\cite{krizhevsky2009learning}\\ (40 labels)\end{tabular} & 
  \begin{tabular}[c]{@{}c@{}}CIFAR-10~\cite{krizhevsky2009learning}\\ (noise 50\% + 250 labels)\end{tabular} \\ \hline
\begin{tabular}[c]{@{}c@{}}FixMatch~\cite{sohn2020fixmatch}\\ (previous SOTA)\end{tabular} &
   86.19{$\pm$3.37} & 79.68{$\pm$4.84}
   \\ \hline
\begin{tabular}[c]{@{}c@{}}\textbf{AggMatch}\\ (Ours)\end{tabular} &
   \textbf{92.54}{$\pm$0.76} & \textbf{85.41}{$\pm$7.07}
   \\ 
\hlinewd{0.8pt}
\end{tabular}
}
 \label{tab:mot}
\end{center}
\end{table}

Leveraging the relationships between different instances would be an alternative solution to address the aforementioned limitations of recent SSL methods~\cite{berthelot2019mixmatch, berthelot2019remixmatch, sohn2020fixmatch, kuo2020featmatch, hu2021simple}, by refining the pseudo label and measuring its confidence.
Interestingly, traditional SSL approaches such as \textit{label propagation}~\cite{zhur2002learning, jaakkola2002partially, zhou2004learning} that propagate the class distribution of labeled data to unlabeled data by considering the manifold structure also leverage the relationships between different instances, but their scalability is limited as the graph should be built for all the data, and they are not differentiable to be learned. Very recently, PAWS~\cite{assran2021semi} attempted to propagate the class distribution of labeled data to unlabeled data within the current mini-batch, similar to label propagation~\cite{iscen2019label, li2020density}, but it highly relies on a large amount of labeled data, thereby limiting the performance on label-scarce scenarios or with noisy labeled data.   

In this paper, we present a novel SSL framework, called AggMatch, that uses the relationships between different instances to achieve more confident pseudo label as shown in~\figref{Fig:intution}. Inspired by cost aggregation in stereo matching literature~\cite{he2016deep, sun2018pwc} and self-attention module in Transformers~\cite{vaswani2017attention}, we present an aggregation module that aggregates initial pseudo labels from labeled and unlabeled data by considering the affinity between the instances, which is measured by using both the feature and class distribution itself. To enlarge the aggregation candidates beyond the current mini-batch, we utilize a queue that memorizes information of previous batch samples during training, allowing for improved scalability, as in~\cite{he2020momentum, jain2020contrastive, liu2021swin}. In addition, we present a class-balanced confidence-aware queue, built by considering the confidence and class distribution and updated with a momentum, which encourages better aggregation. Finally, we also present a novel confidence measure for the pseudo label by measuring the consensus among possible pseudo labels with multiple disjoint subsets of the queue, which helps to improve the robustness to noise.
Experiments on standard SSL benchmarks for image classification~\cite{krizhevsky2009learning, netzer2011reading} and with noisy settings~\cite{sukhbaatar2014training, patrini2017making} prove the effectiveness of our approach over the latest methods. As summarized in~\tabref{tab:mot}, we show the new  state-of-the-art performance on CIFAR-10~\cite{krizhevsky2009learning} with 40 labels and with 50\% noise on 250 labels, which is a large margin of improvement compared to FixMath~\cite{sohn2020fixmatch}, one of the state-of-the-arts.  

The rest of this paper is organized as follows. We discuss related works in ~\secref{sec:related} and introduce the background in ~\secref{sec:preliminaries}. \secref{sec:algo} presents the details of our algorithm. We conduct experiments on extensive SSL benchmarks and provide an ablation study to validate and analyze components of our approach in~\secref{sec:exp}. In~\secref{sec:conc}, we conclude this paper. 

\section{Related Work}
\label{sec:related}
In this section, we first review existing methods for SSL, which are categorized as pseudo-labeling, consistency regularization, and label propagation methods. We then review uncertainty estimation for SSL.
\subsection{Pseudo-Labeling}
\begin{figure}[t]
    \centering
	\subfloat[Pseudo-Labeling]
	{{\includegraphics[width=0.5\linewidth]{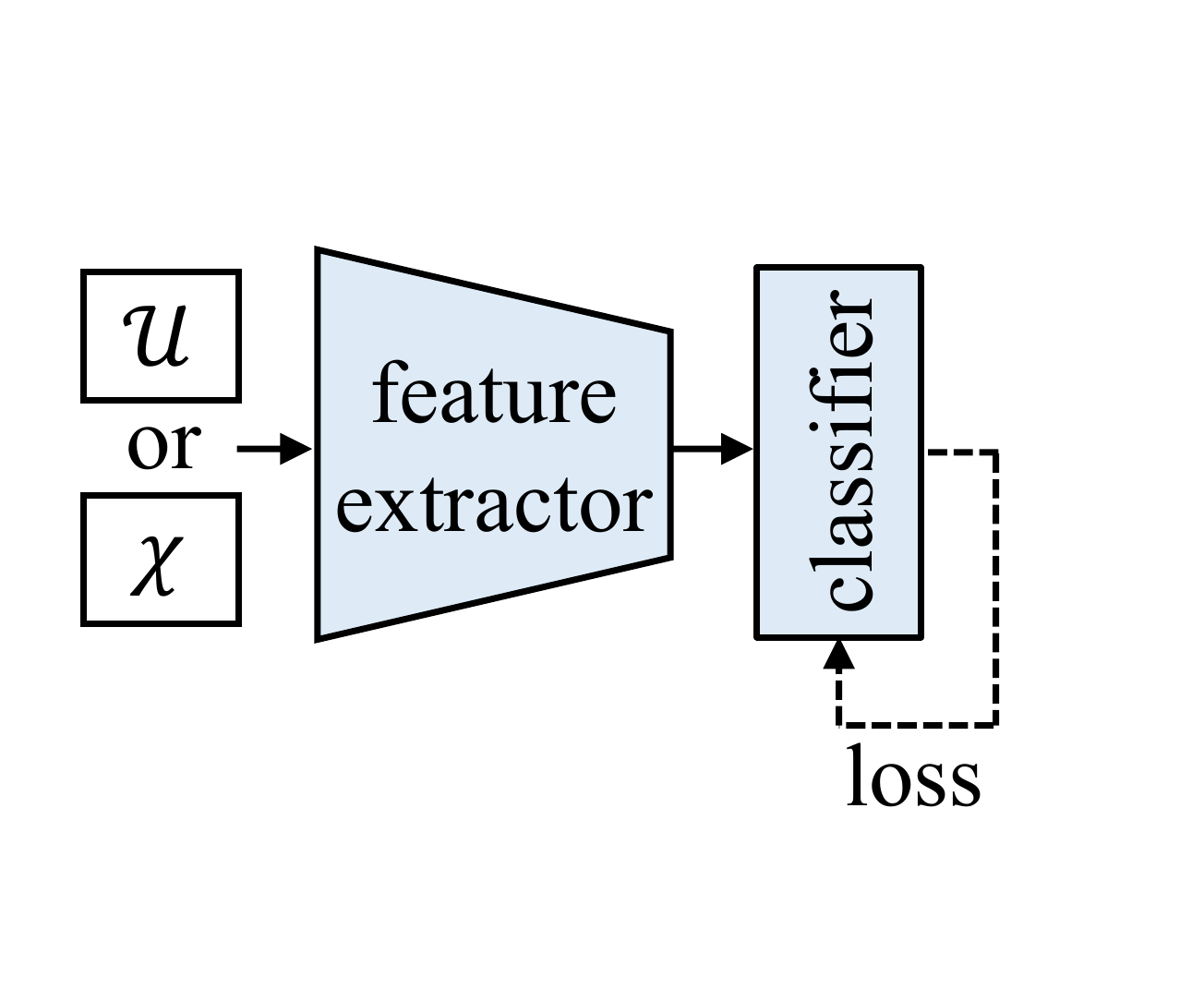}}}\hfill
	\subfloat[Consistency Regularization]
 	{\includegraphics[width=0.5\linewidth]{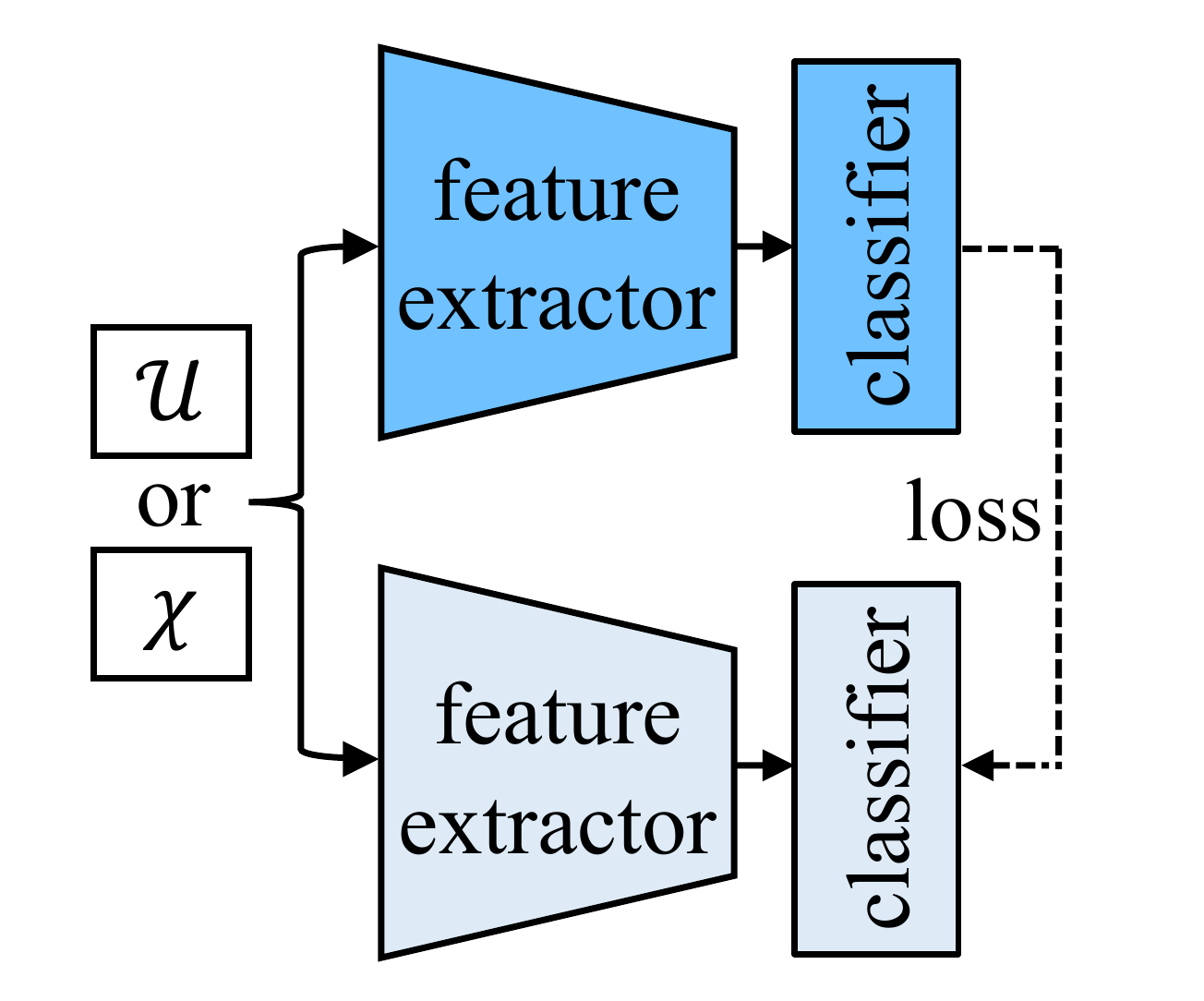}}\hfill \\
 	\subfloat[Label Propagation]
 	{\includegraphics[width=0.5\linewidth]{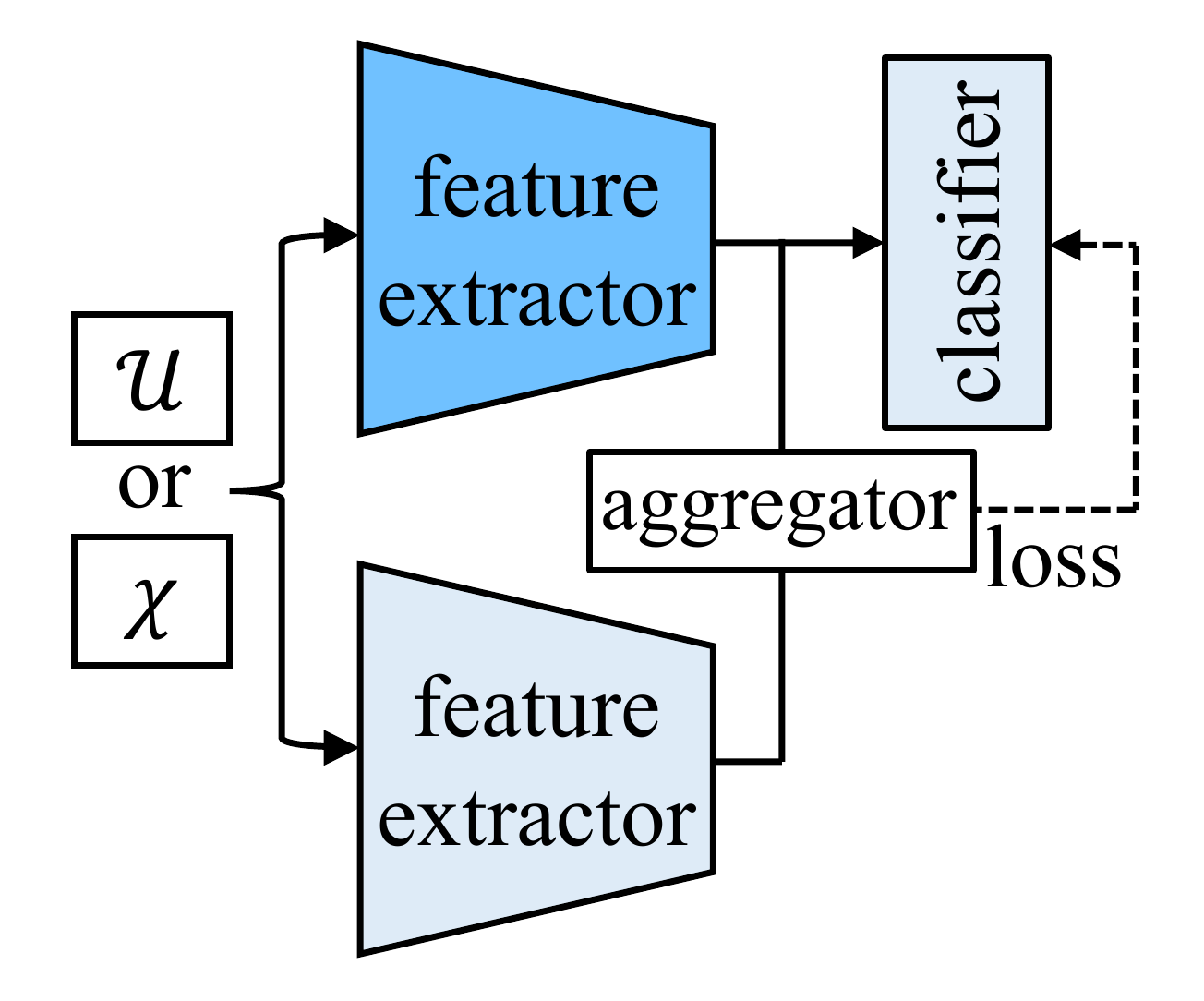}}\hfill
 	\subfloat[Ours]
 	{\includegraphics[width=0.5\linewidth]{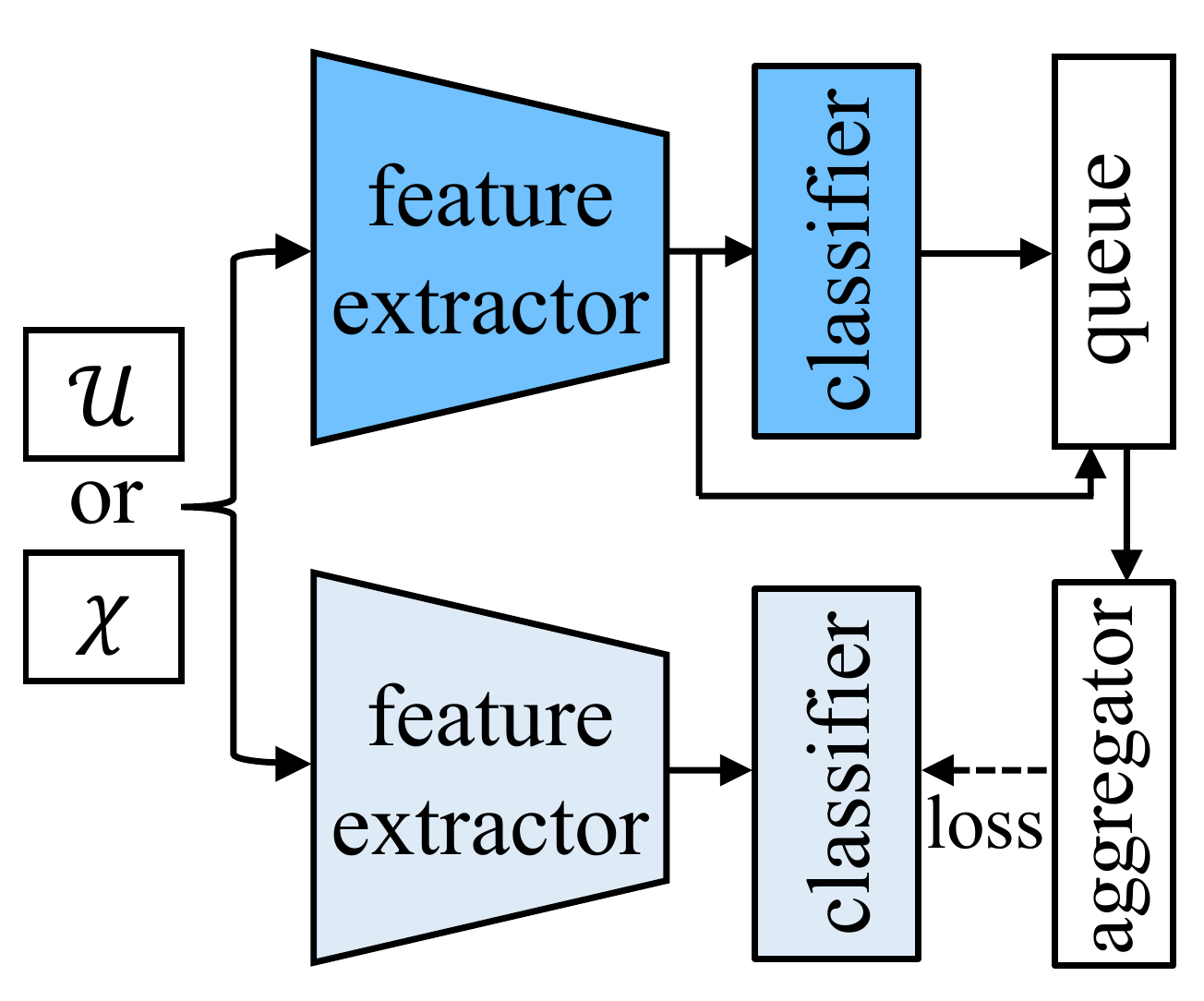}}\hfill \\
	\caption{\textbf{Motivation of our approach:} (a) pseudo-labeling methods~\cite{lee2013pseudo, tarvainen2017mean, xie2020self, pham2021meta}, (b) consistency regularization methods~\cite{sajjadi2016regularization, laine2016temporal}, (c) label propagation methods~\cite{iscen2019label, li2020density, assran2021semi}, and (d) ours.}
	\label{cft_contrastive}
\end{figure}

Pseudo-labeling, also known as self-training~\cite{zoph2020rethinking}, is based on the idea of utilizing the model's prediction itself to explicitly generate an artificial label for unlabeled data as shown in~\figref{cft_contrastive}(a). The pseudo labels from the model predictions can be defined as soft (continuous distribution) or hard (one-hot distribution) pseudo labels~\cite{lee2013pseudo, shi2018transductive, xie2019unsupervised, iscen2019label,arazo2020pseudo, zoph2020rethinking, pham2021meta}. More specifically, a soft and hard version of pseudo-labeling can be designed by sharpening the predicted distribution and introducing argmax of the prediction, respectively. It is closely related to entropy minimization~\cite{grandvalet2005semi, sajjadi2016mutual}, where the model's predictions are encouraged to be low-entropy (i.e., high-confidence) on unlabeled data. Specifically, they~\cite{lee2013pseudo, shi2018transductive,iscen2019label, arazo2020pseudo} and earlier related methods~\cite{scudder1965probability, yarowsky1995unsupervised, riloff1996automatically} first trained a model on labeled samples, and then fined-tuned the pre-trained model using the additional pseudo labels. 
It is also extended to an explicit teacher-student configuration~\cite{xie2020self}, formulated by a pair of networks. In general, a teacher network generates pseudo labels for unlabeled samples, which are used for training a student network. Although pseudo-labeling is simple and intuitive, the initialized model is vulnerable to bias from the limited amount of labeled data. In addition, it heavily relies on the quality of the model’s prediction, thus suffering from confirmation bias~\cite{arazo2020pseudo}, where the prediction errors would accumulate and worsen the model performance.

\subsection{Consistency Regularization} 
Consistency regularization is based on the manifold assumption or the smoothness assumption~\cite{chapelle2009semi} and consequently, it can be regarded to find a smooth manifold given few labeled and a large amount of unlabeled data. As exemplified in~\figref{cft_contrastive}(b), it constraints consistency between model predictions after injecting different noise to the input, hidden states or model parameters, assuming to keep the class semantics unaffected. This idea was first proposed in~\cite{bachman2014learning} and popularized by~\cite{sajjadi2016regularization, laine2016temporal}. Specifically, methods using stochastic data augmentation as perturbation can be defined as follows; domain-specific data augmentation~\cite{laine2016temporal, french2017self, berthelot2019mixmatch, berthelot2019remixmatch}, stochastic regularization~\cite{sajjadi2016regularization, laine2016temporal}, random max pooling~\cite{sajjadi2016regularization}, or adversarial transformation~\cite{miyato2018virtual}. On the other hand, other methods apply the perturbation on the model, such as Dropout~\cite{belkin2004regularization, laine2016temporal, tarvainen2017mean, ke2019dual, park2018adversarial, xie2020self}, running average or time ensemble of past model predictions at different time step (EMA or SWA)~\cite{laine2016temporal, tarvainen2017mean, verma2019interpolation, athiwaratkun2018there}, stochastic depth~\cite{athiwaratkun2018there} or an adversarial perturbation on model's parameter (DropConnect)~\cite{zhang2020wcp}.
Both L2 norm~\cite{sajjadi2016regularization, laine2016temporal, tarvainen2017mean} and cross entropy~\cite{miyato2018virtual, xie2019unsupervised, berthelot2019remixmatch, sohn2020fixmatch} are available as unsupervised loss function for consistency regularization. A further extension of the consistency regularization methods, also called a holistic approach, combines pseudo-labeling and consistency regularization in a unified loss function~\cite{berthelot2019mixmatch, berthelot2019remixmatch, sohn2020fixmatch, kuo2020featmatch}.  
However, the limitation of those consistency regularization methods is that they primarily focus on the relationship between differently augmented samples for one instance itself where once wrong pseudo labeling occurs, model can be overfitted and contaminated. 

\subsection{Label Propagation} 
The key assumption in label propagation is that data points in the same manifold can be considered to share the same semantic label distribution~\cite{zhur2002learning}. This can be closely related to the manifold assumption of graph-based learning algorithms in which nodes consist of the labeled and unlabeled samples, and edges are weighted by the similarity between those nodes. The label propagation aggregates pairwise similarity between features and class probability to infer pseudo labels, as illustrated in~\figref{cft_contrastive}(c).
Traditionally, it was performed in a transductive setting~\cite{zhu2003semi, zhou2004learning}, propagating labels within a given set of unlabeled examples or in an inductive setting~\cite{weston2012deep, tarvainen2017mean}, generalizing inference to new unseen data, while the original training data are discarded.

Recently, some methods~\cite{iscen2019label, li2020density} integrated label propagation in an iterative way with pseudo-labeling. They alternate between training the network with pseudo labels and constructing the affinity graph of nearest neighbors based on the similarity between the feature representations.  
Recent methods in consistency regularization~\cite{assran2021semi, hu2021simple} combine the aggregation technique based on similarity of a mini-batch of labeled or other unlabeled samples to define the aggregated pseudo labels. However, their similarity measurements are implemented between in-batch samples, and then can limits performance depending on the size of the mini-batches.
 
\subsection{Uncertainty Estimation}
Estimating the uncertainty over the predictions of neural networks~\cite{mackay1992practical, chen2014stochastic, welling2011bayesian} often aims to regress a distribution of output~\cite{kendall2017uncertainties, ilg2018uncertainty, kendall2018multi, poggi2020uncertainty}. Uncertainty estimation can be divided into two approaches; \textit{empirical} and \textit{predictive} approach. In \textit{empirical} methods, it can be approximated by sampling a finite number of weight configurations for a given network and computing the mean and variance of the predictions~\cite{graves2011practical, blundell2015weight,lakshminarayanan2016simple}. On the other hand, output distributions can also be trained by \textit{predictive} methods, i.e., a Laplacian or Gaussian distribution~\cite{kendall2017uncertainties}. 

For classification in SSL, the idea of confidence can be represented as a class probability of pseudo-labeling in general. A confidence-based strategy was previously used in~\cite{rosenberg2005semi, french2017self} along with pseudo labeling, encouraging the model to use samples with high confidence (i.e., low entropy). Recently, such confidence strategy is implemented by ~\textit{thresholding} by leaving samples above a pre-defined threshold and filtering out noisy unlabeled data as proposed in UDA~\cite{xie2019unsupervised} and FixMatch~\cite{sohn2020fixmatch}.

However, the pre-defined constant threshold fails to consider different learning status per training iteration and learning difficulty level per sample, leading to slow convergence speed. To solve the trade-off between the quality and the quantity of pseudo labels, we combine uncertainty-based confidence on the unsupervised loss to manipulate the loss magnitude dynamically depending on the uncertainty estimation.

\begin{figure*}[t]
\begin{center}
\includegraphics[width=1\linewidth]{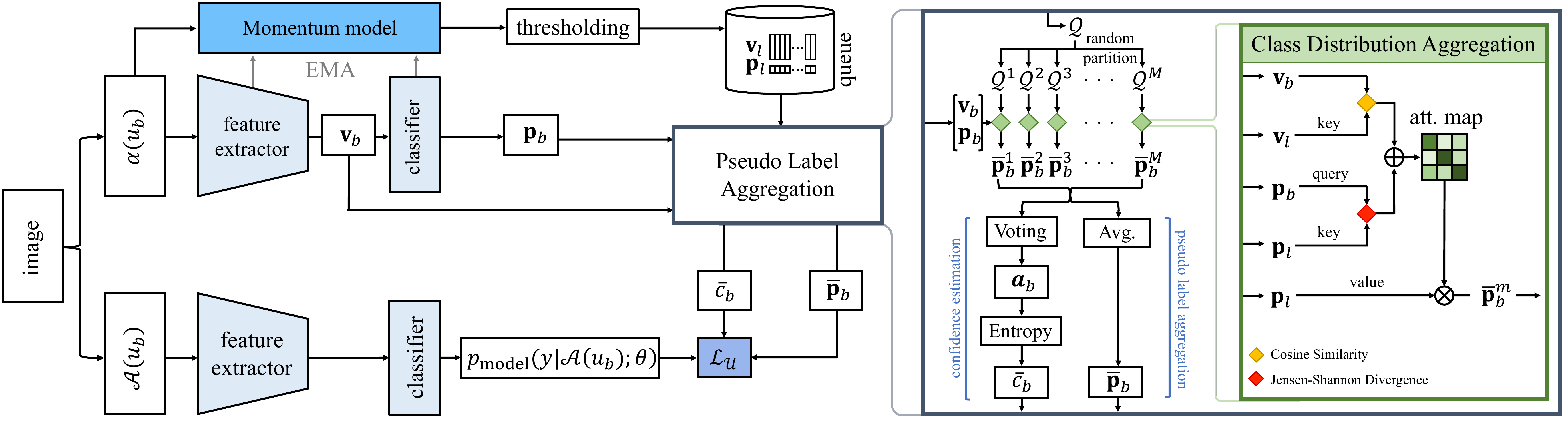}
\end{center}

\caption{\textbf{Network configuration:} For differently perturbed images by weak and strong augmentations, we regularize the model to generate consistent predictions between them. Class distributions of the weakly augmented images are refined by our aggregation module with the help of a large and consistent queue. Our confidence estimator estimates the unreliable samples voted to be uncertain from the hypotheses generated from the randomly and evenly partitioned queue.}
\label{network_overall}
\end{figure*}
\section{Preliminaries}
\label{sec:preliminaries}
Let us define a batch of \textit{labeled} instances as $\mathcal{X}=\{(x_b,y_b):b\in(1,...,B)\}$, where $x_b$ is an instance and $y_b$ is a label representing one of $Y$ labels. In addition, let us define a batch of \textit{unlabeled} instances as $\mathcal{U}=\{u_b:b\in(1,...,\mu B)\}$, where $\mu$ is a hyper-parameter that determines the size of $\mathcal{U}$ relative to $\mathcal{X}$. The objective of SSL is to simultaneously use $\mathcal{X}$ and $\mathcal{U}$ to train a model $p_\mathrm{model}(y|r;\theta)$ taking an instance $r$ as input and outputting a distribution over class labels $y$ with parameters $\theta$, where the learned model is expected to yield better performance than solely using $\mathcal{X}$. 
In general, $p_\mathrm{model}$ consists of two parts; \textit{feature extractor}, which extracts a feature $\mathbf{v}$ from an instance $r$, and \textit{classifier}, which measures a class probability $\mathbf{p}$ from $\mathbf{v}$.

For SSL, consistency regularization methods~\cite{bachman2014learning,sajjadi2016regularization, laine2016temporal} rely on the assumption that the model should generate similar predictions when perturbed versions of the same instance are fed. 
FixMatch~\cite{sohn2020fixmatch} extends this by performing two types of augmentations such as \textit{weak} and \textit{strong}, denoted by $\alpha(\cdot)$ and $\mathcal{A}(\cdot)$, and enforcing the consistency between them with pseudo-labeling~\cite{lee2013pseudo}. The pseudo label is always obtained from \emph{weakly}-augmented version of an unlabeled instance, which enforces the consistency to the model's output for \emph{strongly}-augmented version of the same instance. The loss function is defined as
\begin{equation}
    \frac{1}{\mu B}\sum_{b=1}^{\mu B} {
    c_b
    \mathcal{D}(\mathbf{q}_b,p_\mathrm{model}(y|\mathcal{A}(u_b);\theta))}, 
\end{equation}
where a cross-entropy is used for $\mathcal{D}$, and $\mathbf{p}_b=p_\mathrm{model}(y|\alpha(u_b);\theta)$ denotes a class probability, ${c}_b$ denotes a confidence of $\mathbf{p}_b$ such that ${c}_b=\mathbbm{1}(\mathrm{max}(\mathbf{p}_b) \geq \tau)$. More specifically, the pseudo label $\mathbf{q}_b$ for $\mathbf{p}_b$ can be defined as continuous distribution (\textit{soft}) based on sharpening or one-hot distribution (\textit{hard}) based on argmax operation, respectively. 

One of main limitations of existing SSL methods, including consistency regularization~\cite{sajjadi2016regularization, laine2016temporal, tarvainen2017mean}, pseudo-labeling~\cite{scudder1965probability, yarowsky1995unsupervised, riloff1996automatically, lee2013pseudo} and holistic approaches~\cite{berthelot2019mixmatch, berthelot2019remixmatch}, is that they can only generate confident pseudo labels when a model is sufficiently trained. In this case, the immature model at the early stage of the training may be significantly degraded depending on the quality of pseudo-labeling. For instance, if the labeled data $\mathcal{X}$ has different distribution with unlabeled data $\mathcal{U}$, the size of $\mathcal{X}$ is not enough, or $\mathcal{X}$ and $\mathcal{U}$ are contaminated by noise and bias, the pseudo-labeling overfits to incorrect pseudo labels, which leads to optimizing the model's parameters to undesirable directions, also known as the confirmation bias~\cite{arazo2020pseudo}. To prevent the aforementioned problem, the confidence-based strategy~\cite{rosenberg2005semi, french2017self} is combined with pseudo-labeling to focus on pseudo labels with high confidence by setting a predefined confidence threshold~\cite{xie2019unsupervised, sohn2020fixmatch}. However, estimating the confidence of a pseudo label with a simple threshold can be too strong constraint in early iterations of training. A huge amount of true positive samples may be discarded as false positive without considering the inherent learning difficulties of different samples and learning status during the training.

\section{Our Algorithm: AggMatch}
\label{sec:algo}
\subsection{Motivation and Overview}
In this section, we address the limitations aforementioned of conventional SSL methods~\cite{lee2013pseudo, sajjadi2016regularization, laine2016temporal, tarvainen2017mean, berthelot2019mixmatch, berthelot2019remixmatch,xie2019unsupervised, sohn2020fixmatch} by refining an initial class distribution individually extracted from a \textit{weakly}-augmented version, $p_\mathrm{model}(y|\alpha(u);\theta)$. Based on the observation that similar instance should have similar class distribution~\cite{zhou2004learning}, we present a novel semi-supervised learning framework that \textit{aggregates} initial class distributions both with labeled data $\mathcal{X}$ and unlabeled data $\mathcal{U}$ to provide better pseudo labels for a \emph{strongly}-augmented version, $p_\mathrm{model}(y|\mathcal{A}(u_b);\theta)$. 

In this paper, we mainly study how to accurately \textit{measure} the similarity between instances and how to effectively \textit{aggregate} the class distributions from the candidates by considering the similarities. Inspired by cost aggregation in stereo matching literature~\cite{he2016deep, sun2018pwc} and self-attention module in Transformers~\cite{vaswani2017attention}, we present to measure the similarity of feature embedding by means of query, key and value. A query can be a feature $\mathbf{v}_b$ of an instance and keys can be other features $\mathbf{v}_l$ in $\mathcal{X}$ and $\mathcal{U}$, as well as class distribution $\mathbf{p}_b$ and $\mathbf{p}_l$, respectively. The corresponding initial class distributions $\mathbf{p}_l$ are then considered as values. A  na\"ive framework~\cite{assran2021semi, hu2021simple} is to let this procedure perform within a current batch. The quality of aggregated distribution highly depends on the size of the batch, but a large batch size cannot be guaranteed by limited GPU memories. To resolve this issue, we utilize a queue that memorizes information of previous batch samples during training. By allowing that, the confident samples are selectively enqueued with a momentum model, and the queue slowly and stably changes, enabling consistent propagation. We also measure the confidence of pseudo label by measuring the consensus among its multiple hypotheses with multiple subsets of the queue. Our overall network architecture is illustrated in~\figref{network_overall}.

\subsection{Class Distribution Aggregation}
A class distribution independently predicted from an individual instance often contains ambiguous values and often generates erroneous pseudo labels. To remedy this, we propose an aggregation module that aims to refine the ambiguous or noisy class distributions utilizing the relationship between other confidence-aware samples. Specifically, a class distribution $\mathbf{p}_b$ for $u_b$ is aggregated by $\mathbf{p}_l$ of other instances $r_l$ from the set of $\mathcal{X}$ and $\mathcal{U}$, by considering the similarity between them as
\begin{equation}
    \overline{\mathbf{p}}_b = \sum_{l} \left(\frac{ \exp(\mathcal{S}(u_b,r_l)/\tau_\mathrm{sim}) }
{  {\sum_j} \exp(\mathcal{S}(u_b,{r_j})/\tau_\mathrm{sim}) } \right) \mathbf{p}_l,
\label{equ:aggregation_module}
\end{equation}
where $l$ and $j$ are indexes for all the samples in the set of $\mathcal{X}$ and $\mathcal{U}$, and $\tau_\mathrm{sim}$ is a temperature.
Especially, the similarity function $\mathcal{S}$ measures attention weights between the instances. 
Intuitively, one straightforward way is to only consider class distribution $\mathbf{p}$ based on a class similarity term, but it disregards the noise involved in the class distribution. We present to additionally use a feature similarity term for feature affinity between $u_b$ and $r_l$ defined such that
\begin{equation}
    \mathcal{S}(u_b,r_l)
    = (\mathbf{v}_b\cdot\mathbf{v}_l) / \|\mathbf{v}_b\|\|\mathbf{v}_l\| +\lambda_\mathrm{sim}\mathrm{JS}\left(\mathbf{p}_b\|\mathbf{p}_l\right),
    \label{equ:sim_func}
\end{equation}
where the first term denotes a feature similarity and the second term denotes a class similarity, and $\lambda_\mathrm{sim}$ is a weight parameter. We measure similarity between features by cosine similarity~\cite{kuo2020featmatch, assran2021semi}. We use Jensen-Shannon distance~\cite{cao2019gcnet} $\mathrm{JS}(\cdot\|\cdot)$ for measuring the similarity between distribution $\mathbf{p}_b$ and $\mathbf{p}_l$.
Thanks to this attention, unlike conventional consistency regularization methods~\cite{sajjadi2016regularization, laine2016temporal,tarvainen2017mean,xie2019unsupervised}, ignoring relationships between different instances, our aggregation explicitly exploits such relationships to generate more confident pseudo labels and thus enforces more stable convergence. 
\begin{figure}[t]
\begin{center}
\includegraphics[width=1.0\linewidth]{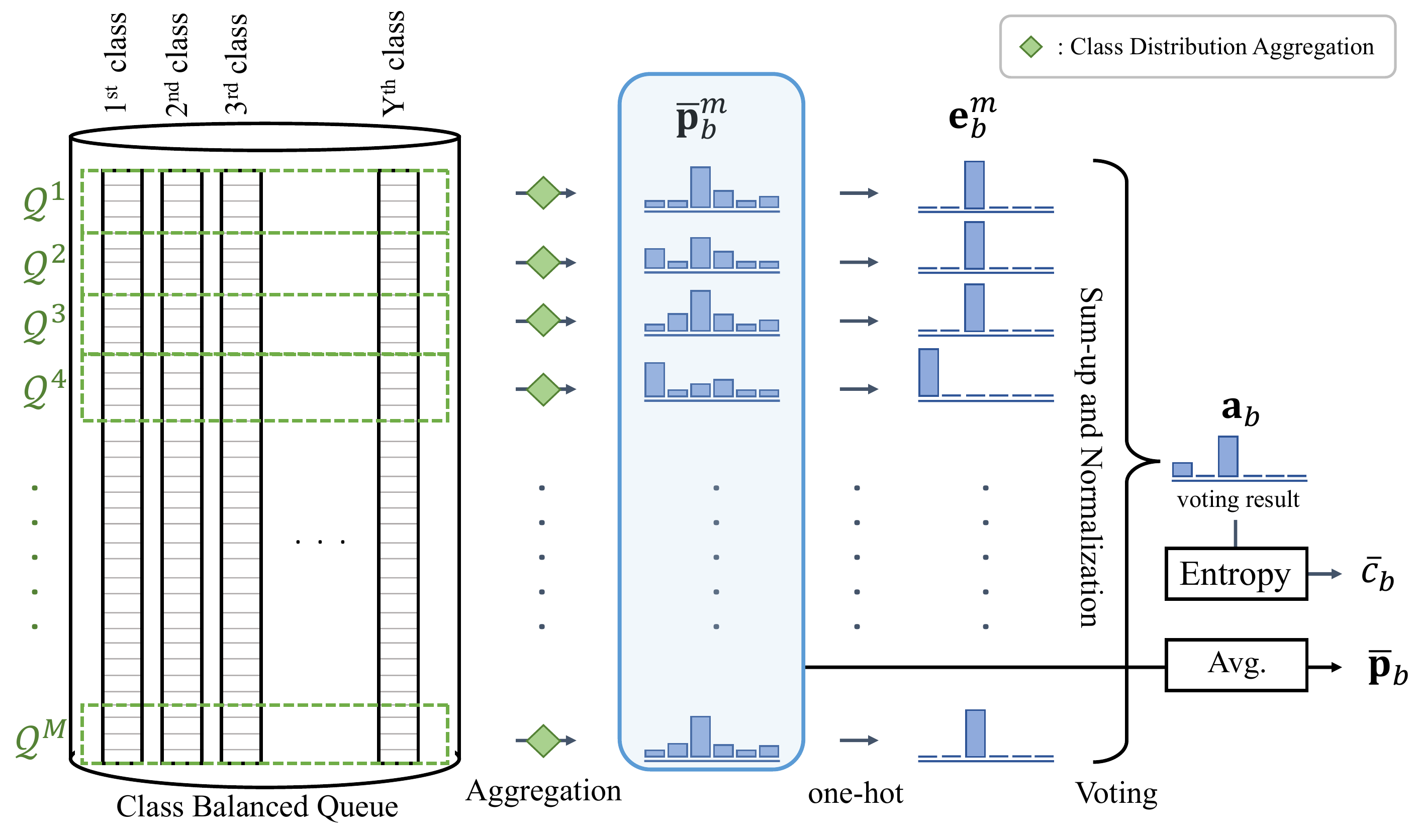}
\end{center}
\caption{\textbf{Illustration of confidence estimation for pseudo label:} We define the confidence of pseudo label as consensus among multiple hypotheses of pseudo label. In specific, the queue is evenly and randomly partitioned as $M$ disjoint subsets to generate the hypotheses. We build aggregated pseudo labels using multiple disjoint subsets of the queue. We get a one-hot hard pseudo label and then apply the voting process for all $M$ predictions to achieve empirical probability, where entropy is used as the confidence of pseudo label.}
\label{fig:confidence_process}
\end{figure}

\subsection{Class-Balanced Confidence-Aware Queue}
An aggregation among current batches $\mathcal{X}$ and $\mathcal{U}$ may be sensitive to the quality of sampled instances in the batch, which prohibits stable convergence. To alleviate this, we exploit a queue that memorizes information of previous batch samples during training, which helps enlarge the aggregation candidates. In specific, at each training step, current batches are enqueued and the oldest are dequeued, enabling the queue to impose ensembles of knowledge among all the samples implicitly. To prevent the propagation of noisy samples, we leave out low confident predictions below a threshold $\tau$. In other fields such as depth completion~\cite{he2016deep, sun2018pwc}, it has also been proven that propagating only confident samples yields better performance. 
The queue is formally defined as $\mathcal{Q}=\{(\mathbf{v}_l,\mathbf{p}_l)\}$. Note that PAWS~\cite{assran2021semi} also tried to exploit the similarity with labeled samples, but the propagation only relies on very sparse labeled samples. Unlike this, thanks to the proposed confident-aware queue, we can easily enlarge the candidate samples by using the unlabeled samples. 
\subsubsection{Class Balancing}
The use of samples in a class-imbalanced queue for aggregation may suffer from a biased prediction problem. We design a class-balanced queue to overcome this, where each sample is enqueued according to its pseudo label prediction. The queue for class $y$ is defined as $\mathcal{Q}_y=\{(\mathbf{v}_{y,l},\mathbf{p}_{y,l}):l\in(1,...,L)\}$ with the number of samples per class is $L$, which is same for all classes.
\subsubsection{Momentum Update}
Using the queue can help enlarge the aggregation candidates, but directly updating the queue may cause consistency problems because measuring attentions between inconsistent features from rapidly updated feature extractor can be poorly operated as in~\cite{he2020momentum, jain2020contrastive, liu2021swin}. To address this issue, we propose a momentum technique for queue update, which mitigates the problem of inconsistency between features from the queue and current batch. The momentum model with parameter $\theta_m$ is manipulated by a momentum coefficient $\lambda_m$ and model parameter $\theta$ such that
\begin{equation}
    \theta_m \leftarrow \lambda_m\theta_m + (1-\lambda_m) \theta.
    \label{equ:momentum}
\end{equation}

It should be noted that Mean Teacher~\cite{tarvainen2017mean} also uses an exponential moving average of model parameters, which was empirically found to improve results. Compared to this, we differ in that we only use this for updating the queue.

\subsection{Confidence Estimation for Pseudo Label} 
SSL generally does not take the model uncertainty into account. FixMatch~\cite{sohn2020fixmatch} uses a simple thresholding technique as confidence measure for pseudo label, but it requires a hand-tuned threshold parameter and also imposes overly strong constraint in early iterations so that most possible confident pseudo labels would be rejected due to their class probability scores below the threshold.

To tackle the aforementioned issues, we present to measure a confidence of aggregated soft pseudo label $\overline{\mathbf{p}}_b$, based on the consensus among multiple hypotheses of pseudo labels. To generate those hypotheses, the queue $\mathcal{Q}$ is evenly and randomly partitioned as $M$ disjoint subsets at each class, i.e., $\mathcal{Q}^m$ for $m=1,...,M$. Given the subsets $\mathcal{Q}^m$, similar to~\equref{equ:aggregation_module}, we aggregate to achieve $\overline{\mathbf{p}}^m_b$. Aggregated class probabilities then undergo a voting process. In specific, we sum up one-hot representation of each prediction such that $\mathbf{a}_b=\frac{1}{M}\sum\nolimits_{m}{\textbf{e}^m_b}$, where $\textbf{e}^m_b$ denotes a one-hot vector by operating argmax on $\overline{\mathbf{p}}^m_b$. From such empirical probability $\mathbf{a}_b$, which represents the occurrence of each class in the hypotheses, we measure the confidence $\overline{c}_b$ of the pseudo label by an entropy~\cite{zhang2019mitigating} such that $\overline{c}_b=\mathrm{exp}\left(\sum\mathbf{a}_b\log{\mathbf{a}_b}\right)$. The final pseudo label $\overline{\mathbf{p}}_b$ can be simply achieved by averaging $\overline{\mathbf{p}}^m_b$ for $m$. By estimating the accurate confidence of pseudo label, our unsupervised loss function enables boosting the performance for SSL even in early stages of training. Our overall confidence estimation process is described in~\figref{fig:confidence_process}.

\subsection{Loss Functions}
Our loss function for unlabeled batch is defined with the aggregated pseudo label $\overline{\mathbf{q}}_b$ and confidence prediction $\overline{c}_b$ as
\begin{equation}
    \mathcal{L}_\mathcal{U} = \frac{1}{\mu B}\sum_{b=1}^{\mu B}{\overline{c}_b \mathcal{D}(\overline{\mathbf{q}}_b,p_\mathrm{model}(y|\mathcal{A}(u_b);\theta))},
    \label{equ:loss_unsup}
\end{equation} 
where the pseudo label $\overline{\mathbf{q}}_b$ is generated by sharpening $\overline{\mathbf{p}}_b$ with temperature scaling $T$ like~\cite{xie2019unsupervised, berthelot2019remixmatch}. A supervised loss for labeled batch is also defined such that
\begin{equation}
    \mathcal{L}_\mathcal{S} = 
    \frac{1}{B}\sum_{b=1}^{B} \mathcal{D}(y_b,p_\mathrm{model}(y|\alpha(x_b);\theta)).
    \label{equ:loss_sup}
\end{equation}
where $y_b$ is the ground-truth label.
Our final loss is as follows: $\mathcal{L}=\mathcal{L}_\mathcal{S}+\lambda\mathcal{L}_\mathcal{U}$ with unsupervised weight parameter $\lambda$. Algorithm~\ref{alg:algorithm} presents the pseudo-code of AggMatch.

\subsection{Relationship with Cognitive Development}
Effectiveness and robustness of AggMatch can be explained through the theory of cognitive development. Vygotsky classified the developmental zone into \textit{the zone of actual development} and \textit{the out of reach zone}, where learners can solve the task on their own and with the help of the more knowledgeable others~\cite{vygotsky1980mind}, respectively. Vygotsky also defined the gap between them as \textit{the zone of proximal development}. The zone of actual development is learned up to the out of reach zone through elaborate scaffolding. When good learning takes place, that is, when the current out of reach zone is changed to the actual zone of development, new out of reach zone and new zone of proximal development are formed. 

Similarly, it can be explained that the label propagation-based methods, such as PAWS~\cite{assran2021semi}, start from the labeled samples and gradually learn the unlabeled data points adjacent to the labeled data points in the feature space corresponding to the zone of proximal development. However, according to Vygotsky's theory of cognitive development, expanding the zone of actual development to the out of reach zone is only possible with the precise guidance of the teacher. For SSL with few labeled data, scaffolding is not sufficient. To solve this problem, many methods form a teacher-student relationship through an asymmetric structure. However, if the teacher is not competent enough, it may not work properly. AggMatch uses a contemplative encoder as a teacher and an impulse encoder as a student according to Kagan's cognitive theory~\cite{kagan1970individual} for more sophisticated scaffolding. The contemplation type encoder is set as a momentum encoder to produce more stable results, and at the same time, the results are thoroughly reviewed through the aggregation module. Therefore, AggMatch can expand the actual zone of development well through more sophisticated scaffold settings.
\begin{algorithm}[t]
\caption{AggMatch}
\label{alg:algorithm}
\hspace*{\algorithmicindent} \textbf{Notation:} confidence threshold $\tau$, weak augmentation $\alpha$, \\
\hspace*{\algorithmicindent}\hspace*{\algorithmicindent}strong augmentation $\mathcal{A}$, queue $\mathcal{Q}$,  model $p_\mathrm{model}$,\\
\hspace*{\algorithmicindent}\hspace*{\algorithmicindent}model parameter $\theta$, momentum model parameter $\theta_m$\\

\hspace*{\algorithmicindent} \textbf{Input:}labeled batch $\mathcal{X}=\{(x_b,y_b)\}_{b=1}^B,$ unlabeled batch \\
\hspace*{\algorithmicindent} \hspace*{\algorithmicindent}$\mathcal{U}=\{u_b\}_{b=1}^{\mu{B}}$

\begin{algorithmic}[1]
\FOR {$b = 1$ to $\mu B$}
    \STATE $\mathbf{v}_{b,w}, \mathbf{p}_{b,w} \gets p_\mathrm{model}(\alpha(\mathit{u}_b);\theta)$
    \STATE $\mathbf{v}_{b,s}, \mathbf{p}_{b,s} \gets p_\mathrm{model}(\mathcal{A}(\mathit{u}_b);\theta)$
    \STATE $\mathbf{v}_{b,w}^{\mathit{m}}, \mathbf{p}_{b,w}^{\mathit{m}} \gets p_\mathrm{model}(\alpha(\mathit{u}_b);\theta_m)$
    \IF {$\max{\mathbf{p}_{b,w}^m} \geq \tau$}
        \STATE $\text{Enqueue}(\mathcal{Q}, (\mathbf{v}_{b,w}^{\mathit{m}}, \mathbf{p}_{b,w}^{\mathit{m}}))$
    \ENDIF
\ENDFOR
\STATE
\FOR {$b = 1$ to $B$}
    \STATE $\mathbf{v}_{b,w}^{m,l}, \mathbf{p}_{b,w}^{m,l} \gets p_\mathrm{model}(\alpha(\mathit{x}_b);\theta_m)$
    \STATE $\mathbf{v}_{b,w}^{l}, \mathbf{p}_{b,w}^{l} \gets p_\mathrm{model}(\alpha(\mathit{x}_b);\theta)$
    \STATE $  (\mathcal{Q}, (\mathbf{v}_{b,w}^{m,l}, \mathbf{p}_{b,w}^{m,l}))$
\ENDFOR
\STATE
\STATE $\mathcal{Q}^1, \mathcal{Q}^2,...,\mathcal{Q}^M \gets \text{RandomSplit}(\mathcal{Q})$ 
\FOR {$b=1$ to $\mu B$}
    \FOR {$m=1$ to $M$}
        \STATE \text{Calculate} $\overline{\mathbf{p}}_b^m$ \text{using~\equref{equ:aggregation_module}}  \\
    \ENDFOR
    \STATE $\overline{c}_b \gets \text{ConfidenceEstimator}(\{\overline{\mathbf{p}}_b^m \mid m=1,...,M\})$
    \STATE $\overline{\mathbf{p}}_b \gets \frac{1}{M} \sum^M_{m=1}{\overline{\mathbf{p}}_b^m}$
    \STATE{$\overline{\mathbf{q}}_b \gets \text{Sharpening}(\overline{\mathbf{p}}_b)$}
\ENDFOR
\STATE \text{Calculate the loss $\mathcal{L}$ via~\equref{equ:loss_unsup} and~\equref{equ:loss_sup}} \\ 
\STATE \text{Update $\theta$ with SGD to minimize $\mathcal{L}$} \\ 
\STATE \text{Update $\theta_m$ with momentum updates using~\equref{equ:momentum}}
\RETURN $\theta$

\end{algorithmic}
\end{algorithm}

\section{Experimental Results}
\label{sec:exp}
\subsection{Experimental Settings}
In this section, we conduct extensive evaluations to justify why our AggMatch can solve the limitations of existing pseudo-labeling, consistency regularization, and label propagation methods; Pseudo-Label~\cite{lee2013pseudo}, pi-Model~\cite{rasmus2015semi}, Mean Teacher~\cite{tarvainen2017mean}, MixMatch~\cite{berthelot2019mixmatch}, UDA~\cite{xie2019unsupervised}, FixMatch~\cite{sohn2020fixmatch}, FeatMatch~\cite{kuo2020featmatch}, and PAWS~\cite{assran2021semi}. We also conduct the experiments with noisy label setting for SSL dataset. To prove the robustness and effectiveness of our methods, we compare our performance to one of the state-of-the-art SSL methods, FixMatch~\cite{sohn2020fixmatch}. We also provide an extensive analysis and ablation studies of key components in our framework. 

\subsubsection{Datasets} 
We use standard benchmarks for SSL, including CIFAR-10/100~\cite{krizhevsky2009learning} and SVHN~\cite{netzer2011reading}. Specifically, CIFAR-10~\cite{krizhevsky2009learning} has 50,000 training images of resolution 32x32 with 10 classes and 10,000 test images. Like CIFAR-10~\cite{krizhevsky2009learning}, CIFAR-100~\cite{krizhevsky2009learning} also has the same number of training/test images and image size but with 100 classes. SVHN~\cite{netzer2011reading} contains close-up of house number images of 32x32, belonging to 10 different classes of digits. The training set has 73,257 images, and the test set contains 26,032 images. In many cases, we perform experiments with fewer labels than previously considered since AggMatch shows competitiveness in extremely label-scarce settings, which is the most difficult setting to generate proper pseudo labels from the biased configuration of few labeled data. 

\subsubsection{Evaluation Metric} 
We compute the mean and standard deviation of the accuracy of the model, trained on three different folds of labeled data, following the standard evaluation protocol of labeling a portion of the training data and leaving the rest unlabeled~\cite{sohn2020fixmatch}.

\subsection{Implementation Details}
Following~\cite{berthelot2019mixmatch, sohn2020fixmatch}, we use the same backbone architecture as WRN-28-2~\cite{zagoruyko2016wide} for CIFAR-10~\cite{krizhevsky2009learning} and SVHN~\cite{netzer2011reading} with a batch size of 64, Wide ResNet-28-8~\cite{zagoruyko2016wide} for CIFAR-100~\cite{krizhevsky2009learning} with a batch size of 48. A hyper-parameter for a unlabeled batch size $\mu$ is set by 7 for both cases. For aggregation, we set temperature by $\tau_{sim}=0.05$, weight for similarity function by $\lambda_{sim}=0.5$, confidence threshold for queue by $\tau=0.95$, and the momentum coefficient for the momentum encoder by $\lambda_m=0.999$, as mentioned by Mean Teacher~\cite{tarvainen2017mean}. Queue size per class $L$ is determined by 2,048 for CIFAR-10~\cite{krizhevsky2009learning} and SVHN~\cite{netzer2011reading}, and 256 for CIFAR-100~\cite{krizhevsky2009learning}. The number of disjoint subset $M$ for estimating uncertainty is determined by 64.
Hyper-parameters for unsupervised loss function, unsupervised loss weight $\lambda $ and sharpening temperature for pseudo-labeling $T$ are set by 1 and 0.5, respectively. To construct the different views, we use two different magnitudes of data augmentation; weak augmentation $\alpha$, consisting of only flip-and-shift augmentation, and strong augmentation $\mathcal{A}$, consisting of randomly selected transformations in RandAugment~\cite{cubuk2020randaugment} as following~\cite{sohn2020fixmatch}. 

\begin{table*}[t]
\caption{\textbf{Quantitative evaluation on CIFAR-10~\cite{krizhevsky2009learning} and SVHN~\cite{netzer2011reading}.} The best results are in bold, and the second best results are underlined. }
\begin{center}
\scalebox{1.0}{
\begin{tabular}{c|ccc|c|cc}
\hlinewd{0.8pt}
\multirow{2}{*}{Methods} & \multicolumn{3}{c|}{CIFAR-10~\cite{krizhevsky2009learning}}  &\multicolumn{1}{c|}{CIFAR-100~\cite{krizhevsky2009learning}}    & \multicolumn{2}{c}{SVHN~\cite{netzer2011reading}}      \\ \cline{2-7} 
                                        & 40                  & 80                 & 250        &400        & 40                  & 250
                                        \\ 
\hline
Pseudo-Label~\cite{lee2013pseudo}                               & -                       & -                  & 51.22±0.43 &- & -                   & 79.79{±1.09}                                \\
pi-Model~\cite{laine2016temporal}                                   & -                   & -                  & 45.74{±3.97} &- & -                   & 81.04{±1.92}                       \\
Mean Teacher~\cite{tarvainen2017mean}                               & -                   & -                  & 67.68{±2.30} &- & -                   & 96.43{±0.11}                      \\
MixMatch~\cite{berthelot2019mixmatch}                              & 51.90{±11.76} & -              & 88.97{±0.85} & 32.39{±1.32} & 57.45{±14.53} & 96.02{±0.26}  \\
UDA~\cite{xie2019unsupervised}                                       & 71.95{±9.64}  & -                  & 91.18{±1.08} & 40.72{±0.88} & 47.37{±20.51} & 94.31{±2.76}   \\
FixMatch (RA)~\cite{sohn2020fixmatch}              & \underline{86.19}{±3.37}  & \underline{92.73}{±0.59} & \textbf{94.93}{±0.65} & 51.15{±1.75} & \textbf{96.04}{±2.17}  & \textbf{97.52}{±0.38}  \\
FeatMatch~\cite{kuo2020featmatch}                 & 58.68{±8.88}  & 83.95{±2.65} & 92.50{±0.64} & - & -                   & \underline{96.66}{±0.19}                                   \\
PAWS~\cite{assran2021semi}                      & 85.74{±2.98}  & 88.72{±0.84}              & 91.18{±0.62} & -                & -                   & -                                                 \\ \hline
\textbf{AggMatch}                                 & \textbf{92.54}{±0.76}       &      \textbf{93.09}{±0.85}              & \underline{94.26}{±0.23} & \textbf{55.58}{±1.57}  & \underline{95.72}{±0.48}               & 95.48{±0.66}                                                        \\ \hlinewd{0.8pt}
\end{tabular}}
\end{center}

\label{quan_table_cifar10_SVHN}
\end{table*}

\begin{table*}[t!]
\begin{center}
\caption{\textbf{Per-class quantitative evaluation on CIFAR-10~\cite{krizhevsky2009learning}.} The best results are in bold, and the lower scores between semantic labels (airplane $\leftrightarrow$ bird, dog $\leftrightarrow$ cat) are underlined respectively.}
\begin{tabular}{c|cccccccccc|c}
\hlinewd{0.8pt}
Methods                        & air. & auto.               & bird                     & cat  & deer & dog  & frog  & horse & ship & truck & total \\ \hline
\multicolumn{1}{c|}{FixMatch~\cite{sohn2020fixmatch}} & \textbf{94.3}     & \multicolumn{1}{c}{\textbf{97.6}} & \multicolumn{1}{c}{\underline{72.6}} & \underline{28.1} & \textbf{96.6} & \textbf{94.1}& 98.1 &    95.6   &  97.5    &     96.6 &87.11 \\ \hline
\textbf{AggMatch}                      & 92.4     & 96.2                     & \textbf{\underline{84.3}}                     & \textbf{\underline{82.7}} & 96.2 & 88.3 & \textbf{98.6} & \textbf{95.8}  & \textbf{97.7}& \textbf{97.9} & \textbf{93.01} \\ 
\hlinewd{0.8pt}
\end{tabular}    
\label{Table:per_class}
\end{center}
\end{table*}

\begin{table}[t]
\caption{\textbf{Evaluation of robustness of AggMatch on CIFAR-10~\cite{krizhevsky2009learning} and SVHN~\cite{netzer2011reading} with noisy settings under various label fractions, 40 and 250 labels.}}
\begin{center}
\small
\scalebox{0.85}{
\begin{tabular}{c|c|c|c|c}
\hlinewd{0.8pt}
\multirow{2}{*}{Methods} & \multicolumn{2}{c|}{CIFAR-10~\cite{krizhevsky2009learning}} & \multicolumn{2}{c}{SVHN~\cite{netzer2011reading}} \\ \cline{2-5} 
 &
  \begin{tabular}[c]{@{}c@{}}40 labels\\ +Noise 25\%\end{tabular} &
  \begin{tabular}[c]{@{}c@{}}250 labels\\ +Noise 50\%\end{tabular} &
  \begin{tabular}[c]{@{}c@{}}40 labels\\ +Noise 25\%\end{tabular} &
  \begin{tabular}[c]{@{}c@{}}250 labels\\ +Noise 50\%\end{tabular} \\ \hline
FixMatch~\cite{sohn2020fixmatch}
&72.18{$\pm$10.95}  
&79.68{$\pm$4.84} 
&\textbf{81.39{$\pm$3.35}} 
& 75.12{$\pm$2.04}   \\ \hline
\textbf{AggMatch}
&\textbf{77.63{$\pm$12.29}}
& \textbf{85.41{$\pm$7.07}} 
&78.66{$\pm$5.82}  
&\textbf{89.47{$\pm$5.54}}   \\ 
\hlinewd{0.8pt}
\end{tabular}
}
\end{center}

\label{quan_noisy_table_cifar10}
\end{table}

\subsection{Experimental Results}
\subsubsection{Results on Standard SSL Benchmarks.}
As shown in Table~\ref{quan_table_cifar10_SVHN}, we conduct experiments on CIFAR-10/100~\cite{krizhevsky2009learning}, and SVHN~\cite{netzer2011reading} focusing on the label-scarce settings. We use the same backbone architecture for a fair comparison to other methods. We show the state-of-the-art accuracy of 92.54\%, outperforming FixMatch~\cite{sohn2020fixmatch} by 6.35\% on CIFAR-10~\cite{krizhevsky2009learning} with four labeled samples per class and also record 93.09\% on CIFAR-10~\cite{krizhevsky2009learning} with eight labeled samples per class. We obtain 55.58\% on CIFAR-100~\cite{krizhevsky2009learning} with four labeled per class setting, which is 4.43\% higher than a previous state-of-the-art method. We also show comparable results on SVHN~\cite{netzer2011reading} dataset by recording 95.72\%, the second-best result on SVHN~\cite{netzer2011reading} with four labeled per class. 

Compared to the results of FeatMatch~\cite{french2017self} and PAWS~\cite{assran2021semi} closely related to ours, which utilize the additional aggregation module, we can prove the robustness and competitiveness of our method. Rather, both of them show lower performance than FixMatch~\cite{sohn2020fixmatch} on CIFAR-10~\cite{krizhevsky2009learning} with 40/80/250 settings and their accuracy rates declines dramatically as the number of labeled data become scarce, specifically from 250 to 40. It can be interpreted that their attention modules significantly rely on the quality and quantity of labeled data. This is because PAWS leverages only a mini-batch of labeled samples for measuring similarity between unlabeled samples, which will lead to bias from the configuration of selected labeled data. FeatMatch~\cite{kuo2020featmatch} utilizes information from both within-class and across-class prototypical representations through clustering in the memory bank for feature-based refinement and augmentation. But the result mentioned above proves that incorrect prototypes extracted from the memory bank, including samples without considering confidence, deteriorate the quality of features. Unlike them, AggMatch shows the state-of-the-art performance in label-scare settings such as CIFAR-10~\cite{krizhevsky2009learning} with 40 labels and 80 labels. We also outperform PAWS and FeatMatch by 6.8\% and 33.9\% at 40 labels setting, respectively. Based on the above results, our method shows powerful competitiveness in label-scarce settings by using aggregation with lots of consistent and confident samples in the queue.


More specifically, We analyze our per-class accuracy on CIFAR-10~\cite{krizhevsky2009learning} with 40 labels, compared to FixMatch~\cite{sohn2020fixmatch} in~\tabref{Table:per_class}. total accuracy of FixMatch~\cite{sohn2020fixmatch} is 87.11\%, but it can be seen that there is a significant difference in accuracy between pairs of semantic labels, which can share the abstract feature space due to the similar appearance between the two classes. Fixmatch~\cite{sohn2020fixmatch} shows that the accuracy between pairs of semantic labels are largely biased on one side of label. For example, the cat label and the bird label have lower accuracy by 66.0\% and 21.7\% compared to their counterparts, the dog label and the 
airplane label, respectively. On the contrary, Aggmatch shows that the accuracy between pairs of semantic labels is quite equivalent to the other. From this observation, we prove that samples, having ambiguous features and class probabilities, can be corrected by aggregating numerous discriminative within-class and across-class samples in the confident-aware queue. And it also proves that the confidence estimation can appropriately measure the confidence of the pseudo-labeling using multiple hypotheses with subsets of the queue.


    

\subsubsection{Results with Noisy Labels} 
In this section, we conduct experiments on CIFAR-10~\cite{krizhevsky2009learning} and SVHN~\cite{netzer2011reading} with noisy label settings to prove that AggMatch performs well in another confirmation bias setting caused by the quality of the label. Although training with noisy labels in supervised learning has been actively discussed~\cite{gu2021realistic}, SSL with noisy labels has not been explored yet. Noisy label settings hinder the model from learning proper representation, closely related to \textit{In-the-wild datasets}, having poor annotation qualities or ambiguous class labels without refining significant noise. We conduct the experiment on SSL benchmarks, CIFAR-10~\cite{krizhevsky2009learning} and SVHN~\cite{netzer2011reading}, with asymmetric label noise to explore the robustness of our approach. The asymmetric label noise mimics practical annotation error in semantically-similar classes, i.e., cats and dogs. Following previous work~\cite{patrini2017making}, we propose asymmetric mapping of the noisy labels for CIFAR-10~\cite{krizhevsky2009learning}: TRUCK $\to$ AUTOMOBILE, BIRD $\to$ AIRPLANE, DEER $\to$ HORSE, CAT $\leftrightarrow$ DOG and for MNIST~\cite{deng2012mnist}: $2 \to 7$, $3 \to 8$, $5 \leftrightarrow 6$, $7 \to 1$. Based on the similarity of the numeric representation, we apply the same asymmetric noise mapping as MNIST~\cite{deng2012mnist} to SVHN~\cite{netzer2011reading} for generating noisy labels.

As shown in Table~\ref{quan_noisy_table_cifar10}, AggMatch outperforms FixMatch~\cite{sohn2020fixmatch} on CIFAR-10~\cite{krizhevsky2009learning} and SVHN~\cite{netzer2011reading} by a large margin. More specifically, AggMatch shows the most significant improvement on CIFAR-10~\cite{krizhevsky2009learning} with 50\% noise on 250 labels setting compared to FixMatch~\cite{sohn2020fixmatch} by 5.73\%. Also, in SVHN~\cite{netzer2011reading}, consisting of more difficult real-world images, AggMatch outperforms FixMatch~\cite{sohn2020fixmatch} by 14.35\%. Even in a more challenging setting in which one of the four labels per class is noise, AggMatch shows 5.5\% performance improvement compared to FixMatch~\cite{sohn2020fixmatch}. Even if there are continuous noise labels, we can build a robust model by maintaining the high quality of pseudo labels through aggregation.


\begin{figure}[t]
    \centering
	\subfloat[Precision]
	{{\includegraphics[width=0.5\linewidth]{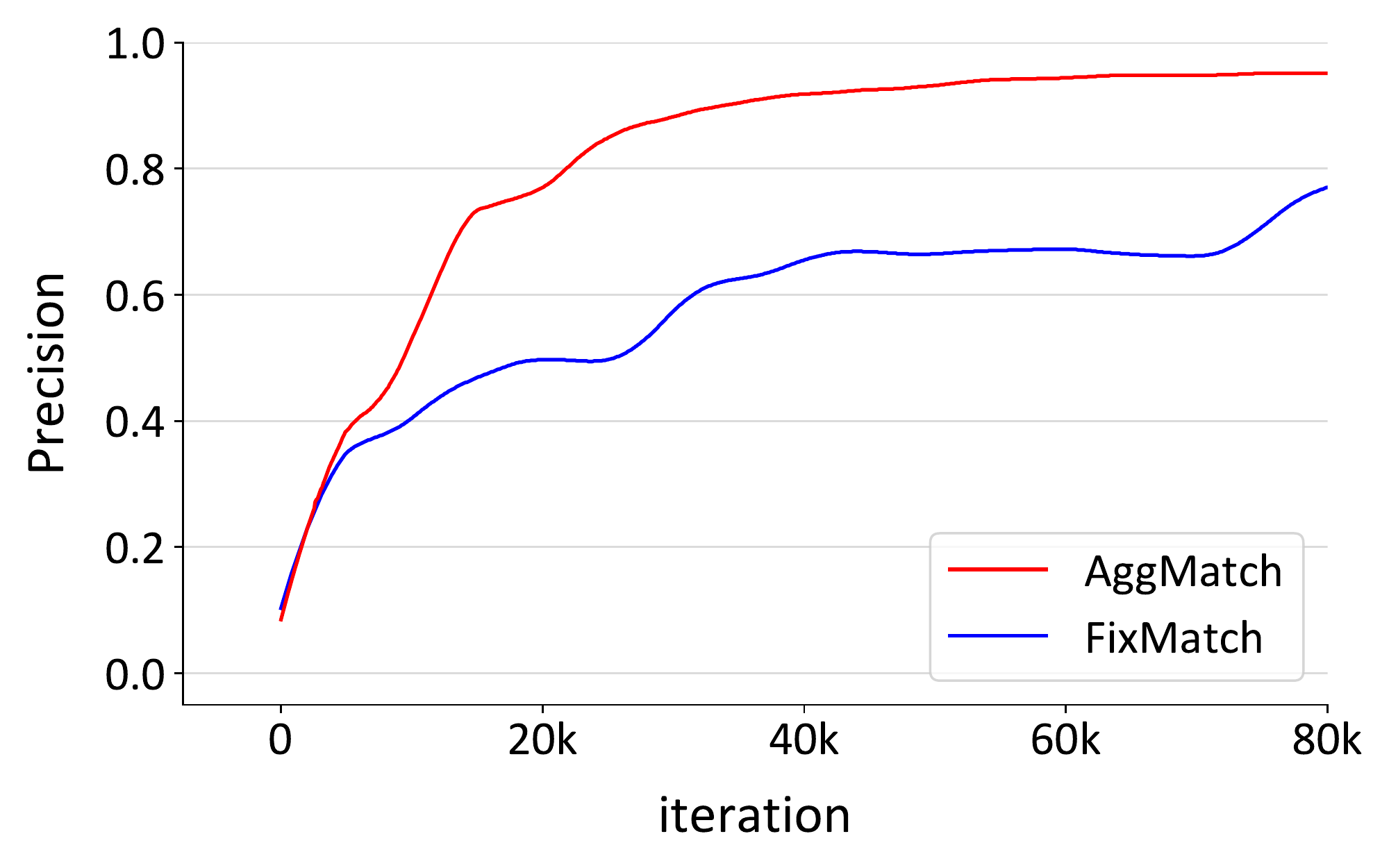}}}\hfill
	\subfloat[Precision on noise setting]
 	{\includegraphics[width=0.5\linewidth]{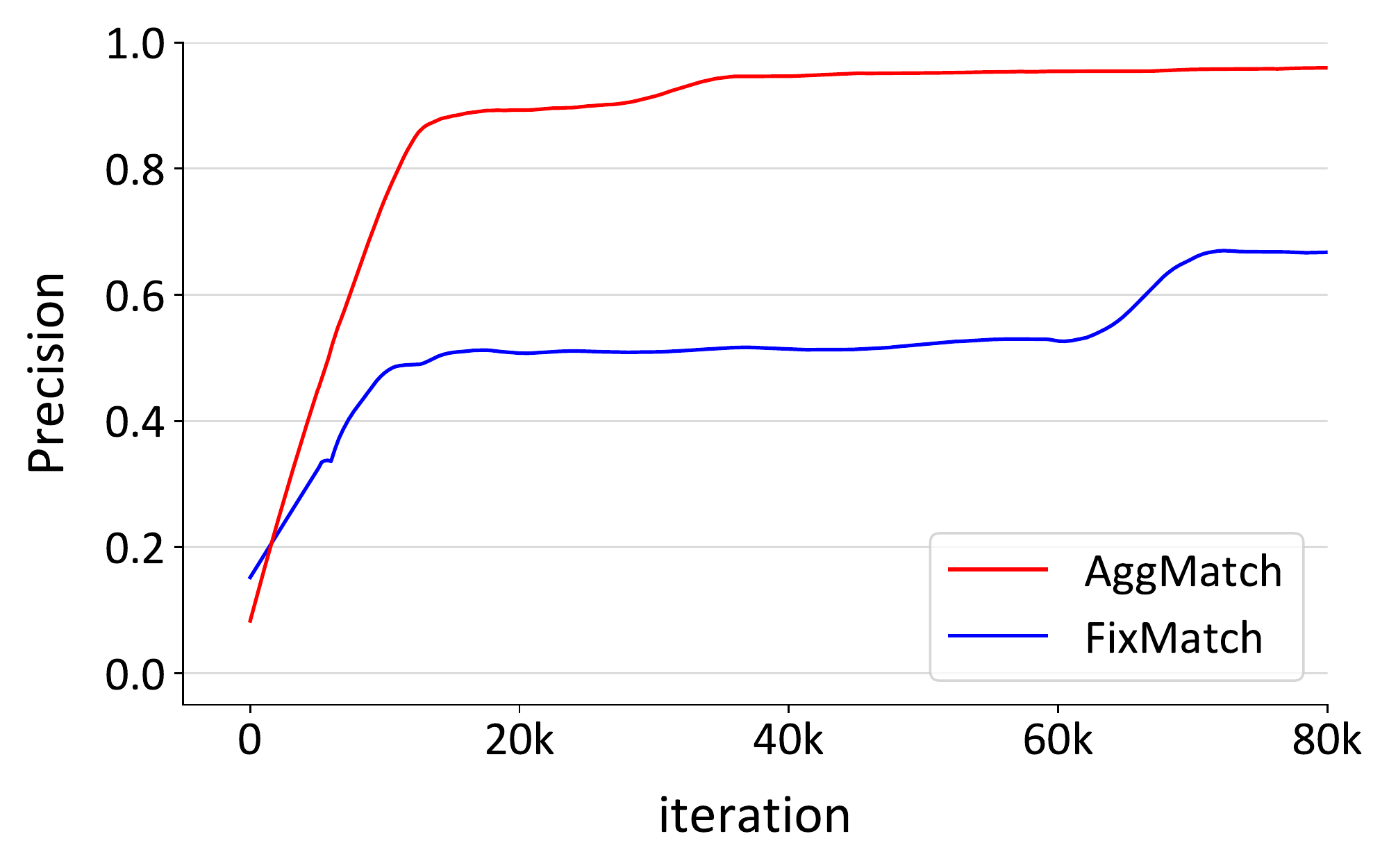}}\hfill \\
 	\subfloat[Recall on noise setting]
 	{\includegraphics[width=0.5\linewidth]{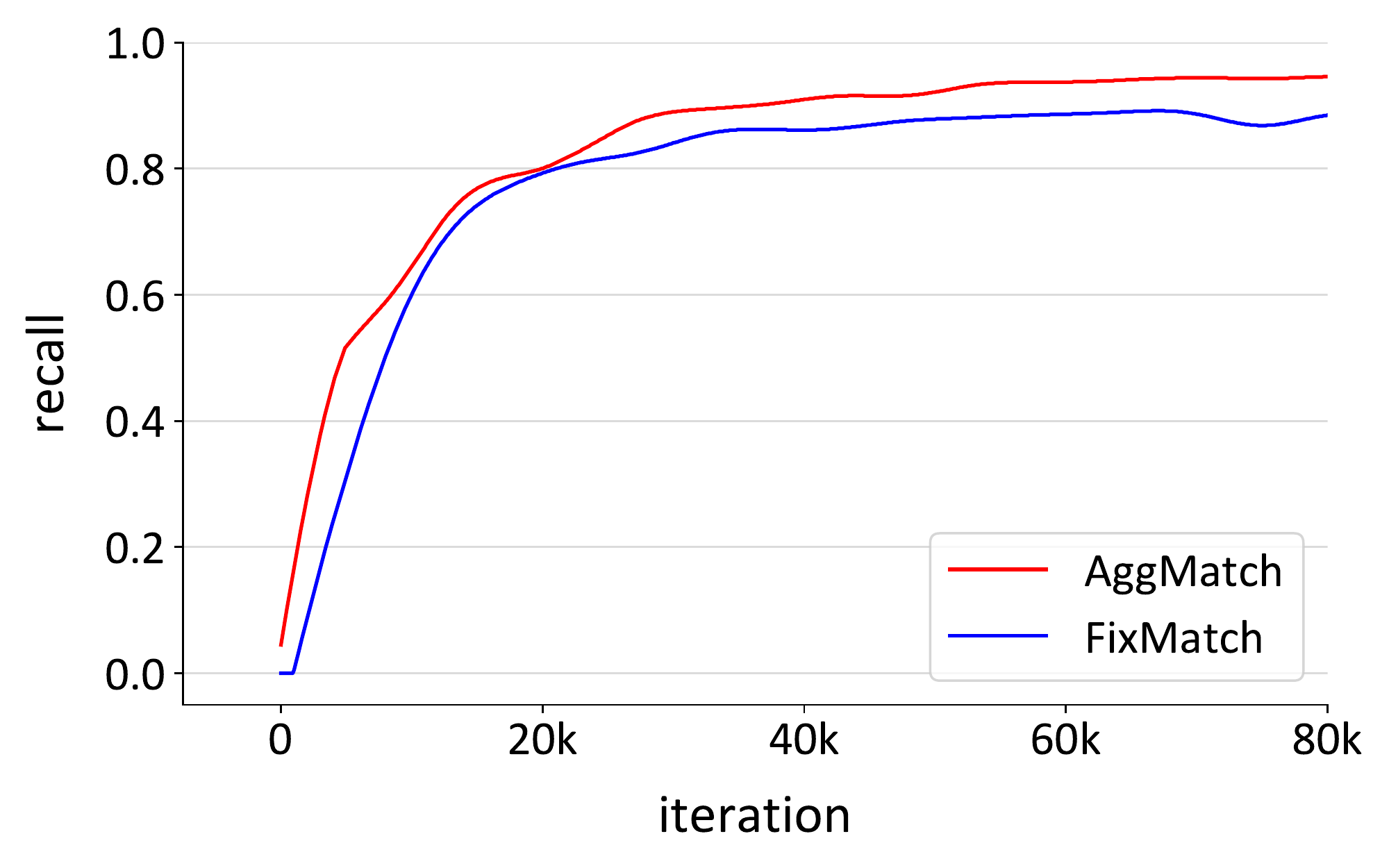}}\hfill
 	\subfloat[Ours]
 	{\includegraphics[width=0.5\linewidth]{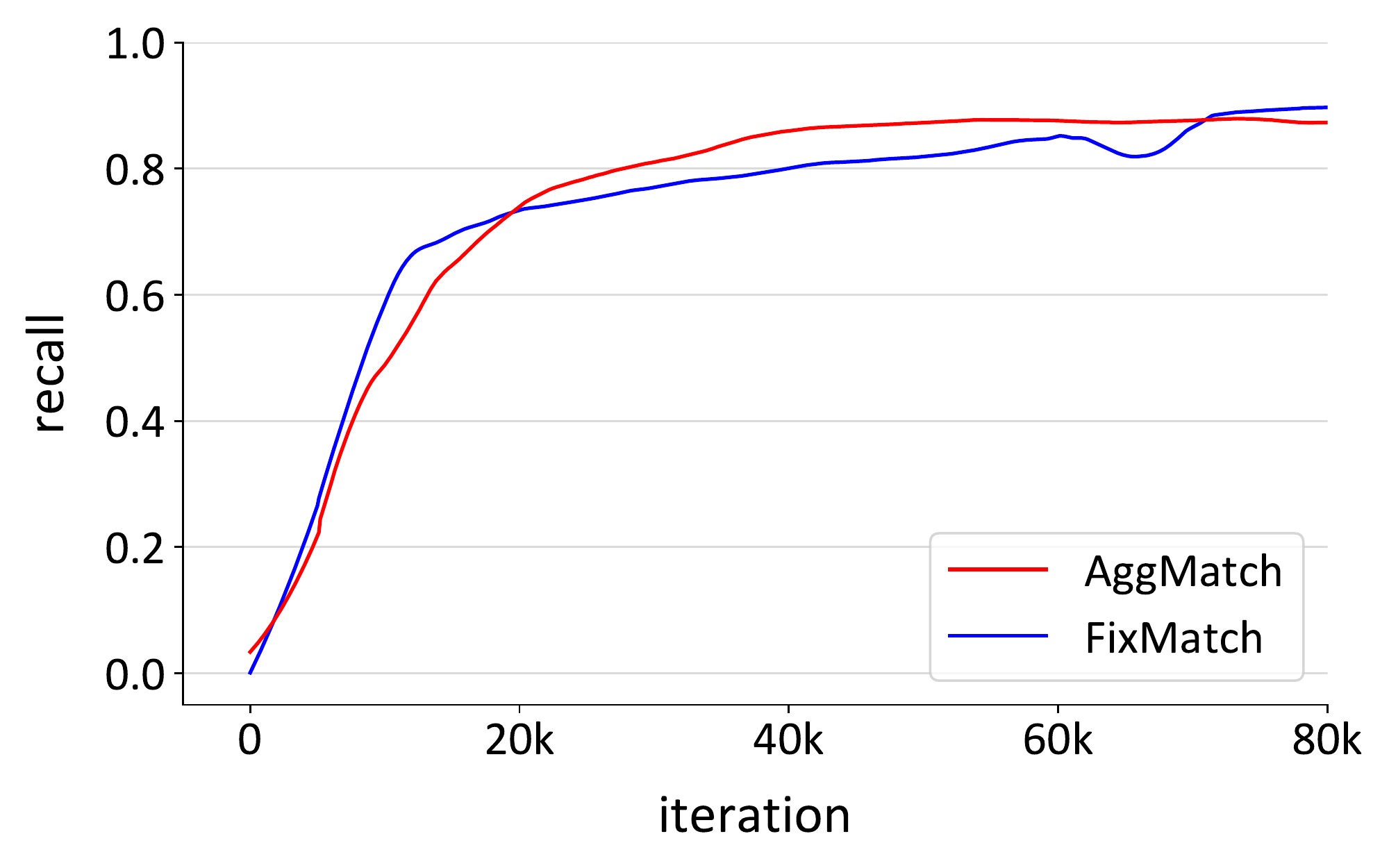}}\hfill \\
    
	\caption{\textbf{Plots of precision and recall:} (a) and (c) are conducted on CIFAR-10~\cite{krizhevsky2009learning} with 40 labels. (b) and (d) are conducted on CIFAR-10~\cite{krizhevsky2009learning} with 25\% noise on 40 labels.}
	\label{fig:uncertainty}
\end{figure}



\subsubsection{Evaluating Confidence Estimation}
\figref{fig:uncertainty} shows the recall and precision evaluation on AggMatch and FixMatch~\cite{sohn2020fixmatch} across the training iterations on CIFAR-10~\cite{krizhevsky2009learning} with 40 labels and also CIFAR-10~\cite{krizhevsky2009learning} with 25\% noise on 40 labels. 
In SSL, it is important to generate accurate pseudo labels because confirmation bias from incorrect pseudo-labeling is the main reason for performance limitation. For this, FixMatch~\cite{sohn2020fixmatch} uses a fixed threshold strategy, using only unlabeled data whose confidence is above the threshold, but this naïve thresholding strategy only involves very high-quality unlabeled data and cannot consider the learning status of the network and the difficulties of different samples. On the other hand, we dynamically adjust the magnitude of unsupervised loss function based on measuring uncertainty by changing the confidence-based strategy from thresholding to weighting. 

The results of precision, defined as the number of the correct pseudo labels over the pseudo labels of the samples in SSL classification, are shown in ~\figref{fig:uncertainty}(a). ~\figref{fig:uncertainty}(a) and  ~\figref{fig:uncertainty}(b) show the significant margin between FixMatch~\cite{sohn2020fixmatch} and AggMatch on CIFAR-10~\cite{krizhevsky2009learning} with 40 labels as gradually formulating queue with the confident-aware samples. 
Also, recall, defined as the number of the correct pseudo labels over the ground-truth labels of the samples in SSL classification, is a trade-off with precision, but AggMatch shows similar or higher recall despite the overwhelming precision as shown in ~\figref{fig:uncertainty}(c) and ~\figref{fig:uncertainty}(d). It can be interpreted that the aggregation is effective for refining label refinement and measuring uncertainty and that weighting strategy can help accelerate training convergence by securing lots of confident pseudo labels.

\subsection{Ablation Study}
\label{sec:ablation}
We also conduct ablation study to evaluate the performance gap of components our overall framework, including more extensive analyses on similarity function and uncertainty estimation. We also conduct experiments on queue size and momentum encoder for validating our method. The dataset used for the experiments is CIFAR-10~\cite{krizhevsky2009learning}, and the number of labeled data for training is limited to four samples per class. 

\subsubsection{Effectiveness of Key Components}
We analyze the key components of the pseudo label aggregation module in AggMatch, one is an pseudo label refinement and the other is a confidence estimation in~\tabref{Tab:ab_key}. We define the baseline as the model without two modules. The two modules were added alternately on the baseline to analyze the effectiveness of each module. The improvement of the aggregation module and confidence estimation module is achieved by 2.92\% / 2.62\% over the baseline, respectively. It proves that each module shows adequate improvement and equally contributes to the performance. The best performance is recorded as 92.54\% with both modules, outperforming the previous state-of-the-art, FixMatch~\cite{sohn2020fixmatch}, by 6.35\%. 

\begin{table}[t]
\caption{\textbf{Ablation study on key components in AggMatch such as pseudo labeling refinement and confidence estimation module.}}
\small
\begin{center}
\begin{tabular}{c|c|c}
\hlinewd{0.8pt}
\multicolumn{2}{c|}{Components}                  & \multirow{1}{*}[-0.4cm]{Accuracy} \\ \cline{1-2}
\shortstack[c]{ \\Pseudo label\\refinement}     & \shortstack[c]{ \\Confidence\\estimation} &  \\ \hline
\multicolumn{1}{c|}{\xmark}   & \xmark                      & 87.64{±2.19}         \\ \hline
\multicolumn{1}{c|}{\cmark} & \xmark                        & 90.56{±2.63}         \\ \hline
\multicolumn{1}{c|}{\xmark } & \multicolumn{1}{c|}{\cmark} & \multicolumn{1}{c}{90.26{±1.93}}     \\ \hline
\multicolumn{1}{c|}{\cmark} & \multicolumn{1}{c|}{\cmark} & \multicolumn{1}{c}{\textbf{92.54}{±0.76}}    \\
\hlinewd{0.8pt}
\end{tabular}
\end{center}

\label{Tab:ab_key}
\end{table}

\subsubsection{Analysis on Each Term of Similarity Function}
In this section, we analyze the similarity function in the aggregation module $\mathcal{S}$.
Unlike previous methods~\cite{cao2019gcnet, kuo2020featmatch, assran2021semi}, we define it as the summation of \textit{feature similarity} term and \textit{class similarity} term based on cosine similarity and Jenson-Shannon divergence on feature vector $\mathbf{v}$ and class distribution $\mathbf{p}$, respectively. Through the ablation study for similarity function in~\tabref{Tab:ab_sim_1}, there is about 3\% difference in accuracy depending on whether ~\textit{class similarity} term is included in the similarity function.

We also conduct a detailed analysis by visualizing the attention weights depending on the similarity function formulation of aggregation. Note that the principal diagonal of attention weights means the similarity magnitude of within-class feature vectors, and other regions mean those of across-class feature vectors. Each row in \tabref{Tab:ab_sim_2} shows the transitions of the attention weights between learned feature vectors based on \textit{feature similarity} term, \textit{class similarity} term and combination of \textit{feature similarity} term and \textit{class similarity} term, respectively as the training of AggMatch progresses. As the training progresses, regardless of the composition of the similarity function, we can confirm that the gap between within-class similarity and across-class similarity is bigger. This means our aggregation is effective for generating discriminative feature vectors and class distributions by optimizing the model, consisting of feature extractor and classifier. When comparing the attention weights in the best model parameters, the diagonal components of attention weight based on the summation of both terms show high contrast with other regions. 
By integrating the results from~\tabref{Tab:ab_sim_1} and~\tabref{Tab:ab_sim_2}, it can be interpreted that aggregating pseudo labels of ambiguous samples, which cannot be refined by \textit{feature similarity} term alone, is complemented by ~\textit{class similarity} term.

\subsubsection{Comparison of Thresholding and Weighting for Confidence Estimation}
Unlike standard confidence-based algorithms~\cite{xie2019unsupervised, sohn2020fixmatch}, we measure confidence as uncertainty-based~\textit{weighting}. More specifically, we can scale the loss magnitude by \textit{weighting} uncertainty estimation based on the consensus between hypotheses from different subsets of the queue. To demonstrate the effectiveness of \textit{weighting}, we compare \textit{thresholding} and \textit{weighting} on uncertainty approach. We can define uncertainty-based \textit{thresholding} such as
$\frac{1}{\mu B}\sum_{b=1}^{\mu B}{\mathbbm{1}(\overline{i}_b\leq\tau_u) \mathcal{D}(\overline{\mathbf{q}}_b,p_\mathrm{model}(y|\mathcal{A}(u_b);\theta))},$ where $\overline{i}_b=-\sum\mathbf{a}_b\log{\mathbf{a}_b}$. The pre-defined threshold value $\tau_{u}$ is set in a range from 0.1 to 1.5 where the lower threshold value uses relatively higher entropy samples for training compared to the higher threshold. 

The result of \textit{weighting} method outperforms ~\textit{thresholding} method by 5.68\%/1.72\%/2.65\%/3.72\%, respectively as shown in~\tabref{Tab:ab_uc}. This is because it is difficult to manipulate the quantity and quality of the samples by using only one pre-defined threshold.
Compared to $\tau_{u}=1.0$, which is an optimal threshold value in the experiment, there is 2.54\% performance decline in $\tau_{u}=1.5$ due to the quality problem, generated by incorrect pseudo labels from low confident samples. On the other hand, we can also prove that quantity problem in $\tau_{u}=0.1$, caused by the few number of high confident samples through 4.26\% performance gap.

\begin{table}[t]
\small
\centering
\caption{\textbf{Ablation on each term of similarity function of the aggregation module.}}
\begin{tabular}{c|c|c}
\hlinewd{0.8pt}
\multicolumn{2}{c|}{Similarity function} & \multirow{1}{*}[-0.4cm]{Accuracy} \\ \cline{1-2}
\shortstack[c]{ \\Feature\\similarity term}     & \shortstack[c]{ \\Class\\similarity term} &  \\ \hline
\cmark                & \xmark                  & 89.58{±5.02}                        \\
\cmark                & \cmark              & \textbf{92.54}{±0.76}             \\ 
\hlinewd{0.8pt}
\end{tabular}%
\label{Tab:ab_sim_1}
\end{table}

\begin{table*}[t] 
\caption{\textbf{Visualization on the transitions of attention weights as training iteration progress depending on the similarity function configuration in aggregation on CIFAR-10~\cite{krizhevsky2009learning}.}}
\begin{center}
\small
\scalebox{0.88}{
\begin{tabular}{c|ccc}
\hlinewd{0.8pt}

\multirow{2}{*}{Similarity function} & \multirow{2}{*}{1000 iter} & \multirow{2}{*}{10000 iter} & \multirow{2}{*}{\textbf{Best}} \\ & & & \\ \hline

Feature similarity term & {\includegraphics[align=c,width=0.3\linewidth]{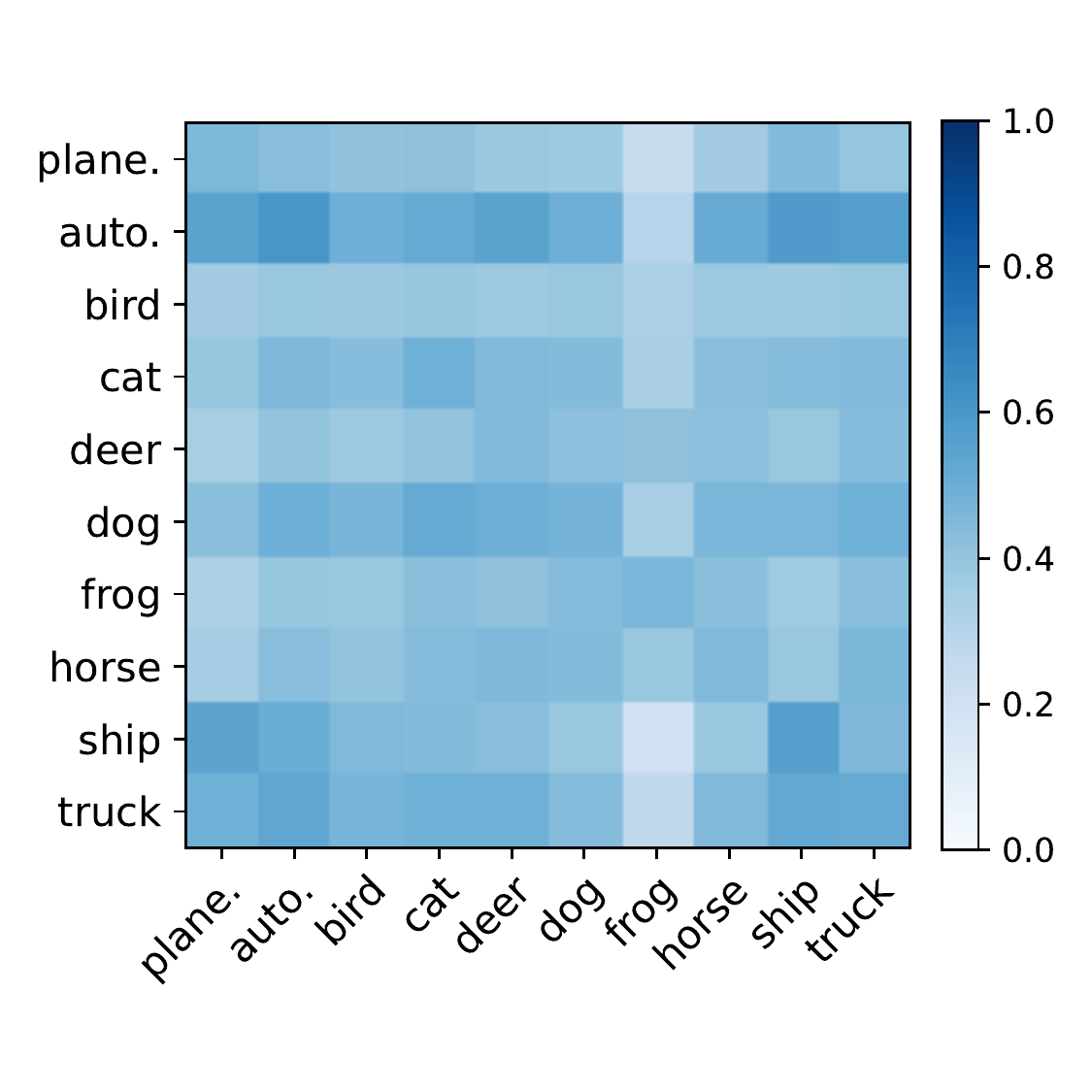}}\hfill &                     {\includegraphics[align=c,width=0.3\linewidth]{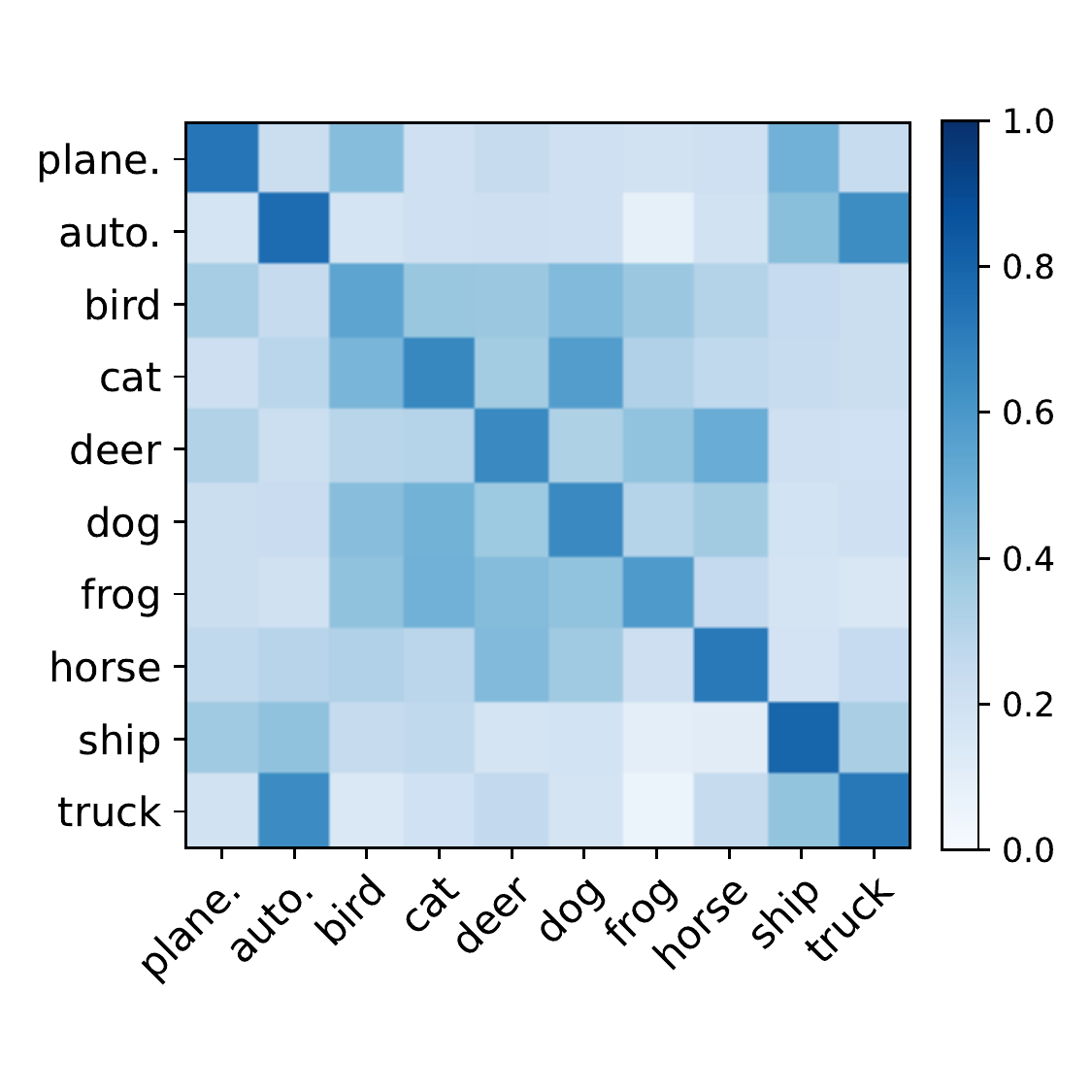}}\hfill & {\includegraphics[align=c,width=0.3\linewidth]{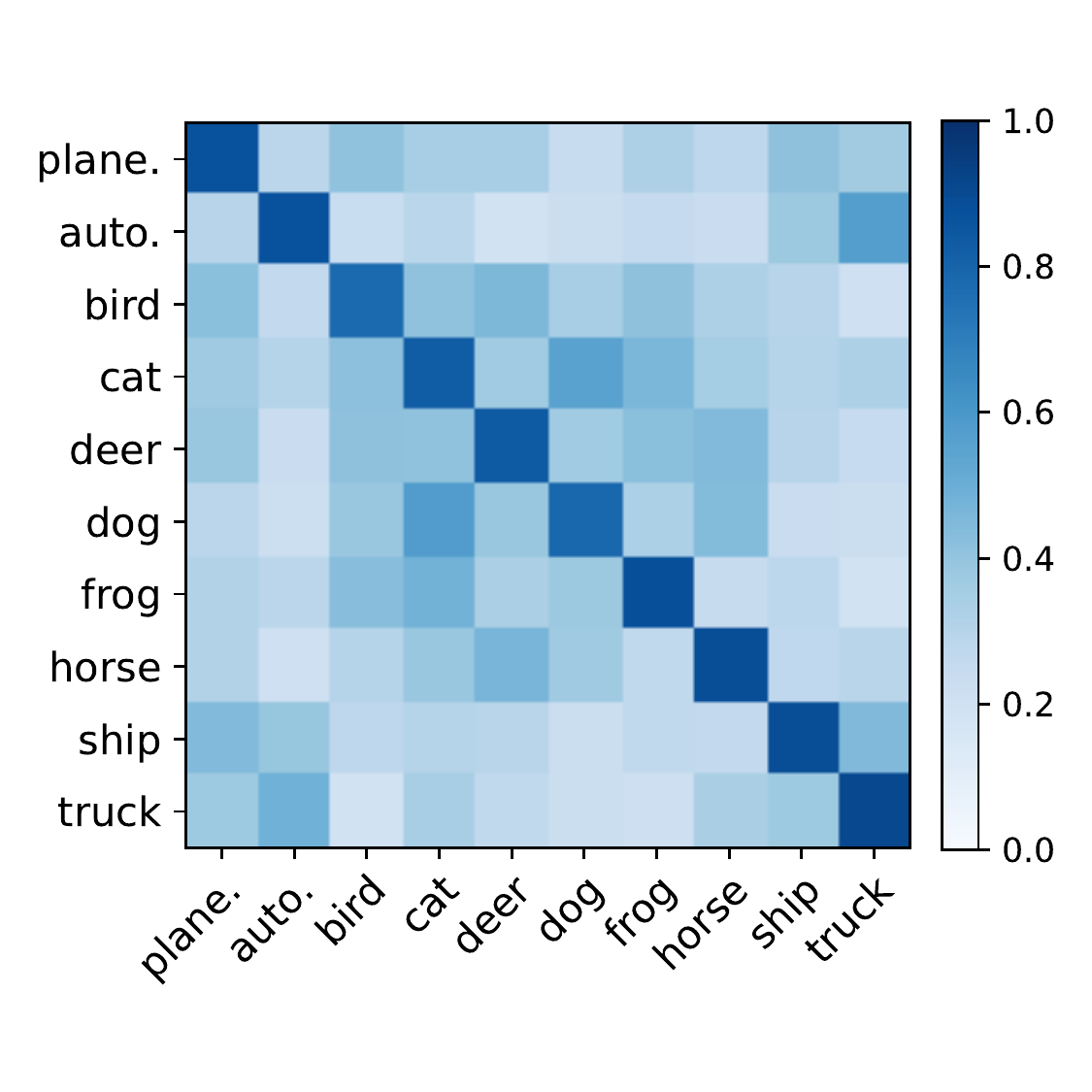}}\hfill  \\ \hline
Class similarity term & {\includegraphics[align=c,width=0.3\linewidth]{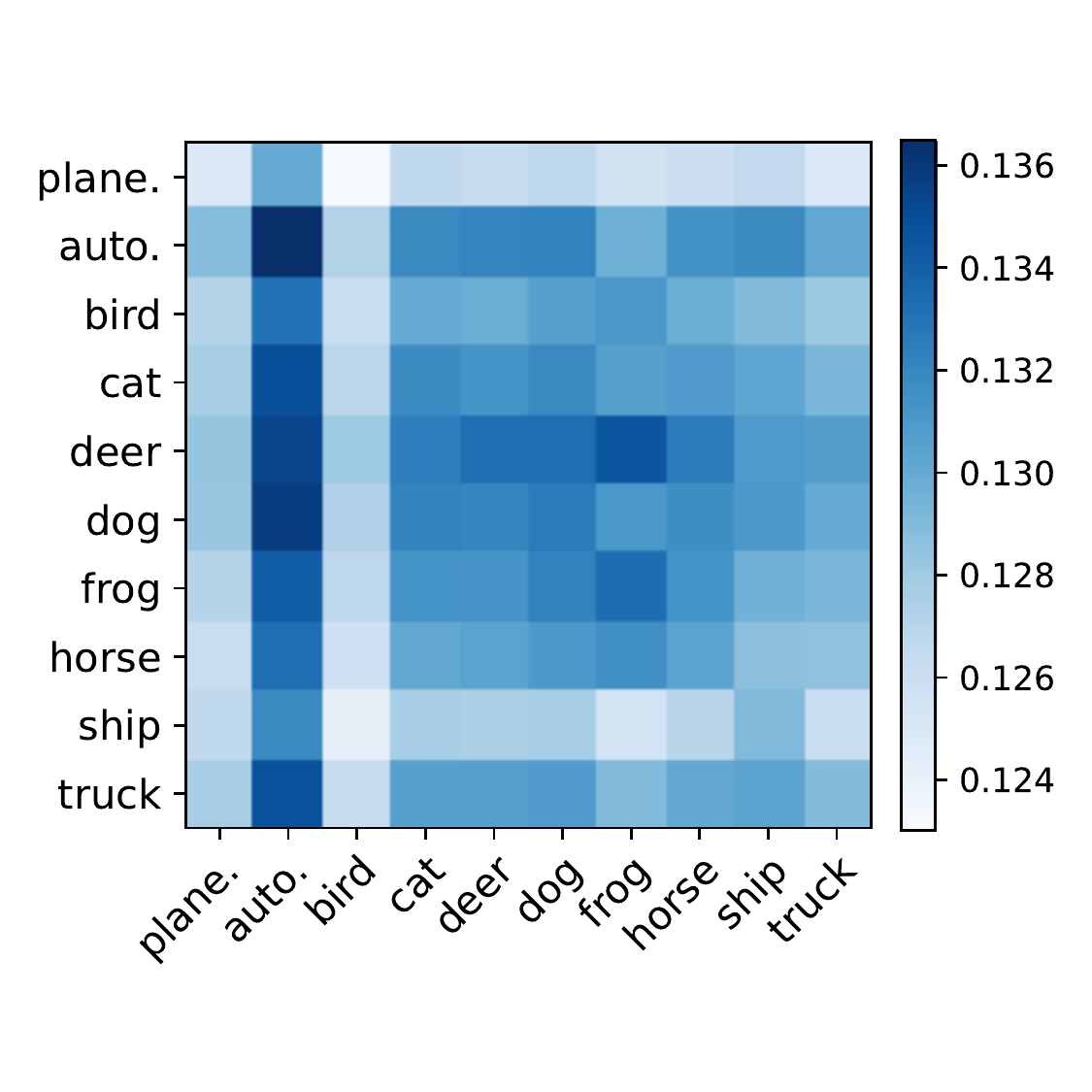}}\hfill &                     {\includegraphics[align=c,width=0.3\linewidth]{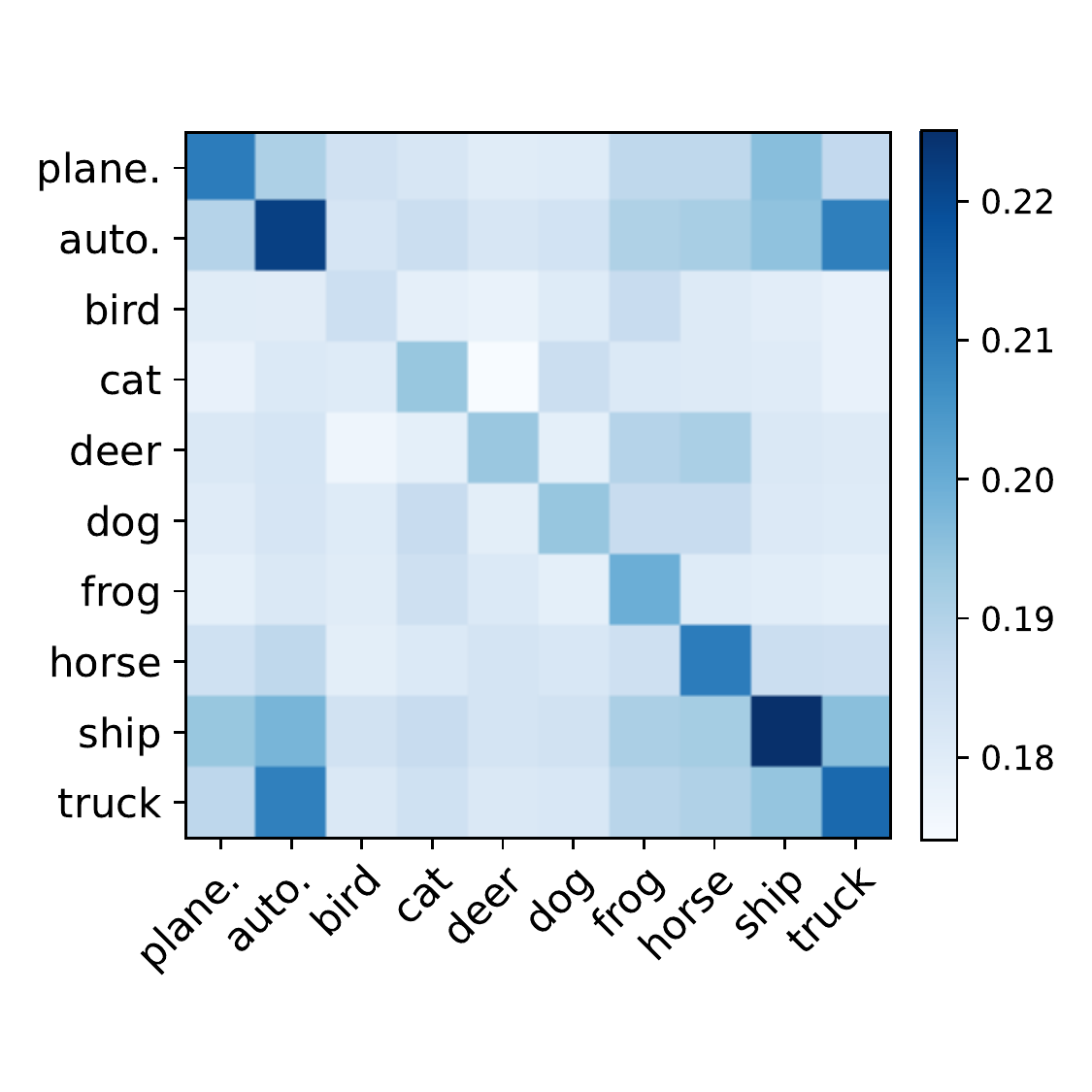}}\hfill & {\includegraphics[align=c,width=0.3\linewidth]{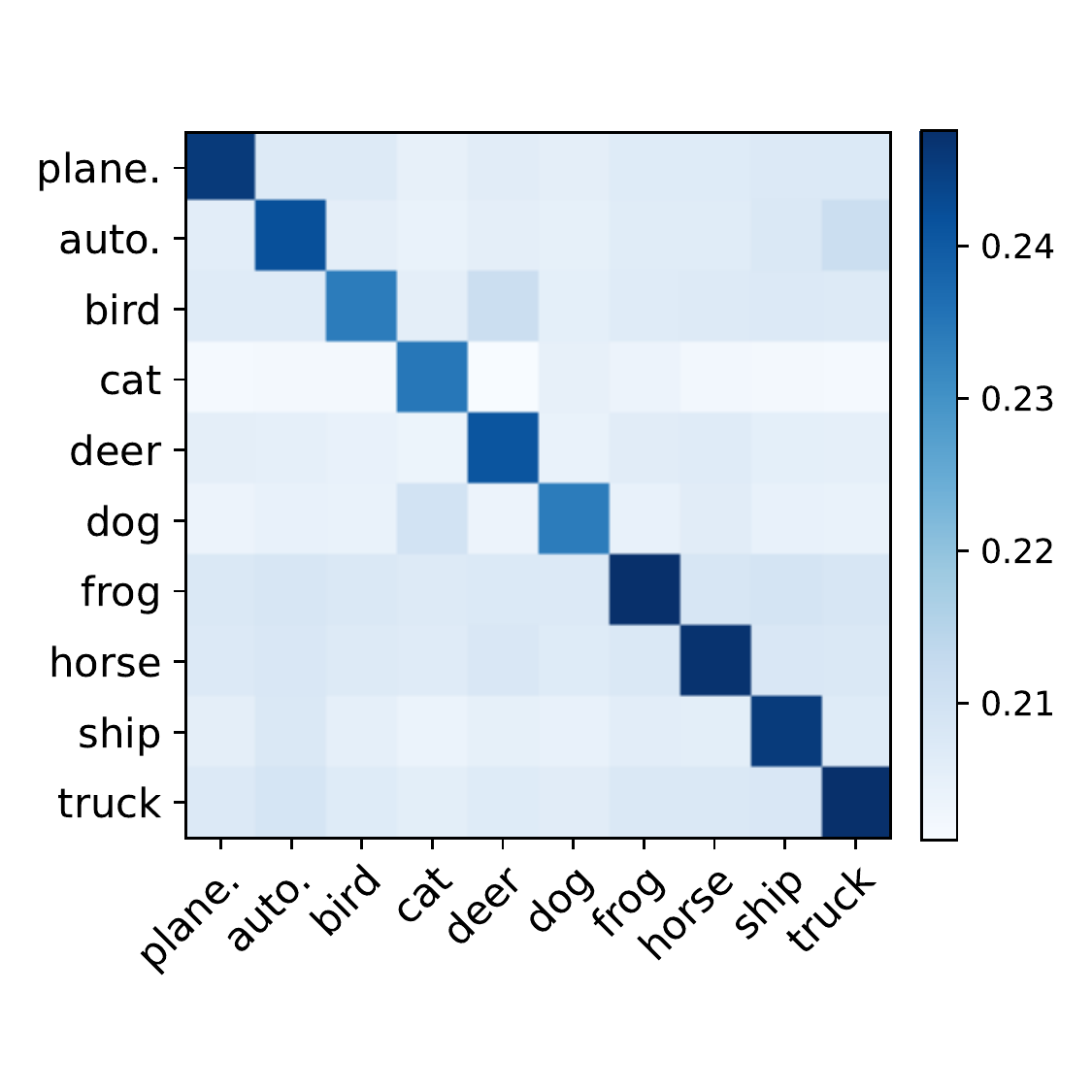}}\hfill  \\ \hline
\begin{tabular}[c]{@{}c@{}}Feature similarity term \\+\\ Class similarity term\end{tabular} & {\includegraphics[align=c,width=0.3\linewidth]{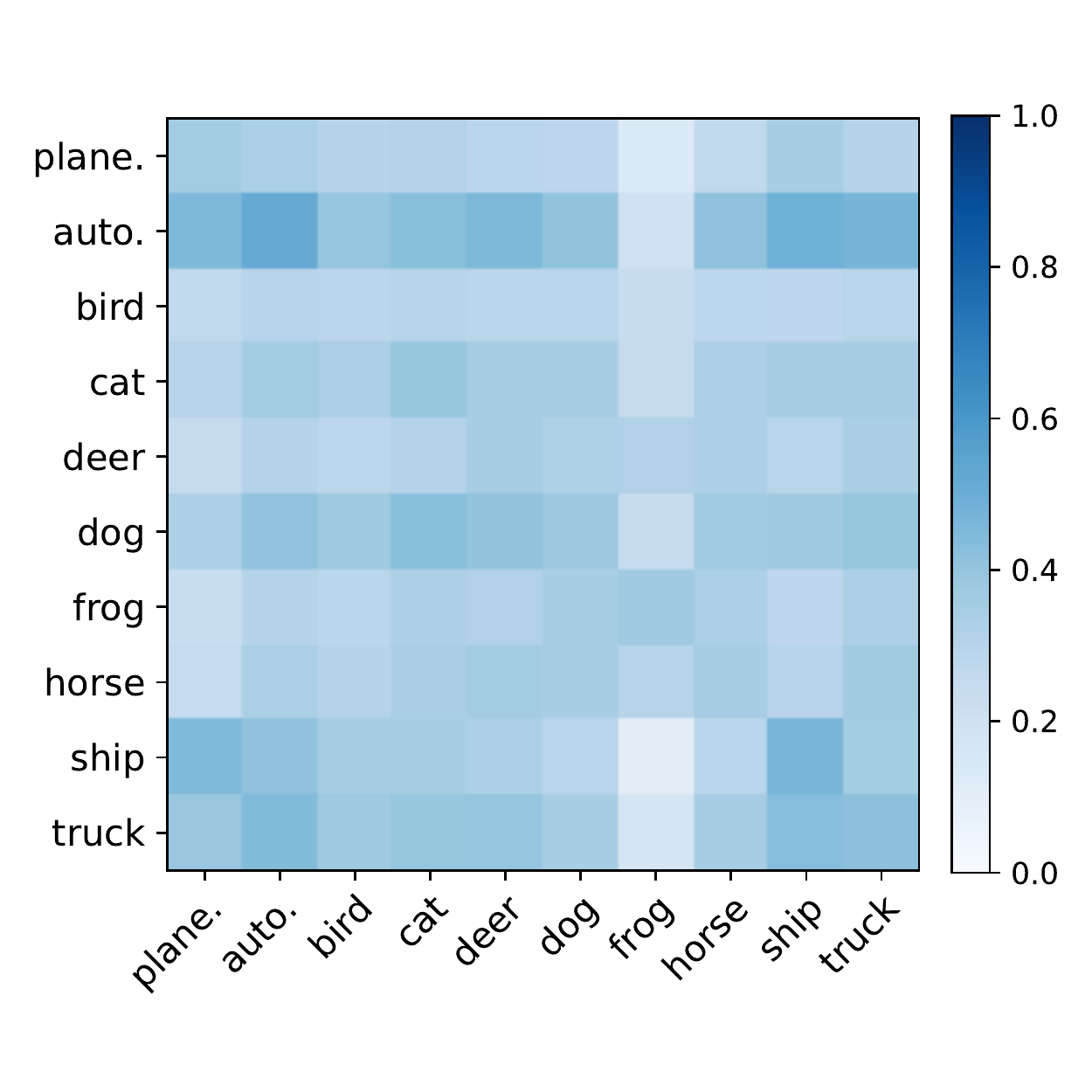}}\hfill &                     {\includegraphics[align=c,width=0.3\linewidth]{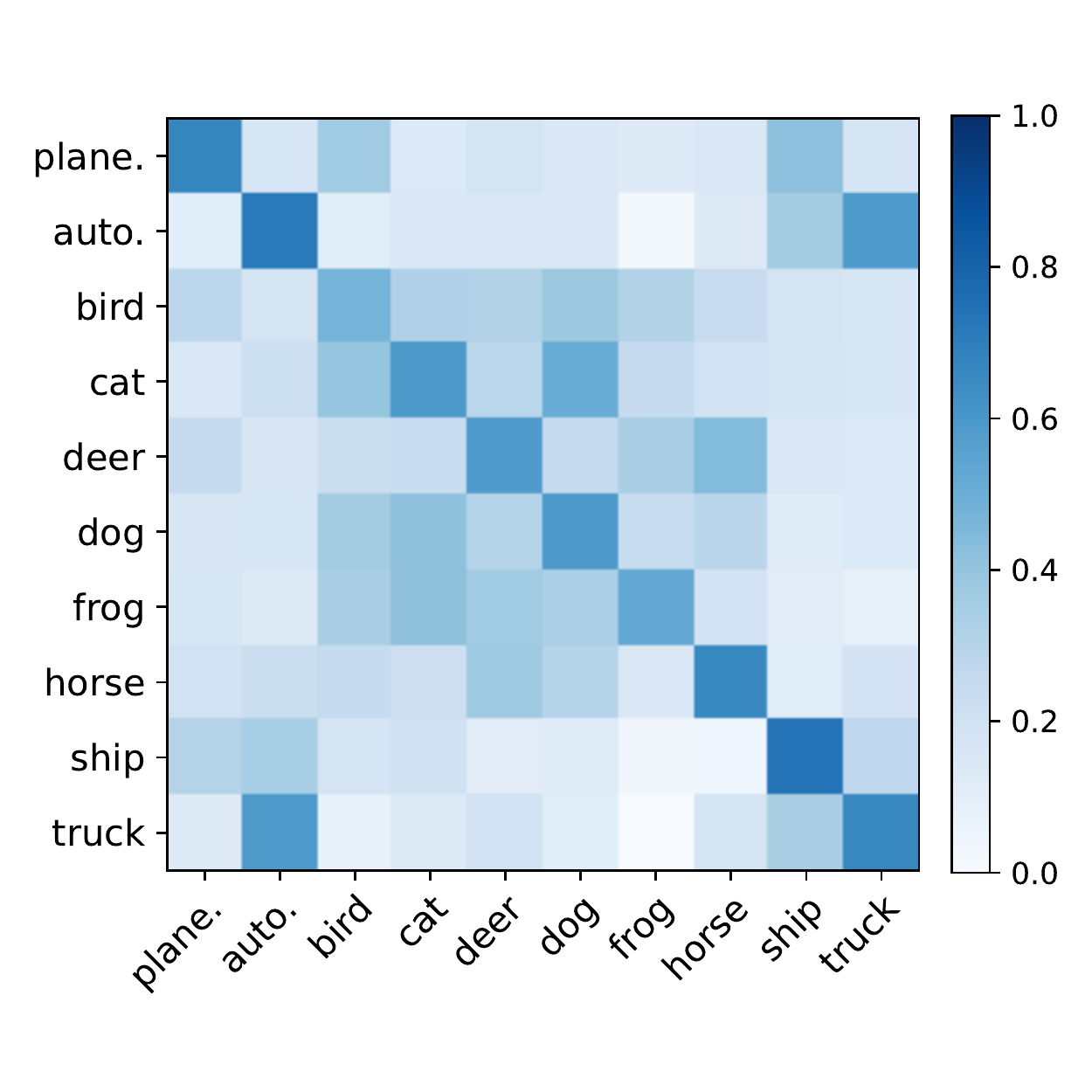}}\hfill & {\includegraphics[align=c,width=0.3\linewidth]{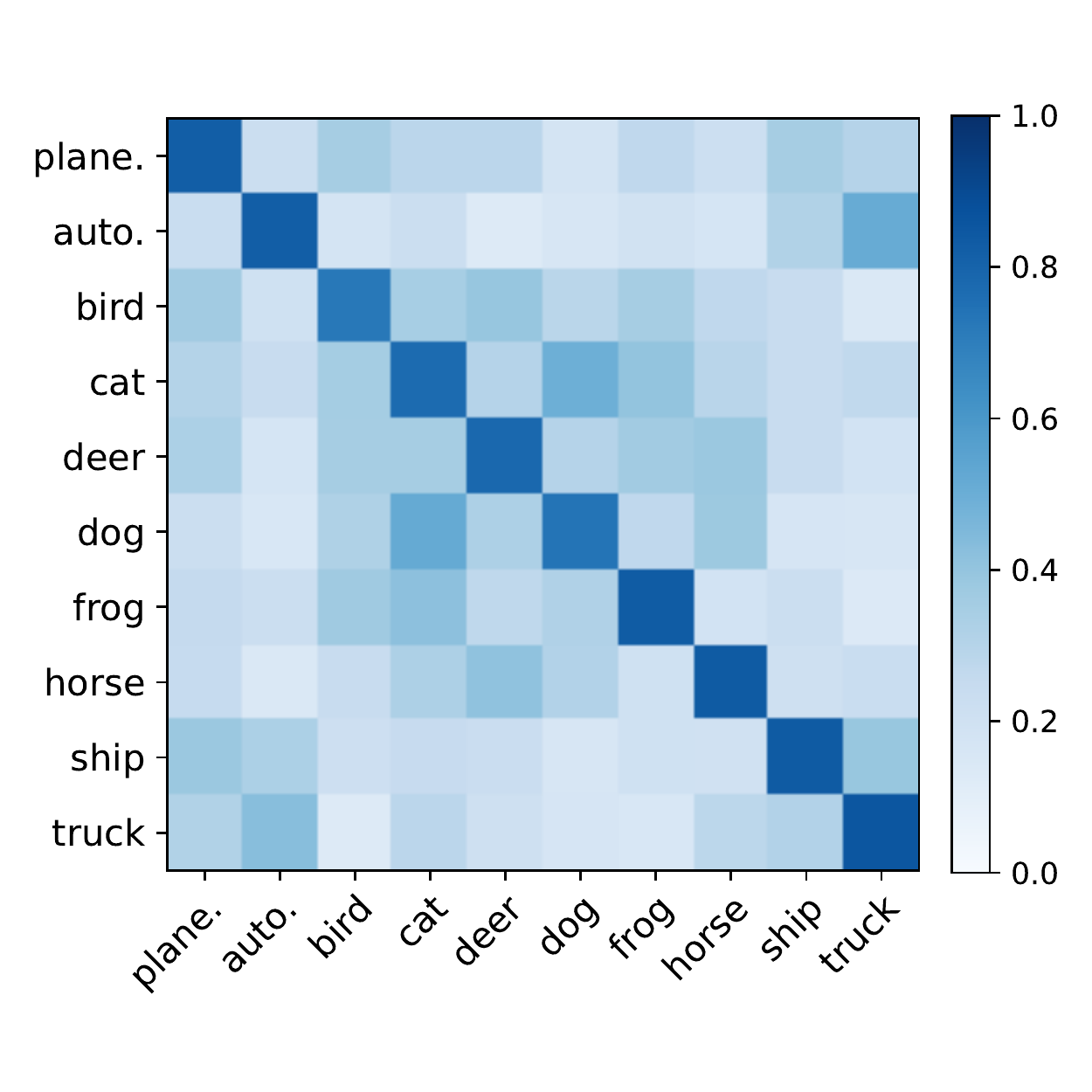}}\hfill \\ 
\hlinewd{0.8pt}
\end{tabular}
}
\end{center}
\label{Tab:ab_sim_2}
\end{table*}

\begin{table}[t]
\caption{\textbf{Comparison of training objective configuration by hand-tuned uncertainty-based thresholding and weighting.}
}

\begin{center}
\small
\begin{tabular}{c|c|c}
\hlinewd{0.8pt}
                                                                             Methods            & $\tau_{u}$ & Accuracy       \\ \hline
\multirow{4}{*}{\begin{tabular}[c]{@{}c@{}}Uncertainty-based thresholding\end{tabular}} & 1.5 & 86.86          \\
                                                                                          & 1.0 & 90.82          \\
                                                                                          & 0.5 & 89.89          \\
                                                                                          & 0.1 & 88.28          \\ \hline
\begin{tabular}[c]{@{}c@{}}Uncertainty-based weighting\end{tabular}                     & \textbf{-}   & \textbf{92.54} \\
\hlinewd{0.8pt}
\end{tabular}
\end{center}

\label{Tab:ab_uc}
\end{table}

\subsubsection{Effects of Queue Size}
To prove the superiority of AggMatch, we conduct an ablation study on the queue size. We reduce the queue size to $L=512$ which is 1/4 times smaller than the original queue size and then record 92.09\% accuracy on the CIFAR-10~\cite{krizhevsky2009learning} with 40 labels. There is only 0.45\% performance decline compared to the original queue size ($L=2,048$). It shows that our competitiveness lies in pseudo-labeling refinement and uncertainty estimation through aggregation with consistent and confident samples in the queue, not mainly depending on the queue size. However, when the queue size is set to $L=2,048$, the queue occupies only 10MB in the GPU memory space during the training. It is an extremely small in volume compared to 16GB of the backbone network, i.e., WideResNet 28-2, having 512 mini-batch sizes. Following ~\cite{he2020momentum}, adopting queue size as 65,356 for the dictionary look-up, we also increase the queue size to maximize our effectiveness.

\subsubsection{Effectiveness of Momentum Encoder}
We conduct an ablation study on the effectiveness of the momentum encoder for aggregation. The samples composing the large queue should be extracted by a slow progressing encoder structure because they come from mini-batches in different time-steps. Therefore, in order to maintain consistency between samples in the queue, we update the encoder for the queue with a momentum-based moving average as in~\cite{he2020momentum}. 

The experiment is conducted at a small queue size, e.g., $L=512$, to minimize the effectiveness of the momentum encoder because the queue can be filled with the same mini-batch samples in the same time-steps with a high probability. We confirm that the momentum encoder is significantly important for maintaining the consistency of the queue through 9.5\% drop in the performance compared to when using the momentum encoder on the CIFAR-10~\cite{krizhevsky2009learning} with 40 labels in ~\tabref{quan_table_cifar10_SVHN}.

\newcolumntype{M}[1]{>{\centering\arraybackslash}m{#1}}
\begin{table*}[t]
\caption{\textbf{Visualization on the transitions of multi-class feature distribution (denoted as dots in colors) of training samples from CIFAR-10~\cite{krizhevsky2009learning} in the feature space at different training iterations.} Data projection in 2-D space is generated by t-SNE~\cite{scikit-learn} based on the feature representation.}
\begin{center}
\small
\scalebox{0.8}{

\begin{tabular}{c|c|c|c|c|c}

\hlinewd{0.8pt}
\multirow{2}{*}{Methods} & \multirow{2}{*}{4,000 iter}   &  \multirow{2}{*}{20,000 iter}     &  \multirow{2}{*}{30,000 iter}  & \multirow{2}{*}{40,000 iter}  & 
\multirow{2}{*}{\textbf{Best}}  \\ &&&&& \\ \hline
FixMatch~\cite{sohn2020fixmatch}& {\includegraphics[align=c,width=0.2\linewidth]{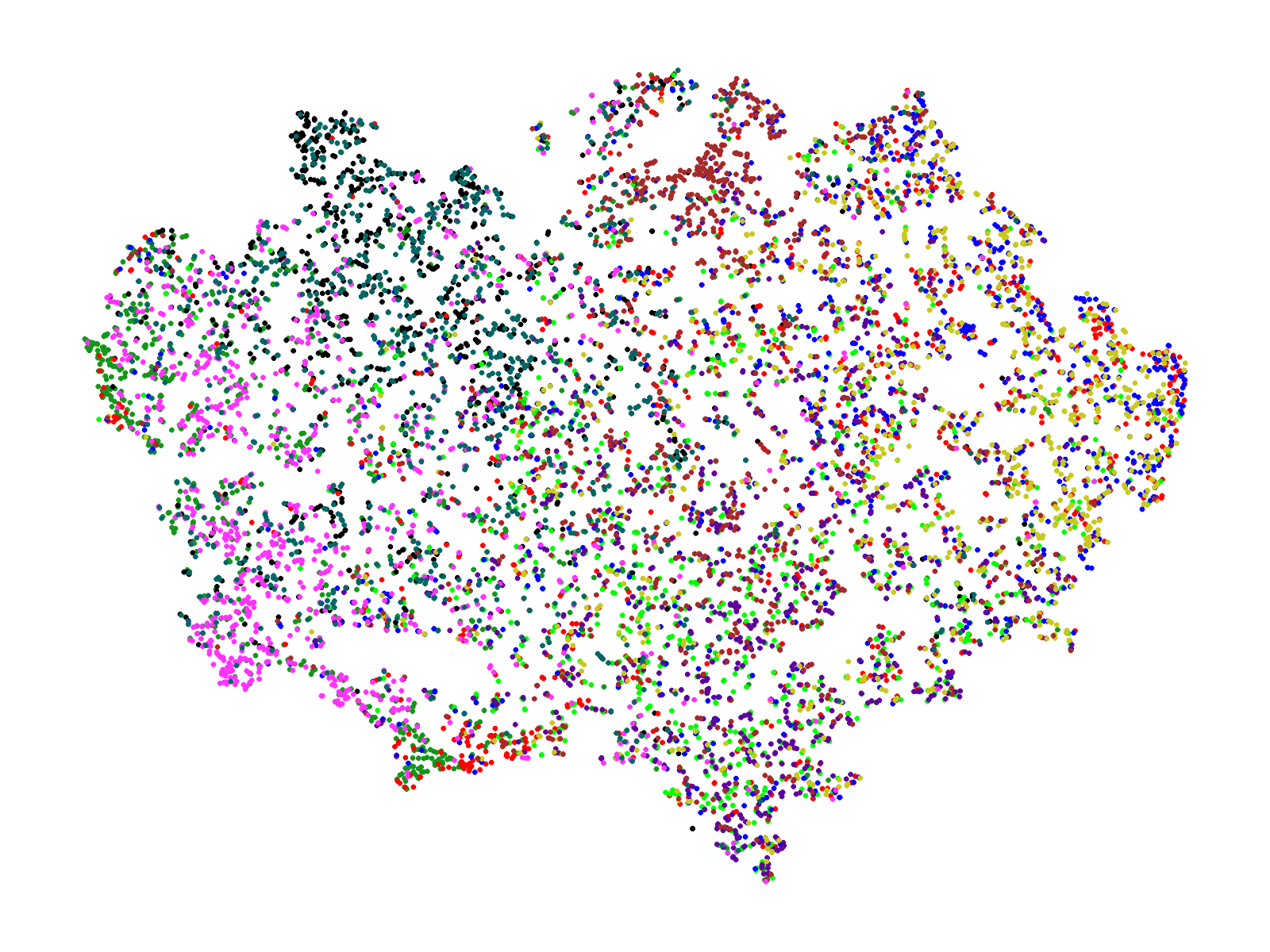}}\hfill &                     {\includegraphics[align=c,width=0.2\linewidth]{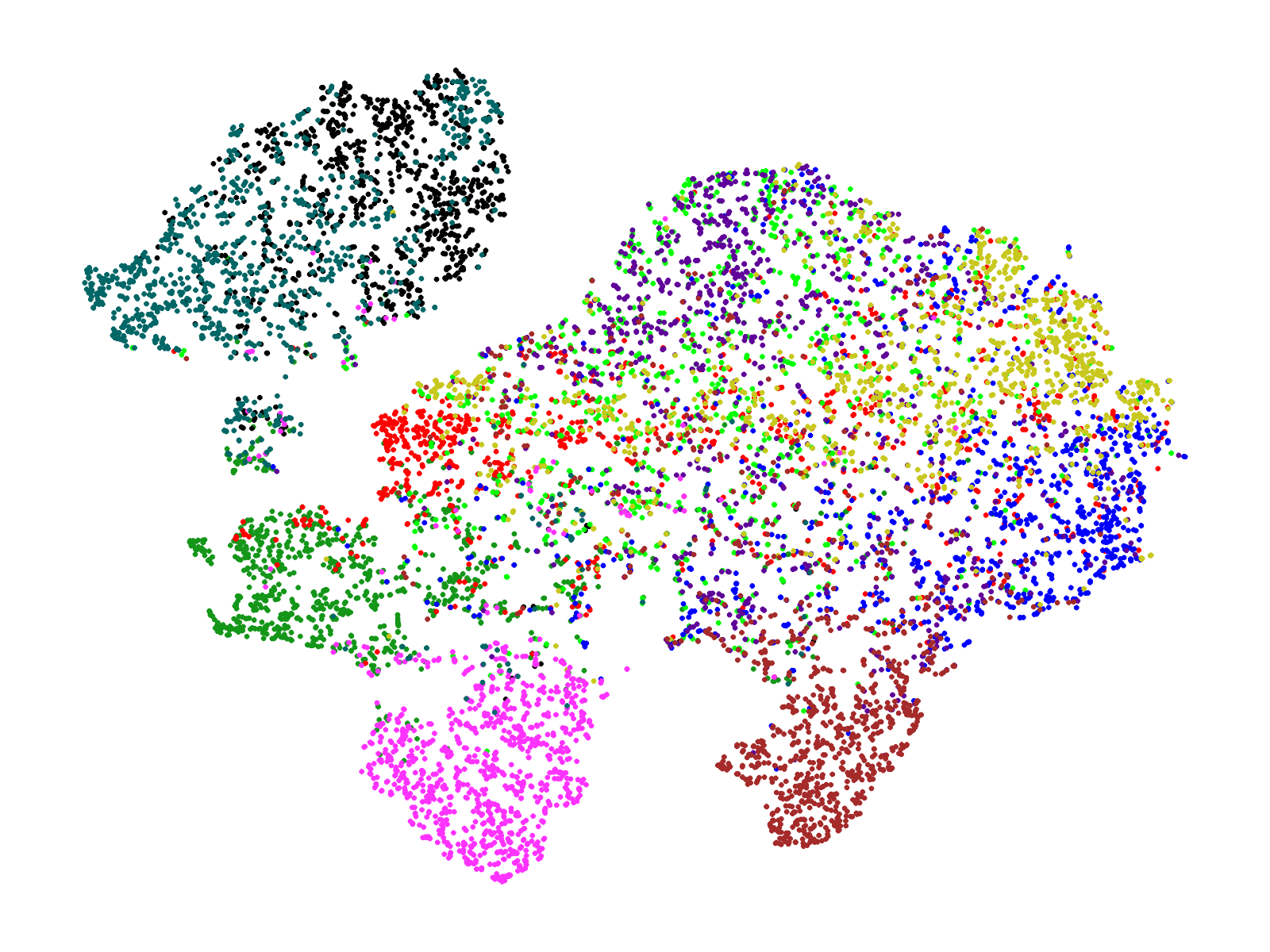}}\hfill & {\includegraphics[align=c,width=0.2\linewidth]{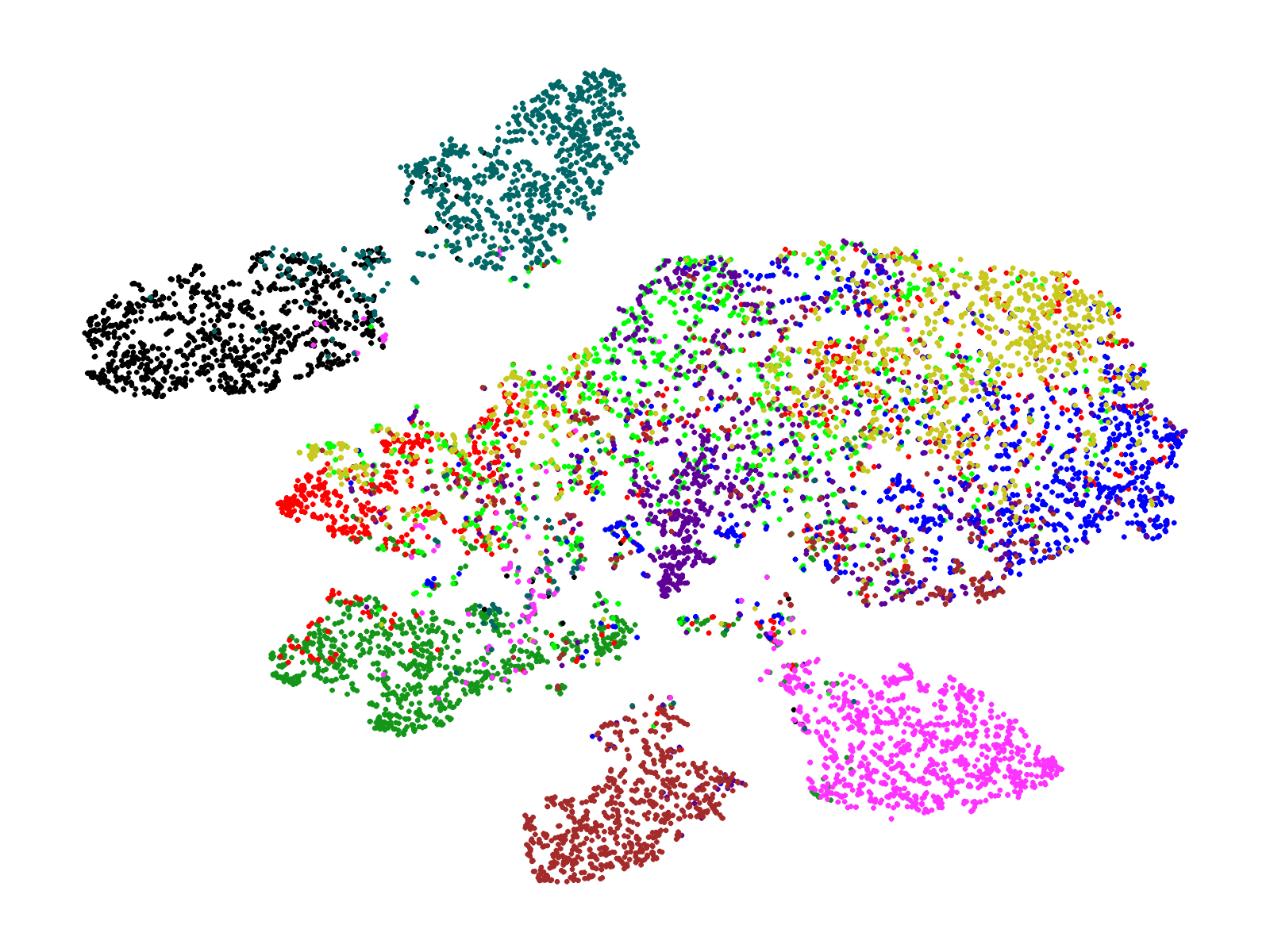}}\hfill  & {\includegraphics[align=c,width=0.2\linewidth]{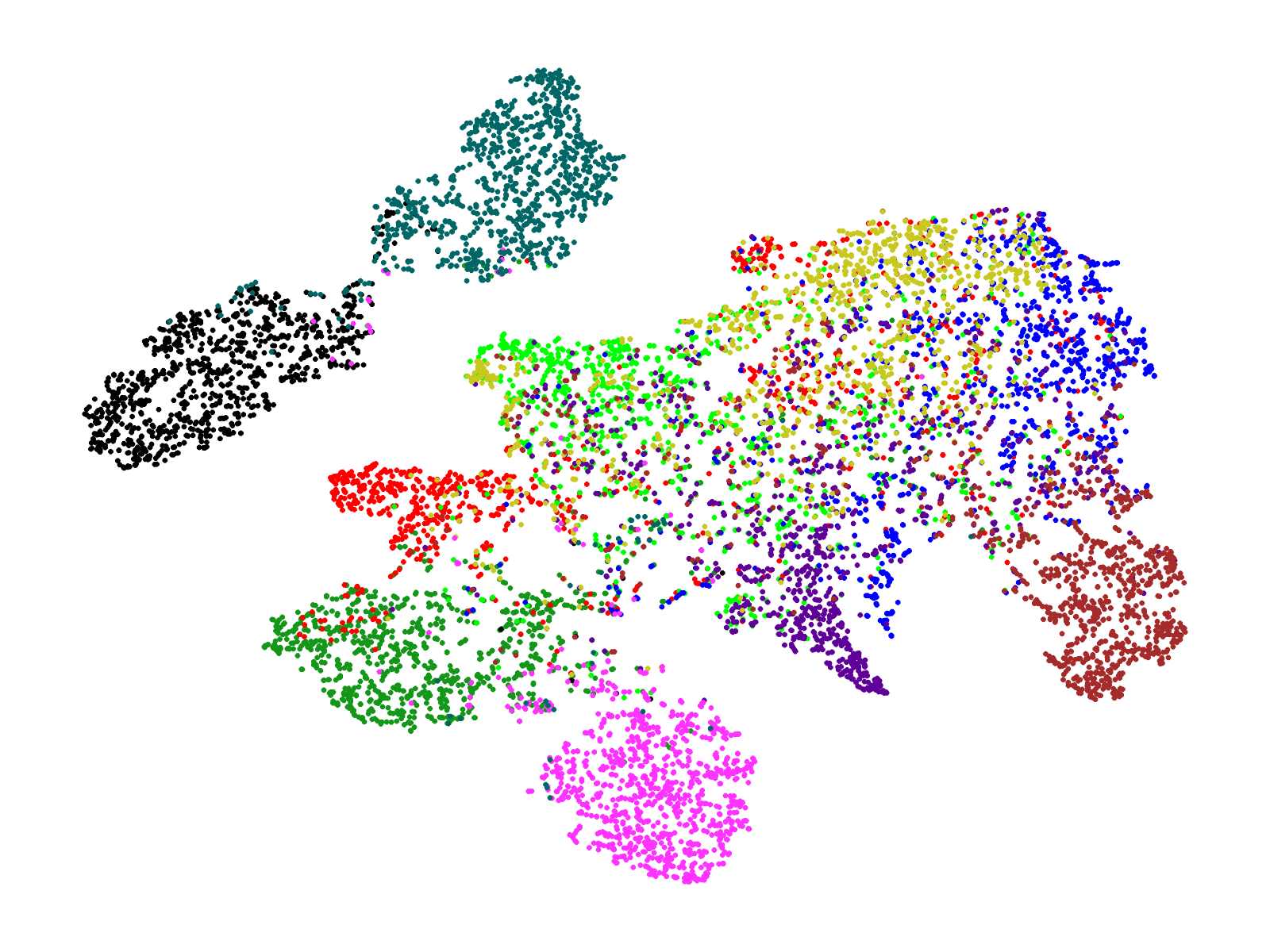}}\hfill  & {\includegraphics[align=c,width=0.2\linewidth]{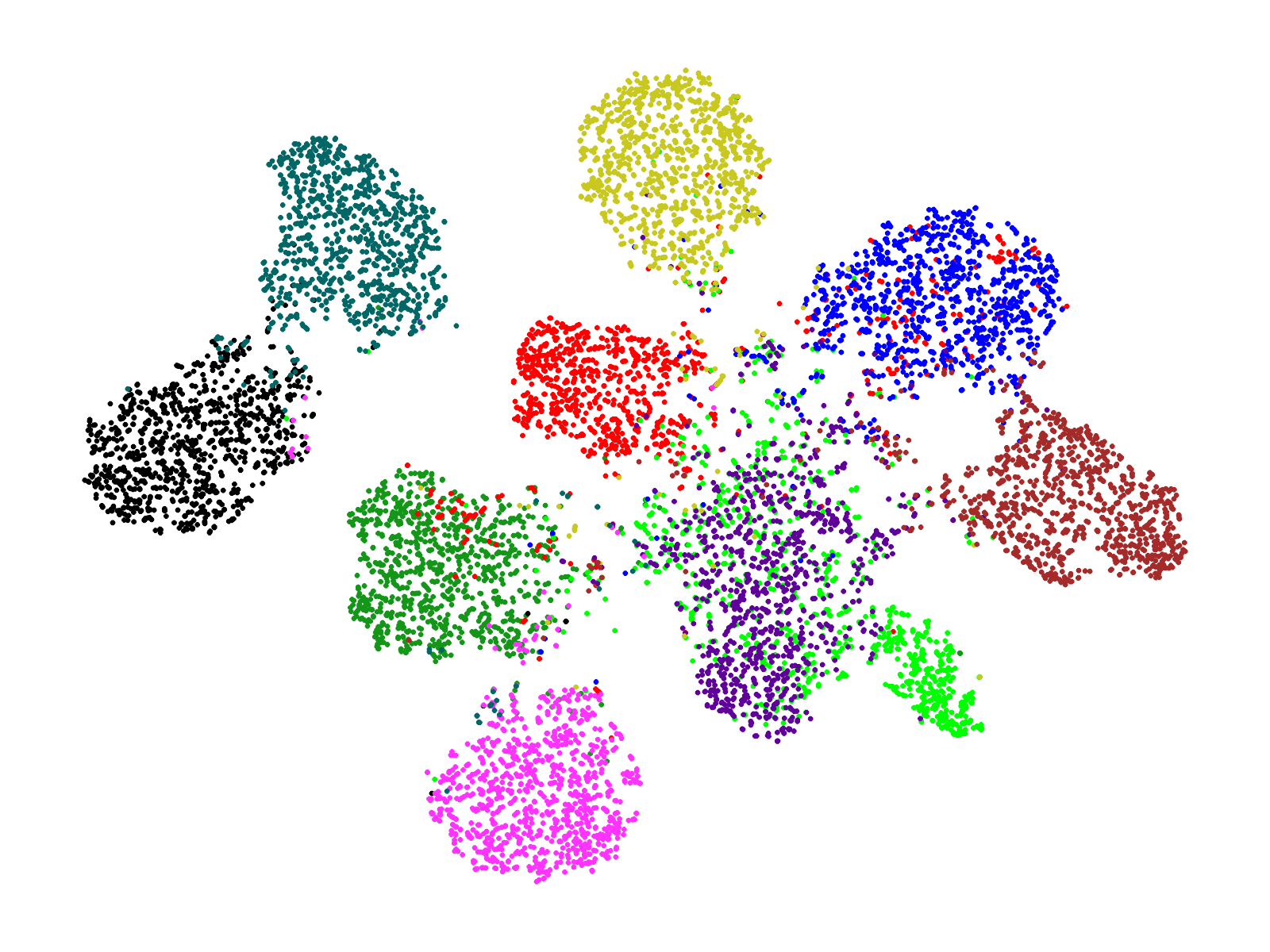}}\hfill 
\\ \hline
\textbf{AggMatch} & {\includegraphics[align=c, width=0.2\linewidth]{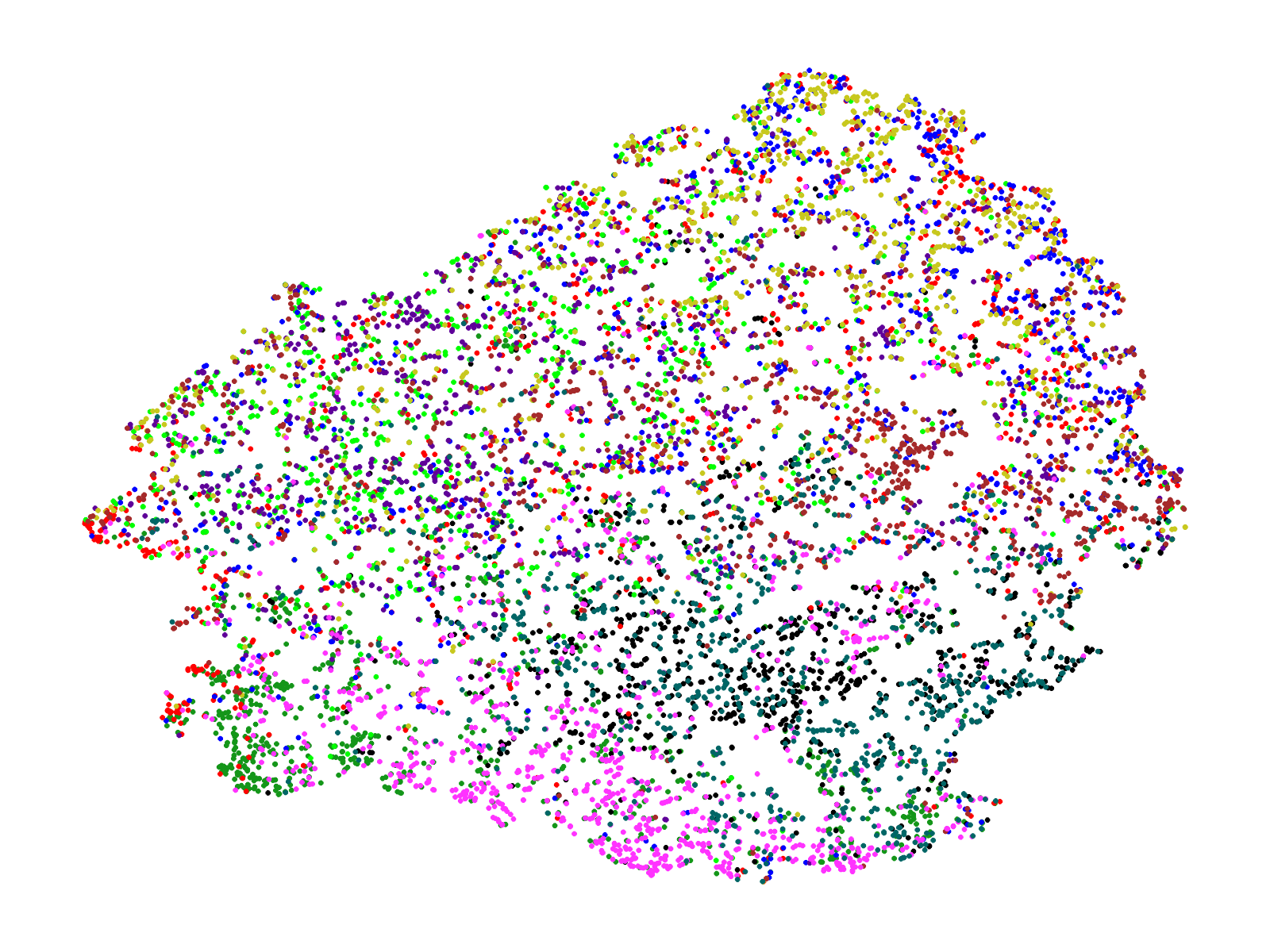}}\hfill &                     {\includegraphics[align=c,width=0.2\linewidth]{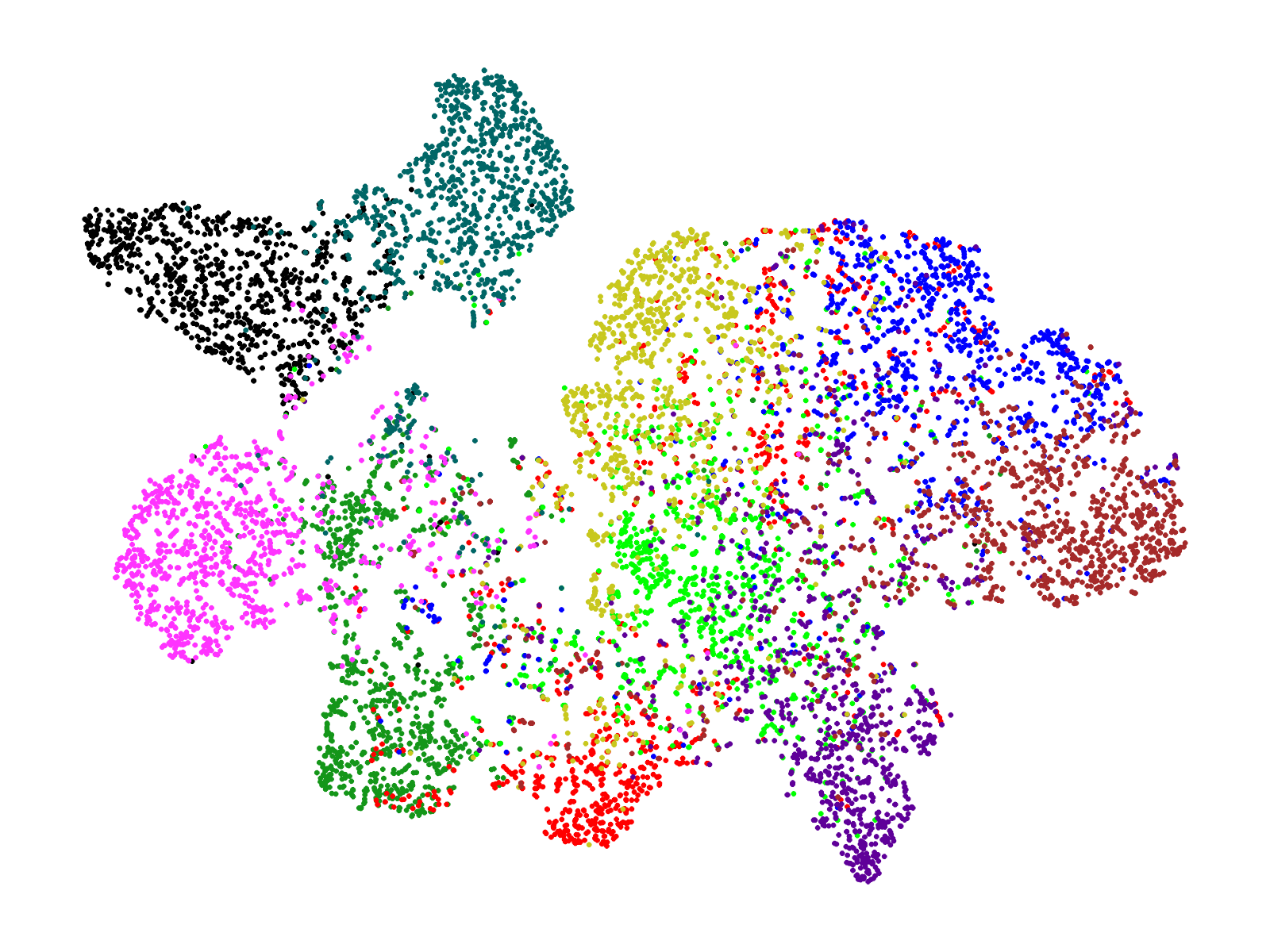}}\hfill & {\includegraphics[align=c,width=0.2\linewidth]{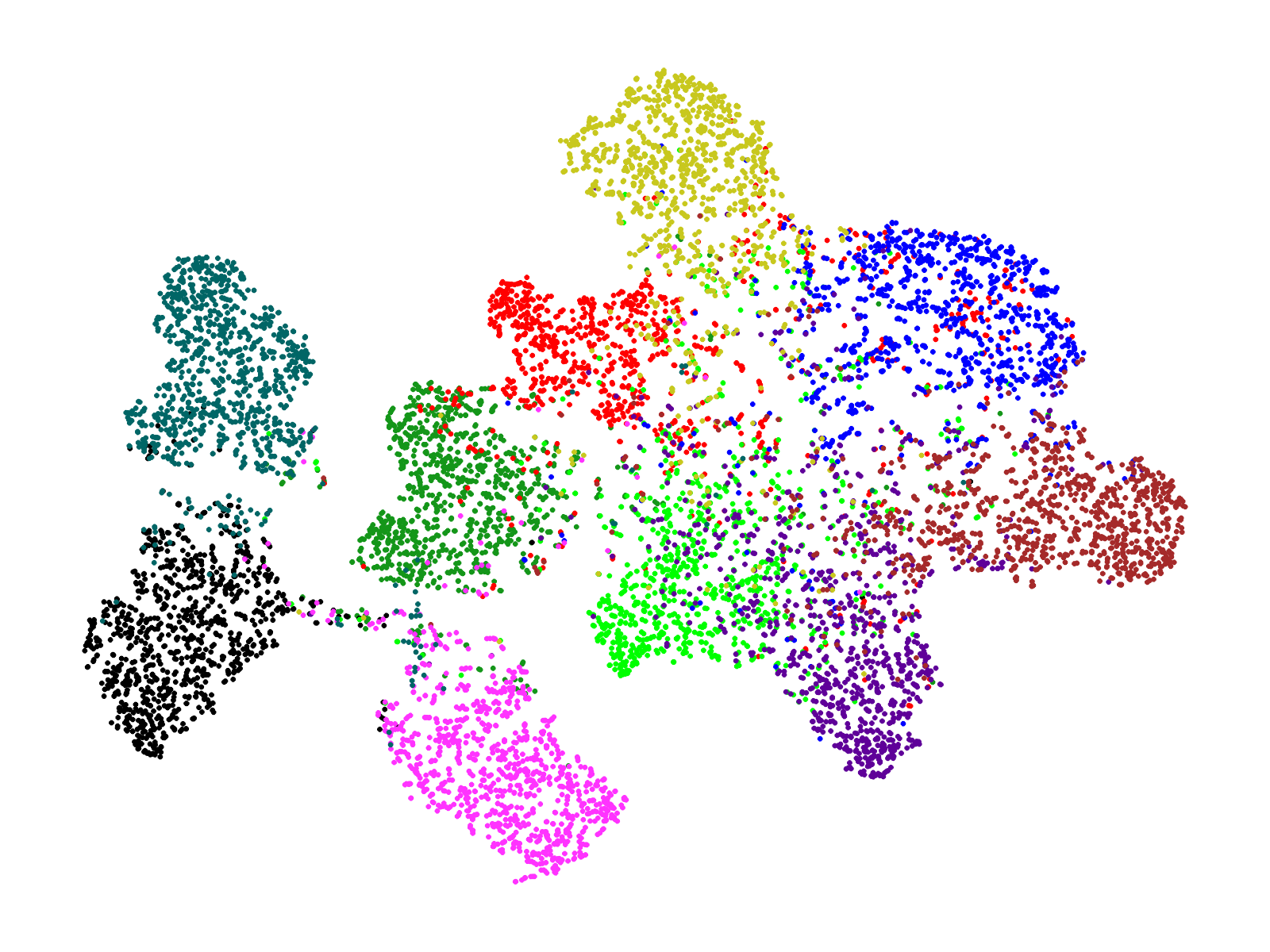}}\hfill  & {\includegraphics[align=c,width=0.2\linewidth]{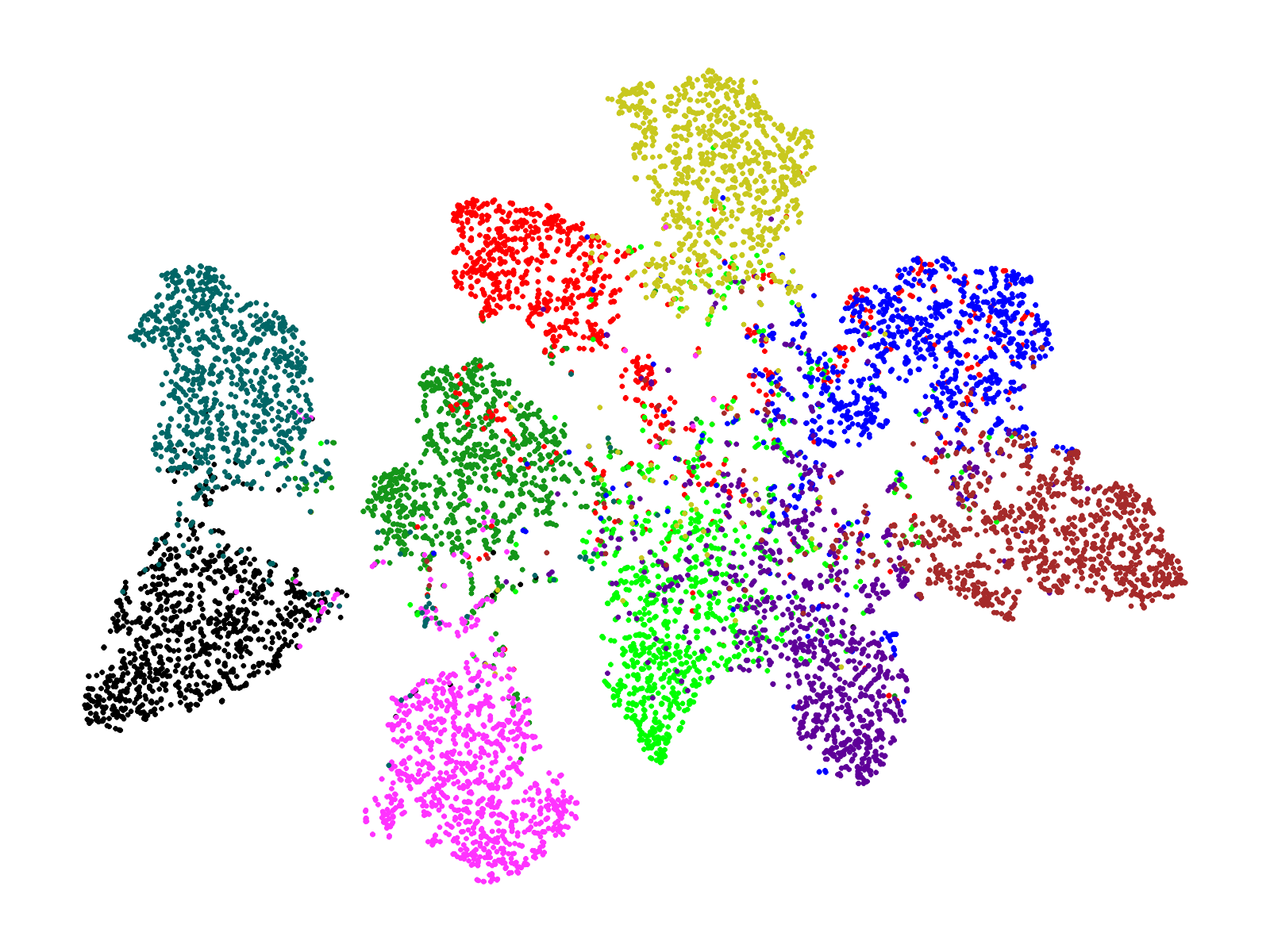}}\hfill  & {\includegraphics[align=c,width=0.2\linewidth]{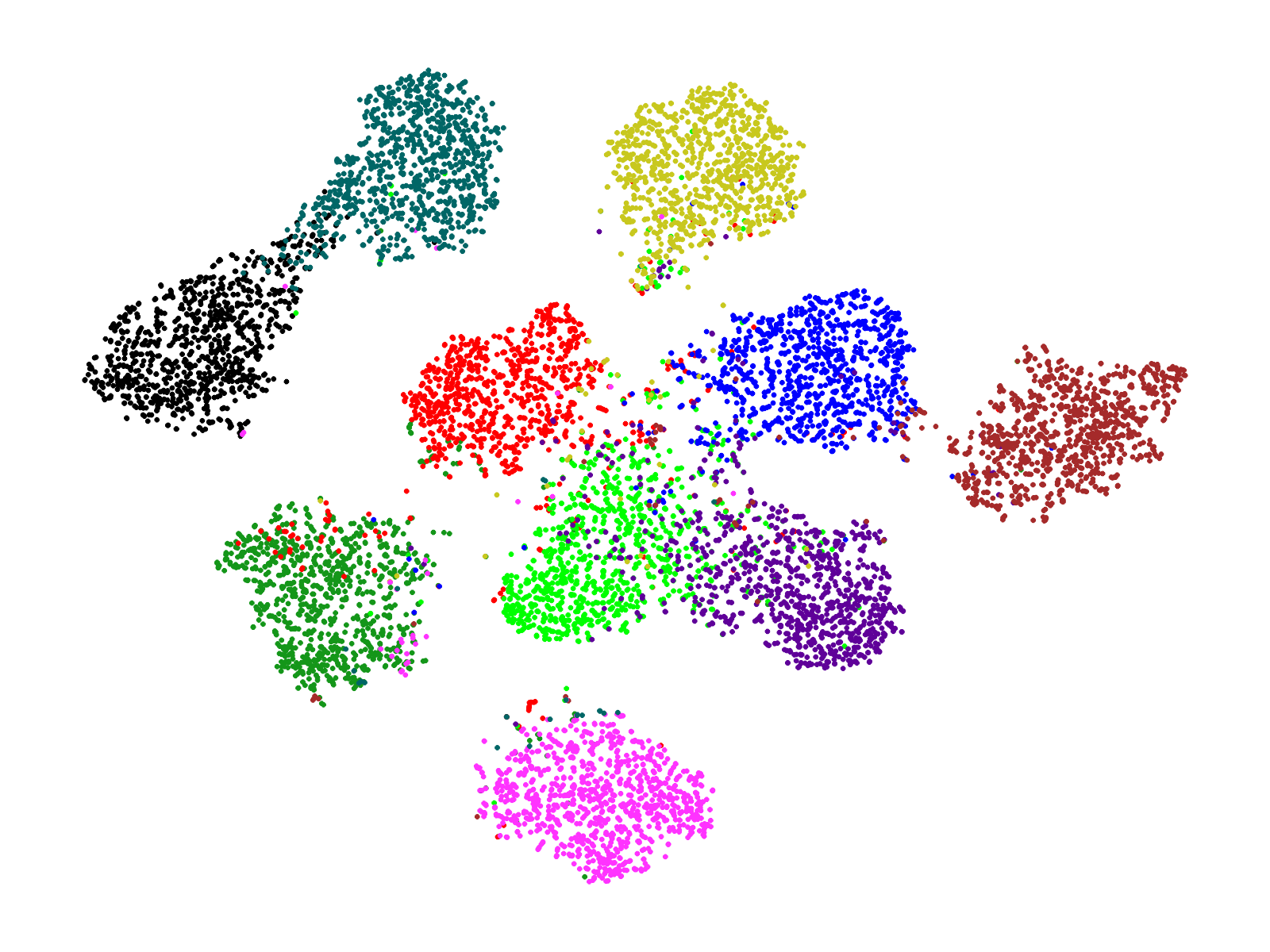}}\hfill \\ \hlinewd{0.8pt}
\end{tabular}
}\
\end{center}

\label{Tab:ana_feat_1}
\end{table*}

\begin{figure}[t]
    \centering
	\subfloat[FixMatch~\cite{sohn2020fixmatch}]
	{{\includegraphics[width=0.5\linewidth]{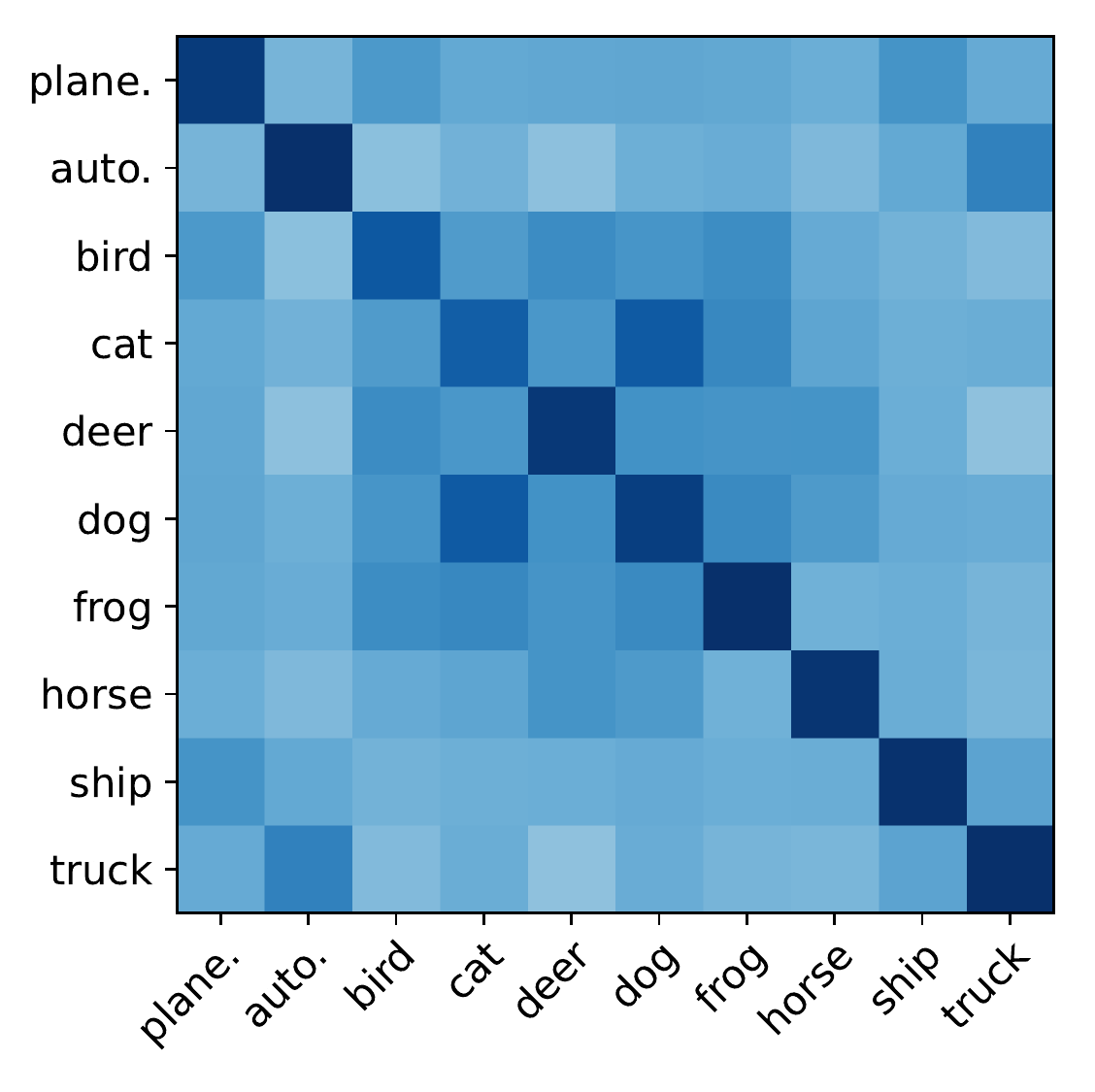}}}\hfill
	\subfloat[\textbf{AggMatch}]
 	{\includegraphics[width=0.5\linewidth]{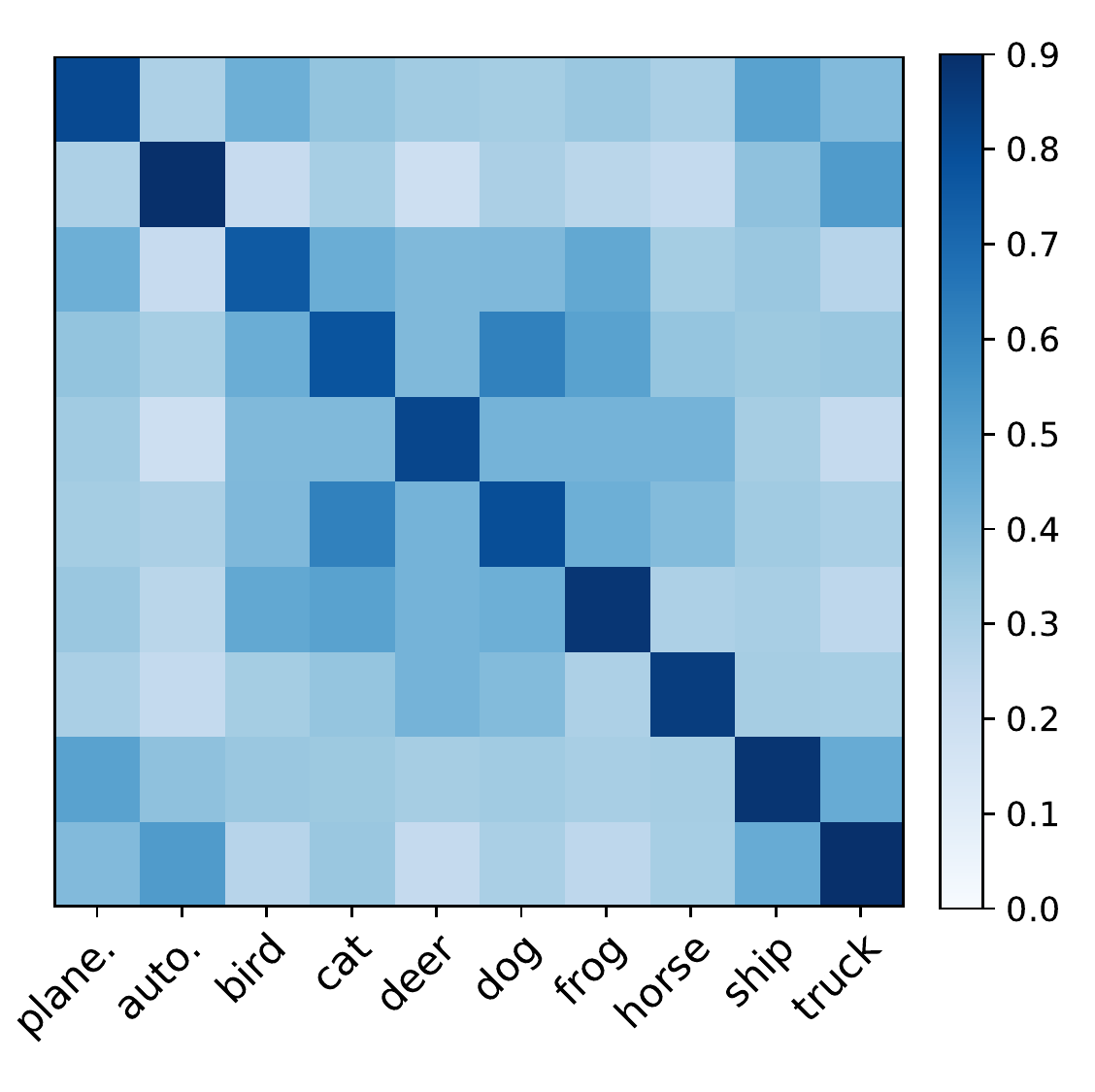}}\hfill \\
	\caption{\textbf{Visualization of feature attention map of FixMatch~\cite{sohn2020fixmatch} and AggMatch.}}
	\label{Fig:ana_feat_2}
\end{figure} 


\begin{table}[t]
    \caption{\textbf{Examples of pseudo labels by FixMatch~\cite{sohn2020fixmatch} and AggMatch.}} 
\large
\begin{center}
\scalebox{0.65}{
  \begin{tabular}
      {M{1.5cm}M{1.5cm}M{1.5cm}M{1.5cm}M{1.5cm}M{1.5cm}} 
\hlinewd{0.8pt}
      {Img} &{\includegraphics[width=0.7in]{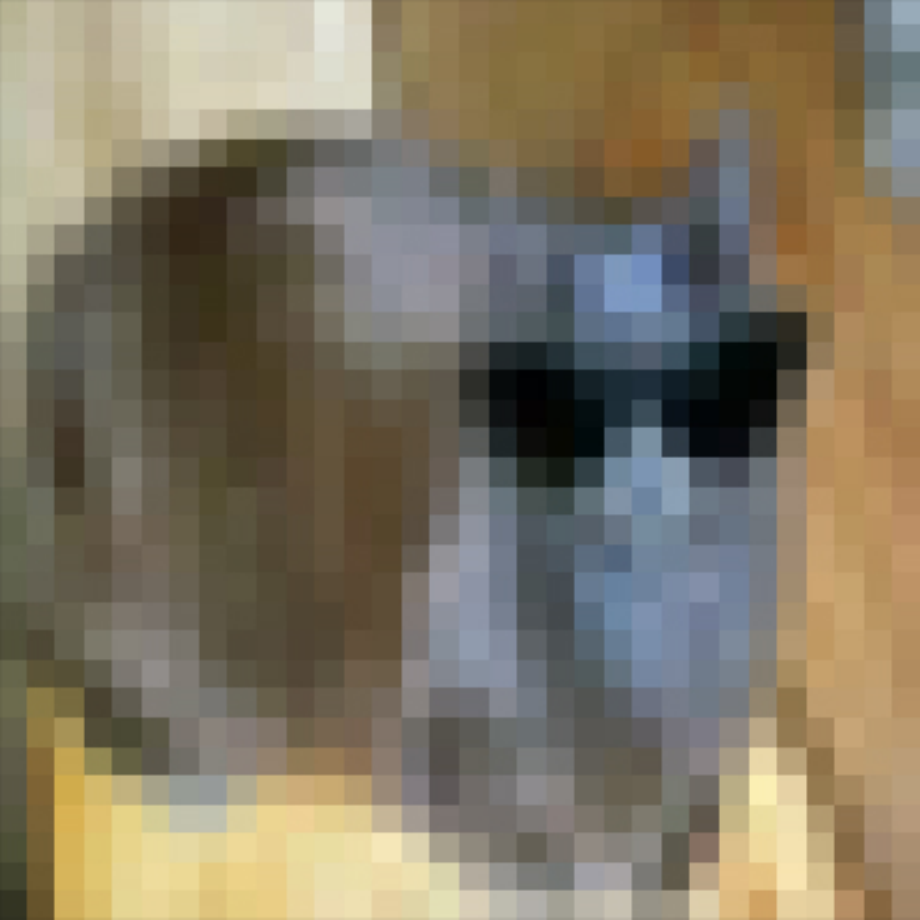}}\hfill & {\includegraphics[width=0.7in]{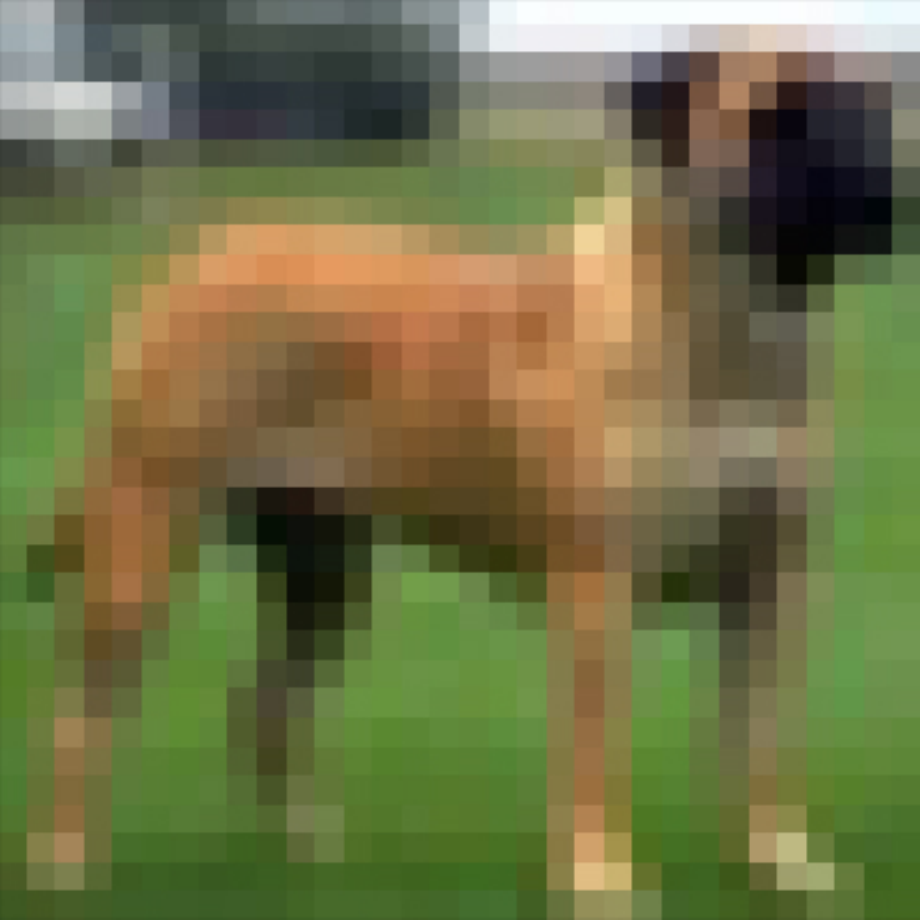}}\hfill & {\includegraphics[width=0.7in]{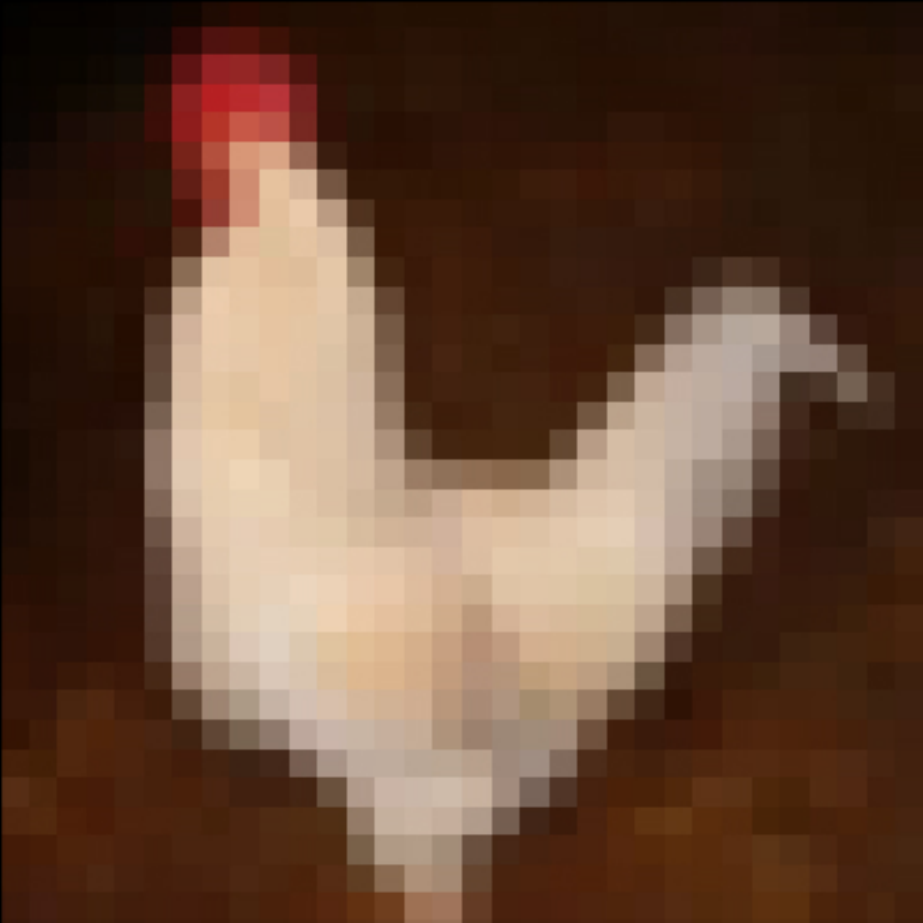}} & {\includegraphics[width=0.7in]{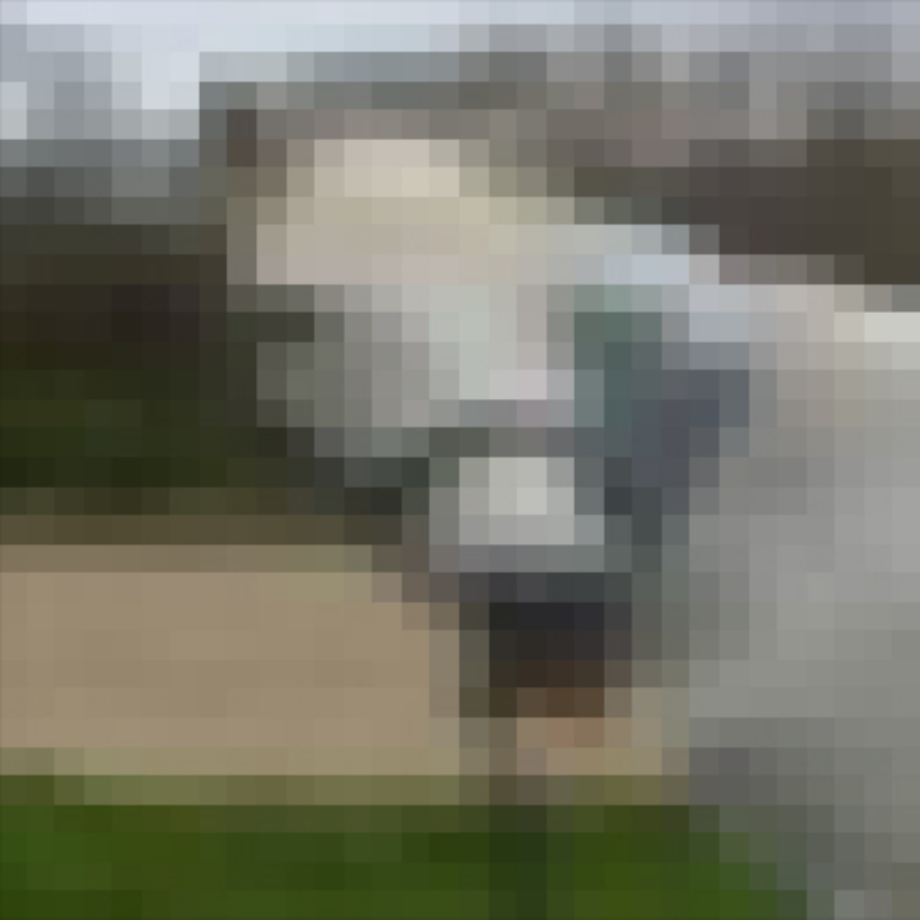}} & {\includegraphics[width=0.7in]{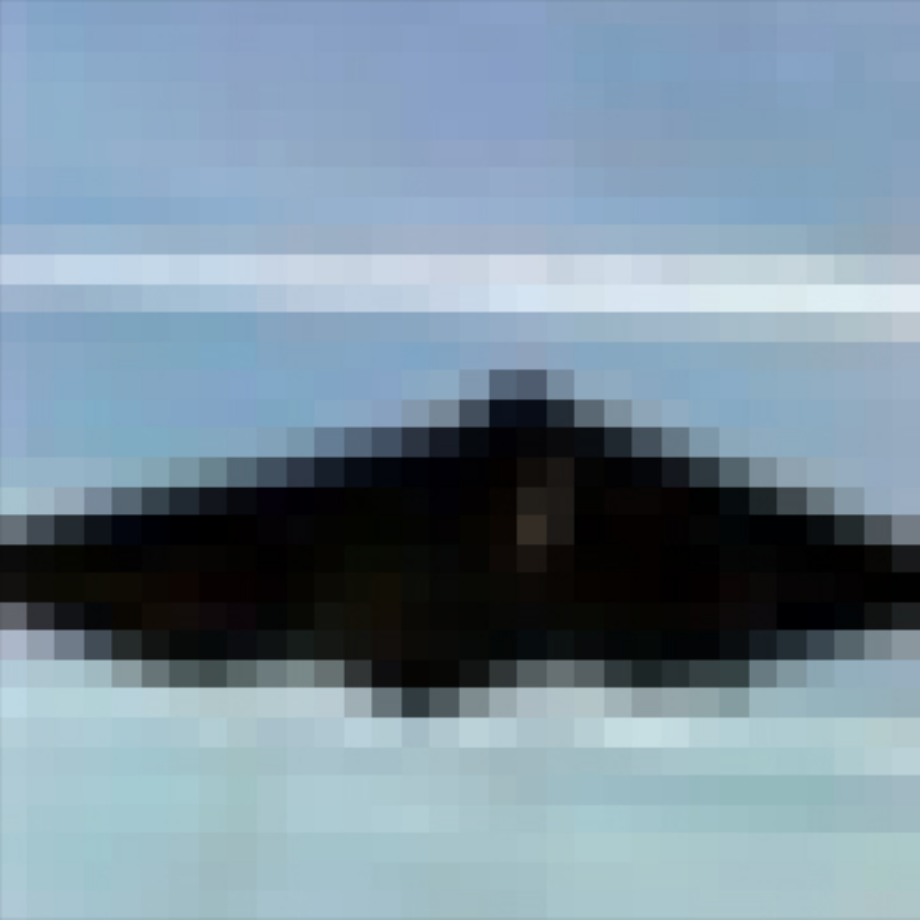}}\\
      \hline 
      GT label                  & cat & dog & bird & horse & airplane \\ \hline
      \multirow{2}{*}{FixMatch~\cite{sohn2020fixmatch}} & dog & horse & dog    & deer & cat \\
                                & (1.00) & (0.97)&(1.00) & (0.96) & (0.99) \\
      \hline
      \multirow{2}{*}{\textbf{AggMatch}}                                                               & cat  & dog    & bird & horse   & airplane\\
                                & (1.00)&(0.99)& (0.98)&(0.99)& (1.00)\\
\hlinewd{0.8pt}
  \end{tabular}
  }
\end{center}
\label{Tab:ana_class_1}
\end{table}

\begin{table}[t]
\caption{
\textbf{Visualization on the predictions of selected unlabeled samples from CIFAR-10~\cite{krizhevsky2009learning}.} The blue bar corresponds to the misclassified class. The yellow bar corresponds to the ground-truth class and the value of the ground-truth class is represented below the bar graph.}
\begin{center}
\large
\scalebox{0.7}{
\begin{tabular}{ccccc} 
\hlinewd{0.8pt}

Img & {\includegraphics[align=c,width=0.7in]{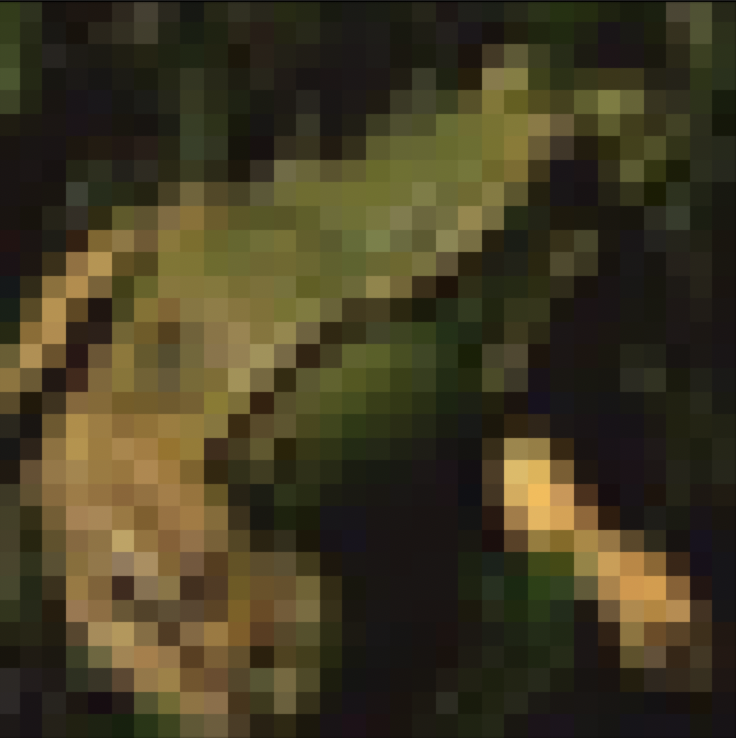}}\hfill &                     {\includegraphics[align=c,width=0.7in]{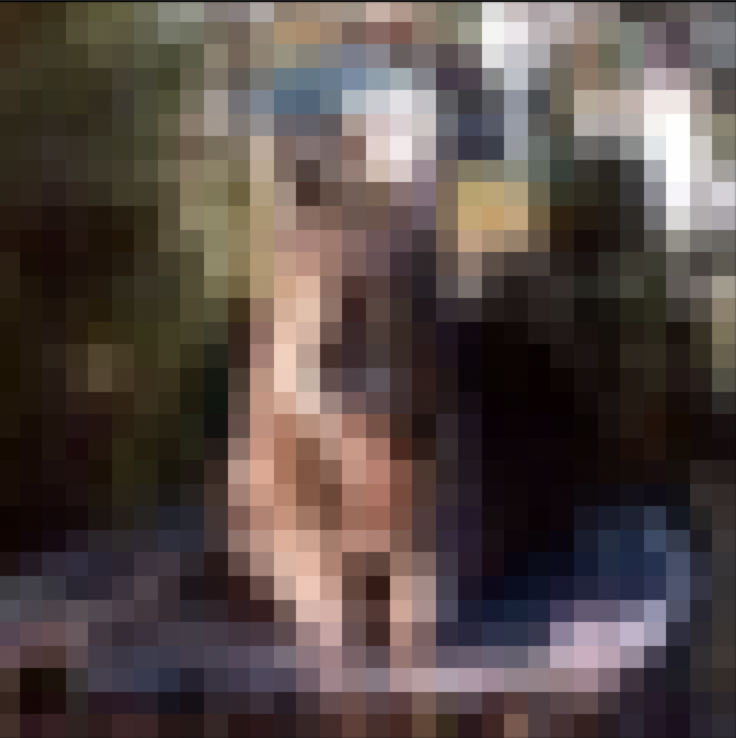}}\hfill & 
{\includegraphics[align=c,width=0.7in]{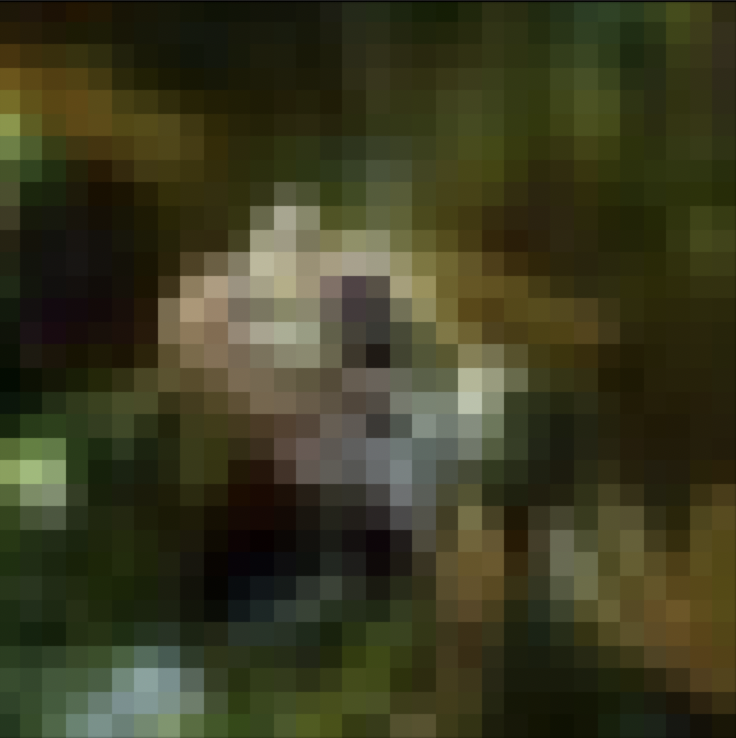}}\hfill  &
{\includegraphics[align=c,width=0.7in]{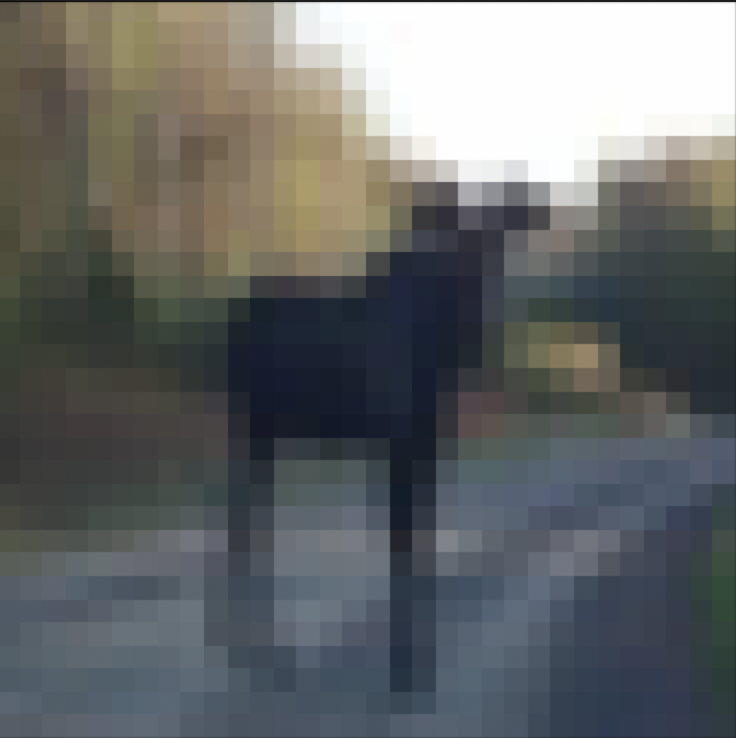}}\hfill  \\
\hline
GT label & frog  & cat & frog & dog \\\hline
before  & {\includegraphics[width=0.25\linewidth]{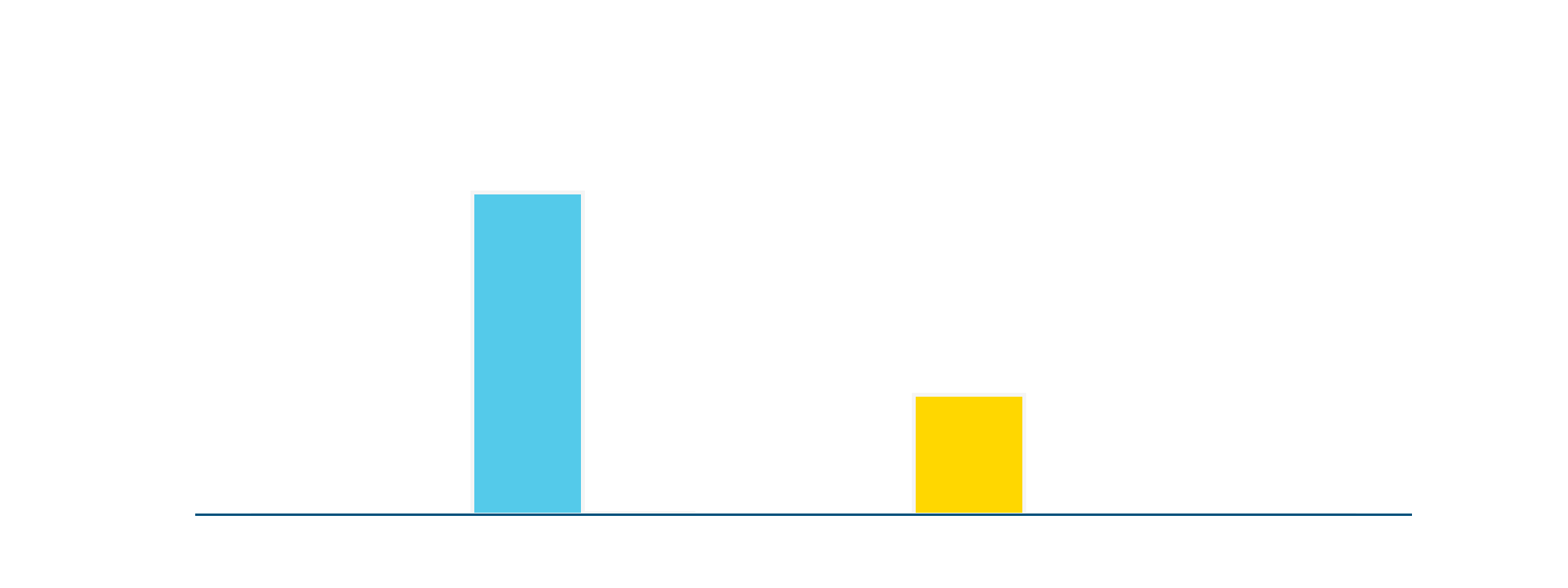}}\hfill &                 {\includegraphics[width=0.25\linewidth]{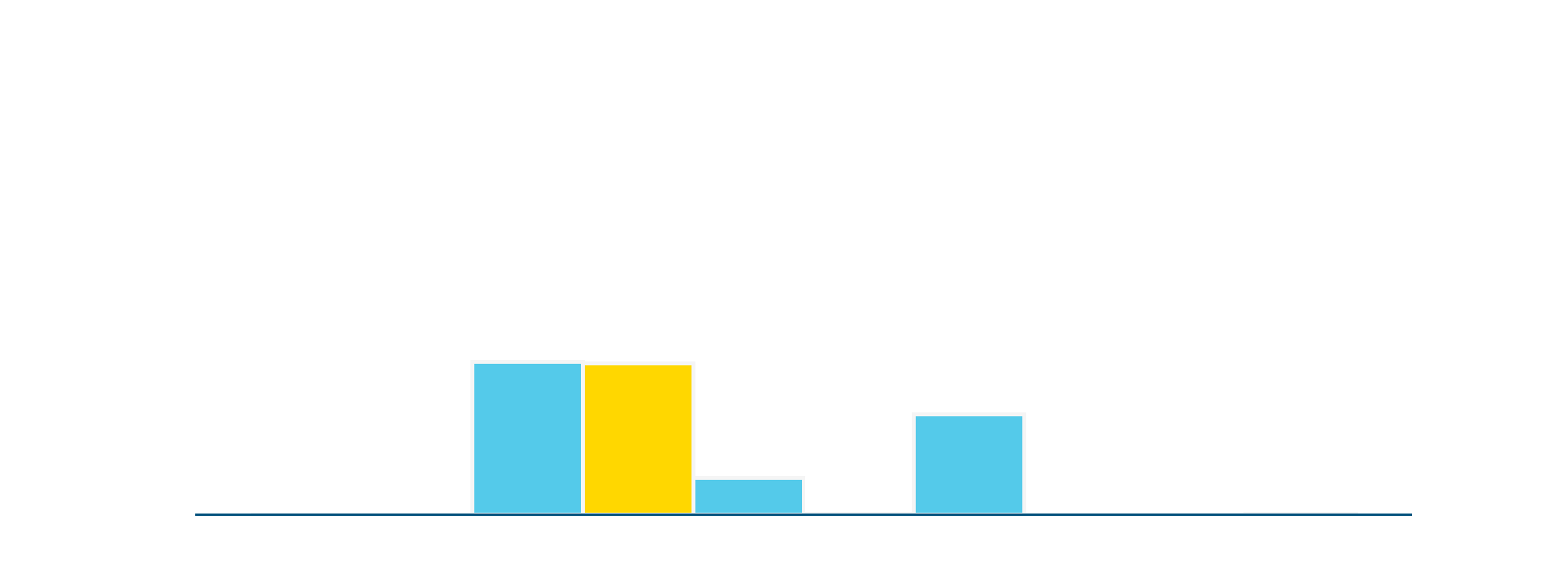}}\hfill &
{\includegraphics[width=0.25\linewidth]{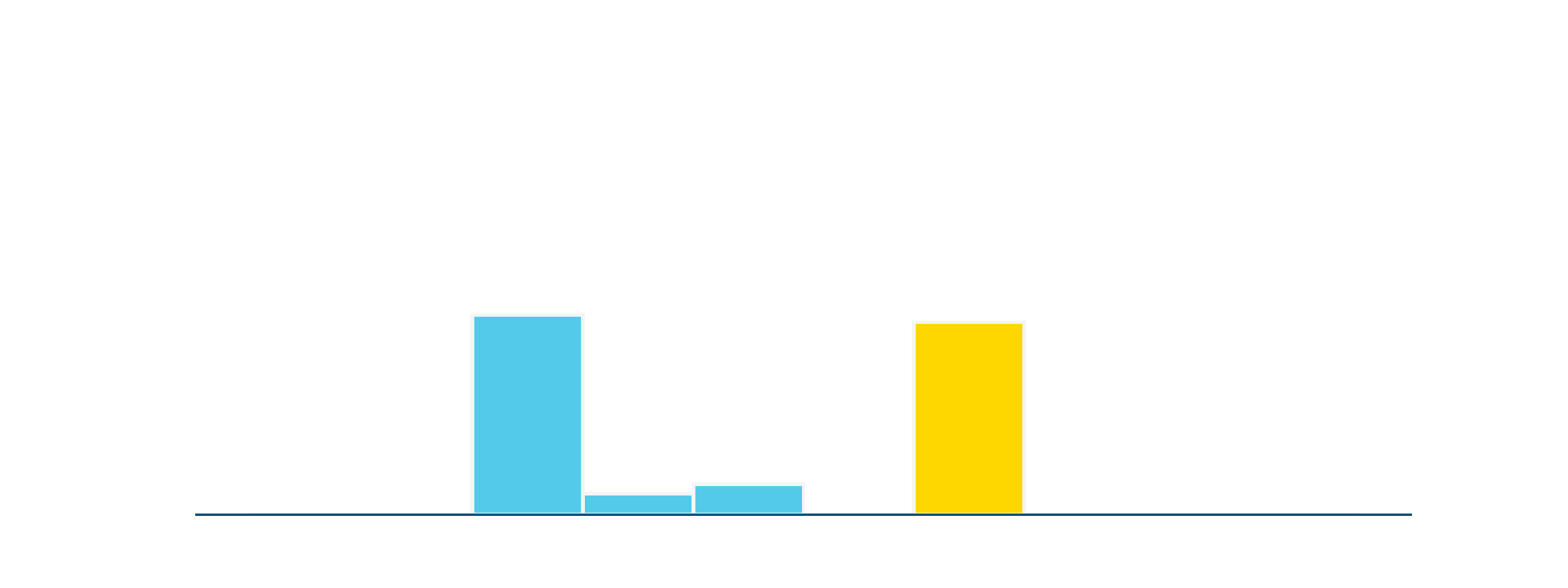}}\hfill  &
{\includegraphics[width=0.25\linewidth]{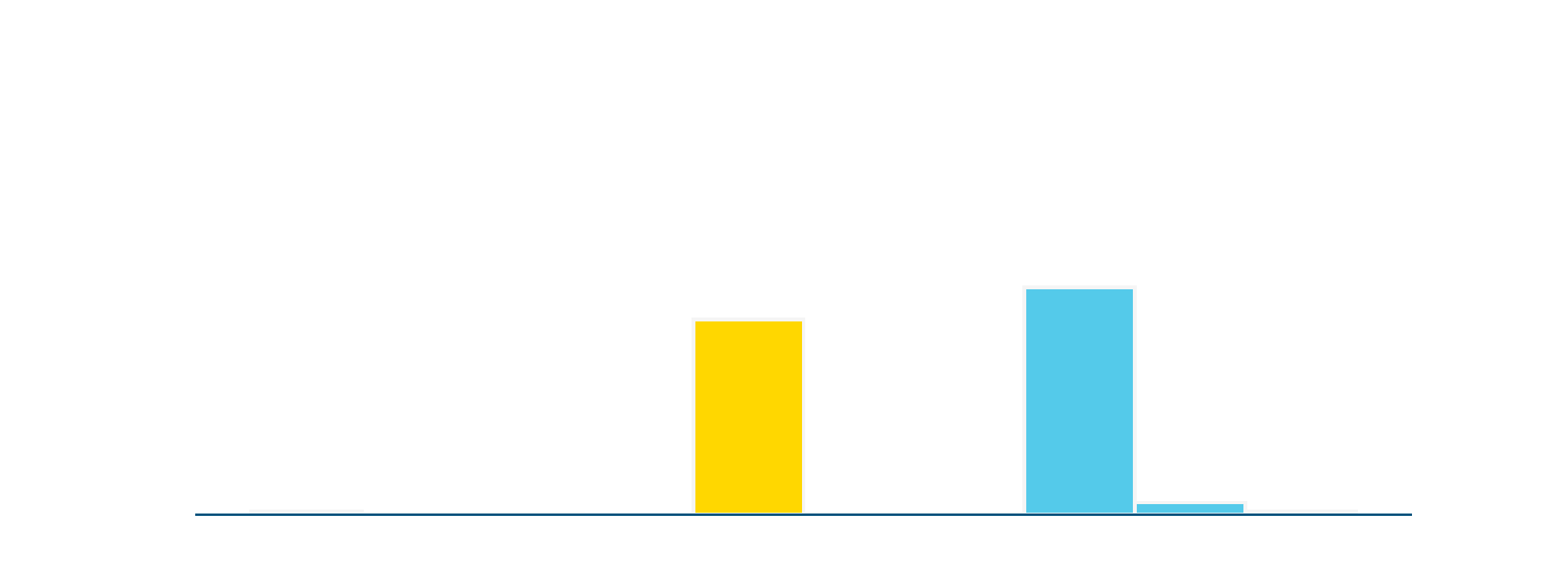}}\hfill \\
 & 0.27  & 0.34 & 0.43 & 0.44 \\
\hline
after  & {\includegraphics[width=0.25\linewidth]{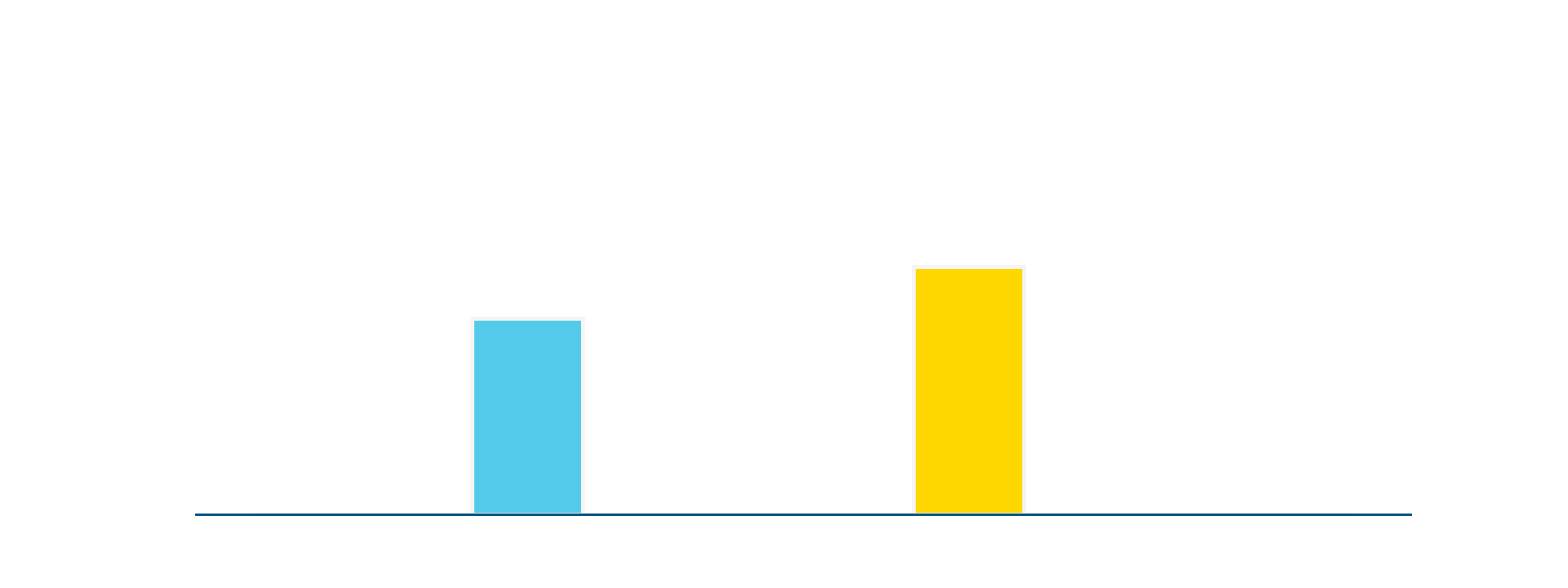}}\hfill &                     {\includegraphics[width=0.25\linewidth]{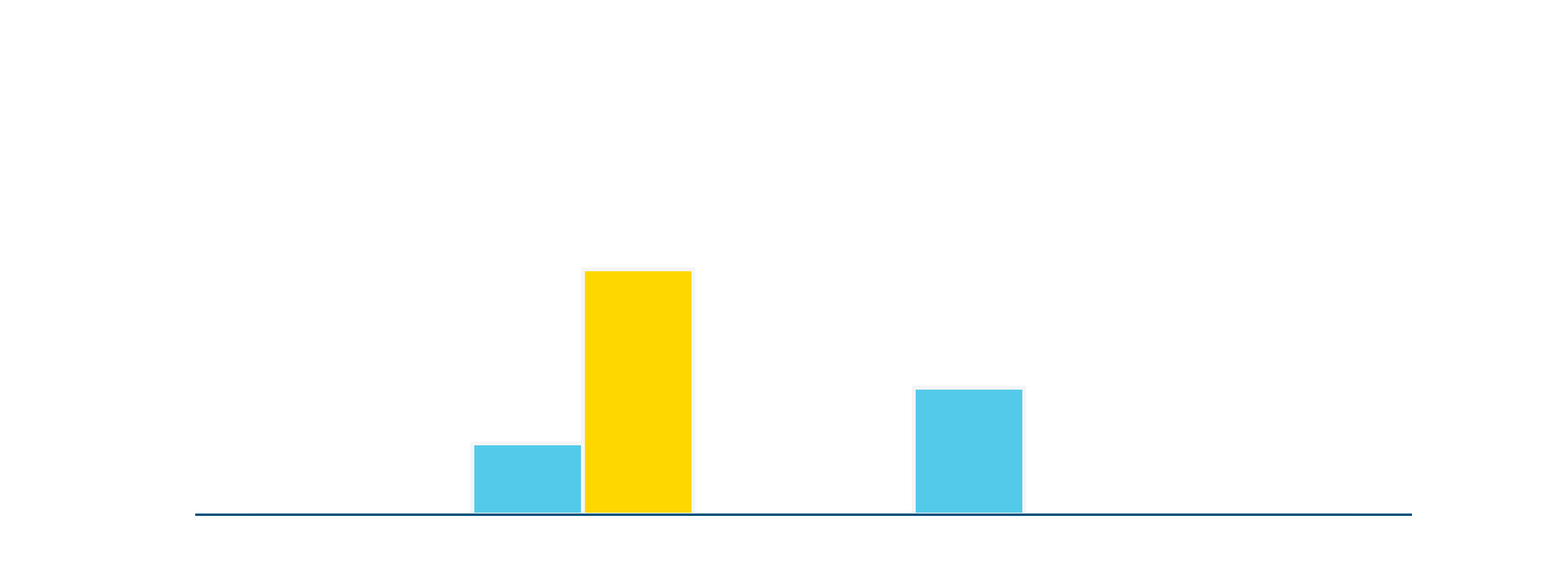}}\hfill &
{\includegraphics[width=0.25\linewidth]{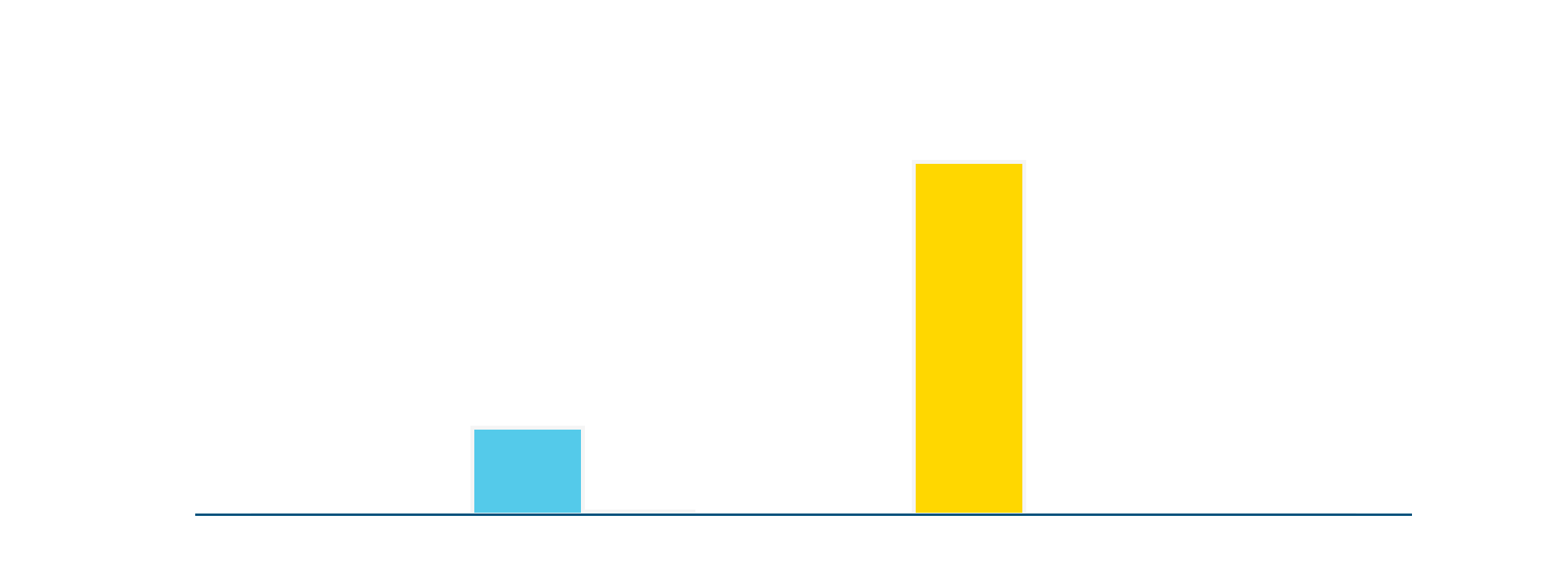}}\hfill  &
{\includegraphics[width=0.25\linewidth]{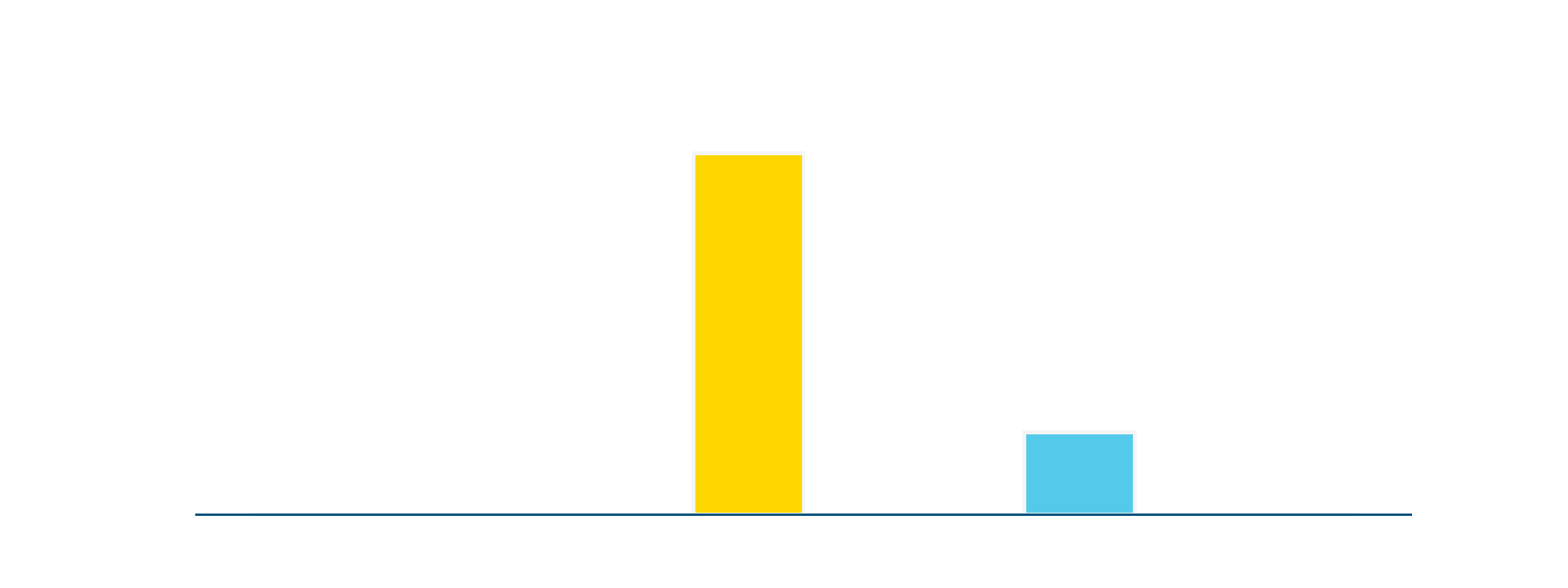}}\hfill  \\
 & 0.56  & 0.55 & 0.79 & 0.81 \\
\hlinewd{0.8pt}
\end{tabular}
}\end{center}
\label{Tabl:ana_class_2}
\end{table}

\subsection{Analysis}
\label{sec:analysis}
In this section, we conduct intensive experiments to analyze the effectiveness of our key contributions, a similarity function and a confidence estimation, respectively. We provide extensive visualization using the model weights from specific training iterations in CIFAR-10~\cite{krizhevsky2009learning} test dataset. We also use the same experiment settings as~\secref{sec:ablation}.

\subsubsection{Visualization of Feature Distribution}
We compare the feature distribution between AggMatch and FixMatch~\cite{sohn2020fixmatch} as shown in~\tabref{Tab:ana_feat_1}. For visualizing high-dimensional representations of model predictions, the complex feature matrices are then transformed into two-dimensional points by using t-Distributed Stochastic Neighbor Embedding (t-SNE)~\cite{scikit-learn} to effectively display the feature manifolds. As training iteration progresses, the t-SNE of AggMatch shows that the structure of data points is increasingly well-organized and the misclassified regions where two representations from different classes coexist are also significantly decreased. This indicates that our aggregation module effectively refines pseudo labels based on well-organized representations that express distinctive characteristics of each class. 

~\figref{Fig:ana_feat_2} visualizes attention distribution through the learned attention weights from in-batch of images to quantify the relationship between within-class and across-class relationships. We can confirm that the weight of the diagonal position from AggMatch, equivalent to the inter-class similarity, is significantly higher than other regions. Compared to FixMatch~\cite{sohn2020fixmatch}, it can be seen that the difference between within-class and across-class attention weights is much larger, which means that AggMatch has distinctive feature representations for each class.     
 

\subsubsection{Visualization of Pseudo-labeling Refinement}

Wrong pseudo-labeling of unlabeled samples with high confidence is considered the main factor of the confirmation bias. As shown in~\tabref{Tab:ana_class_1}, FixMatch~\cite{sohn2020fixmatch} has the pseudo-label misclassification problem, even using high confident pseudo labels, since it relies solely on its own model output of the sample without considering the relationship between other samples. On the other hand, AggMatch aggregates pseudo labels by considering the relationship between confident-aware samples in the queue and finally generates more accurate pseudo labels. 
~\tabref{Tabl:ana_class_2} visualizes the pseudo-labeling refinement of AggMatch by comparing the class distribution before and after aggregation. It can be seen that the class probability, which would has the wrong one-hot pseudo label by the argmax operation is refined to the appropriate pseudo labels after aggregation. 

\section{Conclusion}
\label{sec:conc}
In this paper, we have proposed a novel semi-supervised learning (SSL) framework, dubbed AggMatch, that aggregates initial pseudo labels by considering the similarity between the different instances to achieve more confident pseudo labels. We have introduced a class-balanced confidence-ware queue with the momentum model for more stable and consistent aggregation. We have also proposed a novel confidence measure for the pseudo label by considering the consensus among multiple pseudo labels with different subsets of the queue. We have shown that our method surpasses the current state-of-the-art in several benchmarks even with noise settings. A natural next step, which we leave for future work, is to examine how AggMatch could be used to learn a model in a self-supervised setting without any labeled data.
\bibliography{egbib}
\bibliographystyle{IEEEtran}

\end{document}